\documentclass[twoside,11pt]{article}

\usepackage[preprint]{jmlr2e}

\newcommand*\highlight{%
\color{red}}

\ShortHeadings{Benchmarking Graph Neural Networks}{Dwivedi, Joshi, Luu, Laurent, Bengio and Bresson}
\firstpageno{1}

\usepackage[utf8]{inputenc} 
\usepackage[T1]{fontenc}    
\usepackage{hyperref}       
\usepackage{url}            
\usepackage{booktabs}       
\usepackage{amsfonts}       
\usepackage{nicefrac}       
\usepackage{microtype}      
\usepackage{verbatimbox}

\usepackage{graphicx}
\usepackage{multirow}
\usepackage{xcolor}
\usepackage{subfig}
\usepackage{caption}
\usepackage{wrapfig}
\usepackage{enumitem}
\usepackage{amsmath}
\usepackage{amssymb}

\usepackage{cancel}

\newcommand{\crossmark}{\textit{\sffamily x}}

\newcommand{\best}[1]{{\color{red}\textbf{#1}}}
\newcommand{\good}[1]{{\color{violet}#1}}

\definecolor{darkgreen}{rgb}{0.0, 0.56, 0.0}

\newcommand{\edits}[1]{{\color{black}#1}}
\newcommand{\finaledits}[1]{{\color{black}#1}}

\usepackage{lastpage}
\jmlrheading{23}{2022}{1-\pageref{LastPage}}{5/22; Revised
11/22}{12/22}{22-0567}{Vijay Prakash Dwivedi, Chaitanya K. Joshi, Anh Tuan Luu, Thomas Laurent, Yoshua Bengio, Xavier Bresson}
\ShortHeadings{Benchmarking Graph Neural Networks}{Dwivedi, Joshi, Luu, Laurent, Bengio and Bresson}

\begin{document}

\title{Benchmarking Graph Neural Networks}

\author{\name Vijay Prakash Dwivedi$^1$ \email vijaypra001@e.ntu.edu.sg 
      \AND
      \name Chaitanya K. Joshi$^2$ \email chaitanya.joshi@cl.cam.ac.uk 
      \AND
      \name Anh Tuan Luu$^1$ \email anhtuan.luu@ntu.edu.sg
      \AND
      \name Thomas Laurent$^3$ \email tlaurent@lmu.edu 
      \AND
      \name Yoshua Bengio$^4$ \email yoshua.bengio@mila.quebec 
      \AND
      \name Xavier Bresson$^5$ \email xaviercs@nus.edu.sg \\
      \addr $^1$Nanyang Technological University, Singapore, $^2$University of Cambridge, UK, $^3$Loyola Marymount University, USA, $^4$Mila, University of Montr\'eal, Canada, $^5$National University of Singapore}

\editor{Joaquin Vanschoren}

\maketitle

\begin{abstract}

In the last few years, graph neural networks (GNNs) have become the standard toolkit for analyzing and learning from data on graphs. This emerging field has witnessed an extensive growth of promising techniques that have been applied with success to computer science, mathematics, biology, physics and chemistry. 
But for any successful field to become mainstream and reliable, benchmarks must be developed to quantify progress.
This led us in March 2020 to release a benchmark framework that i) comprises of a diverse collection of mathematical and real-world graphs, ii) enables fair model comparison with the same parameter budget to identify key architectures, iii) has an open-source, easy-to-use and reproducible code infrastructure, and iv) is flexible for researchers to experiment with new theoretical ideas. As of December 2022, the GitHub repository\footnote{The framework is hosted
at \url{https://github.com/graphdeeplearning/benchmarking-gnns}.} has reached 2,000 stars and 380 forks, which demonstrates 
the utility of the proposed open-source framework through the wide usage by the GNN community. In this paper, we present an updated version of our benchmark with a concise presentation of the aforementioned framework characteristics, an additional 
medium-sized molecular
dataset AQSOL, similar to the popular ZINC, but with a real-world measured chemical target, and discuss how this framework can be leveraged to explore new GNN designs and insights. As a proof of value of our benchmark, we study the case of graph positional encoding (PE) in GNNs, which was 
introduced with this benchmark 
and has since spurred interest of exploring more powerful PE for Transformers and GNNs in a robust experimental setting.

\end{abstract}



\section{Introduction}

Graph neural networks 
have benefitted from a great interest recently with numerous methods being developed for diverse domains including chemistry~\citep{duvenaud2015convolutional,gilmer2017neural}, physics~\citep{cranmer2019learning,sanchez2020learning}, social sciences~\citep{monti2019fake}, transportation \citep{derrow2021eta}, knowledge graphs~\citep{schlichtkrull2018modeling,chami2020low}, recommendation~\citep{monti2017geometric,ying2018graph}, and neuroscience~\citep{griffa2017transient}. Developing powerful and
expressive GNN architectures is a key concern towards practical applications and real-world adoption of graph machine learning.
However, tracking progress is often challenging in the absence of a community-standard benchmark as models that are evaluated on traditionally-used datasets with inconsistent experimental comparisons make it difficult to differentiate complex, simple and graph-agnostic architectures \citep{Hoang2019RevisitingGN,chen2019powerful, errica2019fair}.

\begin{wrapfigure}{r}{0.65\textwidth}
    \vspace{-6pt}
    \centering
    \includegraphics[width=1\linewidth]{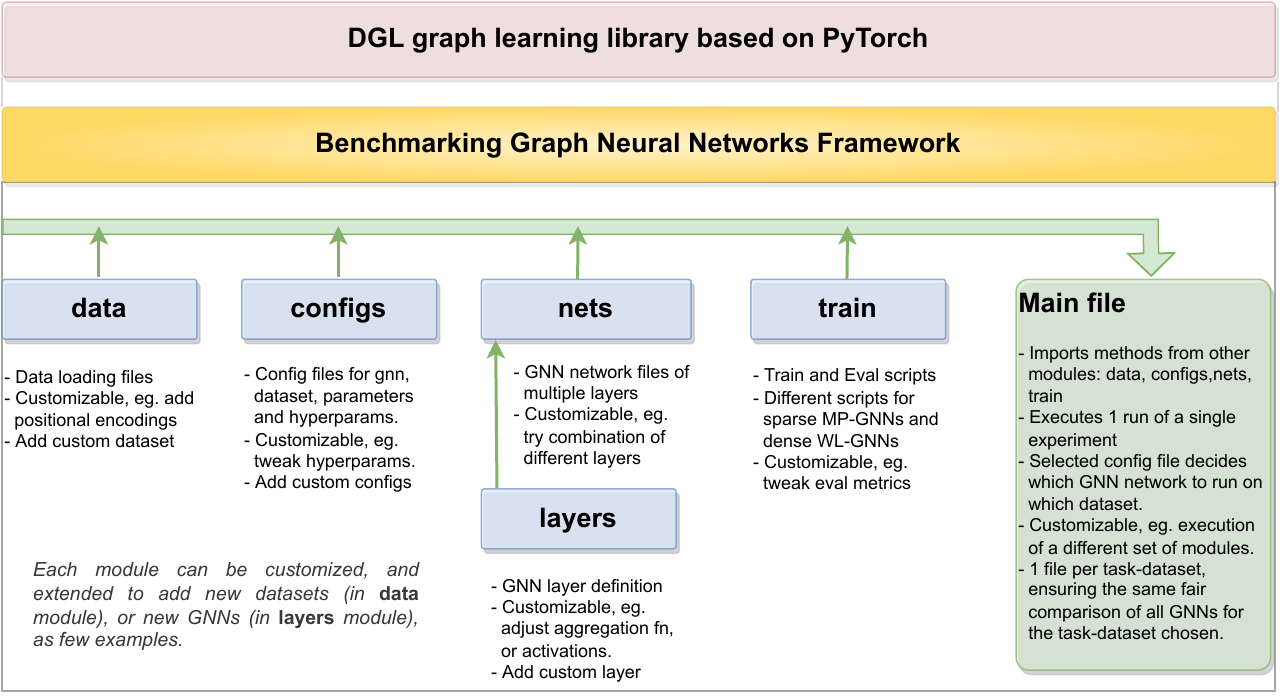}
    \vspace{-20pt}
    \caption{Overview sketch of the proposed GNN benchmarking framework with 
    different
    modular components. This benchmark is built upon DGL and PyTorch libraries.
    }
    \label{fig:framework_diagram}
\end{wrapfigure}

We introduce 
an open-source GNN benchmarking framework (see Fig \ref{fig:framework_diagram})
that brings forward a set of diverse medium-scale datasets
which are discriminative to benchmark different GNN models when compared fairly on fixed parameter budgets. The existing collection of datasets, the protocol to use the same parameter budgets for comparison, and the modular coding infrastructure has been widely used to prototype powerful GNN ideas and develop new insights, 
as shown by 2000+ stars and 380+ forks of the GitHub repository from its first release in March 2020, and 470+ citations gathered by the ArXiv technical report according to Google Scholar. Aspects of the benchmark have 
led to facilitating several interesting studies for GNNs such as on (i) the aggregation functions and filters \citep{corso2020principal, tailor2021we, elhag2022graph}, (ii) improving expressive power of GNNs \citep{valsesia2021ran, bouritsas2022improving, bevilacqua2021equivariant}, (iii) pooling mechanisms \citep{mesquita2020rethinking}, (iv) graph-specific normalization and regularization \citep{chen2022learning, zhou2020towards, zhang2021ssfg}, and (v) GNNs' robustness and efficiency \citep{wei2022evaluating, tailor2020degree}
among other ideas contributed in the literature. 
In this paper, we provide an updated overview of the proposed framework that extends on the previous collection of datasets to 
(a) include a number of essential mathematical datasets which can be used to test specific theoretical graph properties, and (b) incorporate another molecular dataset, AQSOL \citep{sorkun2019aqsoldb} that has real-world experimental solubility targets unlike ZINC's computed targets, resulting in a collection of 12 datasets (see Table \ref{tab:data_summary}).
The remainder of the paper discusses a 
proof of concept of
the benchmark that can be used to explore and develop new insights for GNNs.

\section{Overview of GNN Benchmarking Framework}
\label{sec:overview_gnn_benchmarking_framework}

\textbf{Datasets.} 
Collecting representative, realistic and medium-to-large scale graph datasets presents several challenges.
It is unclear what theoretical tools can define the quality of a dataset or validate its statistical representativeness for a given task.
Similarly, there are several arbitrary choices when preparing graphs, such as node and edge features. 
Finally, very large graph datasets also present a computational challenge and require extensive GPU resources to be studied~\citep{chiang2019cluster,rossi2020sign, hu2021ogb}. 

\begin{wrapfigure}{r}{0.48\textwidth}
\vspace{-1pt}
    \scalebox{0.55}{
    \begin{tabular}{l|c|c}
    \toprule
    \textbf{\hspace{1cm} Domain \hspace{1cm}} & \textbf{Dataset} & \textbf{Task}  \\
      \midrule
      
      \multicolumn{3}{l}{\textbf{A. \textsc{Real World Graphs}}}\\
    \midrule
      \multirow{2}{*}{Chemistry}
      & ZINC & \multirow{2}{*}{Graph Regression} \\
      & AQSOL & \\
      \cmidrule{1-3}
      \multirow{2}{*}{Social/Academic Networks}
      & OGBL-COLLAB &  Edge Classification \\
      & WikiCS &  Node Classification \\
      \cmidrule{1-3}
      \multirow{2}{*}{Computer Vision} 
      & MNIST & \multirow{2}{*}{Graph Classification} \\
      & CIFAR10 &  \\
      \midrule
      \multicolumn{3}{l}{\textbf{B. \textsc{Mathematical Graphs}}}\\
    \midrule
      \multirow{2}{*}{Mathematical Modelling} 
      & PATTERN &  \multirow{2}{*}{Node Classification} \\
      & CLUSTER &  \\
      \cmidrule{1-3}
      \multirow{1}{*}{Combinatorial Optimization}
      & \multirow{1}{*}{TSP} & \multirow{1}{*}{Edge Classification} \\
      \cmidrule{1-3}
      Isomorphism & CSL &  Graph Classification \\
      \cmidrule{1-3}
      Cycles in Graphs & CYCLES  & Graph Classification \\
      \cmidrule{1-3}
      Multi Graph Properties & GraphTheoryProp & Multi Node/Graph Task\\
      \bottomrule
    \end{tabular}
    }
    \makeatletter\def\@captype{table}\makeatother
    \caption{Summary statistics of datasets included in the 
    benchmark. Additional details 
    in Appendix Table \ref{tab:data_stats} and Sec. \ref{sec:expsetup}.
    }
    \vspace{-6pt}
    \label{tab:data_summary}
\end{wrapfigure}

On account of such challenges, we present in our benchmark a collection of 12 graph datasets, listed in Table \ref{tab:data_summary}, which are (i) collected from real-world sources and \edits{generated} from mathematical models, (ii) of medium-scale size suitable for academic research, (iii) representative of the three fundamental 
learning tasks at graph-level, node-level and edge-level, and (iv) from diverse end-application domains.
These datasets are appropriate to statistically separate the performance of GNNs on specific graph properties, hence fulfilling the academic mission to identify first principles.

\noindent\textbf{Coding Infrastructure.}
Our benchmarking infrastructure builds upon PyTorch \citep{paszke2019pytorch} and DGL \citep{wang2019dgl}, and has been developed with the following fundamental objectives:
(a) Ease-of-use and modularity, enabling new users to experiment and study the building blocks of GNNs;
(b) Experimental rigour and fairness for all models being benchmarked; and
(c) Being future-proof and comprehensive for tracking the progress of graph ML tasks and new GNNs.
At a high level as sketched in Fig \ref{fig:framework_diagram}, our benchmark unifies independent components for:
(i) Data pipelines; 
(ii) GNN layers and models; 
(iii) Training and evaluation functions;
(iv) Network and hyperparameter configurations; and 
(v) Scripts for reproducibility. 
This standardized framework has been of immense help to the community as aforementioned about its wide community usage. It has enabled researchers to explore new ideas at any stage of the pipeline without setting up everything else. 
We direct readers to the \texttt{README} user manual included in our GitHub repository for detailed instructions on using the coding infrastructure.

\noindent\textbf{Parameter Budgets for Fair Comparison.}
One goal of this benchmark is not to find the optimal hyperparameters for a specific model (which is computationally expensive), but to compare the model and their building blocks within a budget of parameters.
Therefore, we decide on using two model parameter budgets: i) 100k for each GNN for all the datasets, and ii) 500k for GNNs where the scalability of a model to larger parameters and deeper layers are investigated. 
The layers and dimensions are selected accordingly to match these budgets.

\edits{\noindent\textbf{Discussion on Design Choices.} \textbf{First}, our motivation behind the medium-scale datasets in the benchmark 
is to enable swift yet reliable prototyping of GNN research ideas as we can achieve statistical difference in GNN performance within 12 hours of single experiment runs (see Appendix \ref{sec:hardware}).}
Medium-scale datasets are arguably more informative than small datasets and more feasible than large-scale datasets in the academic-scale research.
\edits{\textbf{Second}, our coding infrastucture with standard protocols has enabled fair comparison of GNNs something that was lacking in prior literature \citep{errica2019fair}. \textbf{Third}, a fixed budget of model parameters for each GNN model allows for fair comparison of different architectures. In the absence of such design choice, it is comparatively difficult to conclude whether a better performing model's gain arises from its architectural design or extra learning capacity brought by additional model parameters. \textbf{Finally}, the aforementioned decisions can be refined and extended to allow further flexibility as elaborated in Appendix \ref{sec:elaboration_design_choices}.
}

\vspace{-4pt}
\section{How can the benchmark be used to explore new insights?}
The proposed benchmarking framework can be used to test new research ideas at the level of data preprocessing, improving the GNN layers and normalization schemes, or even to substantiate the performance of a novel GNN model. Such studies are conveniently facilitated given the set of diverse datasets and the rigorous comparison of different experiments on same parameter budgets. At any stage, a modular component of the framework, 
such as \textbf{data}, \textbf{layers}, etc., can be modified and multiple experiments on the datasets can be conducted fairly and with ease. Indeed, we employ the framework to perform multiple studies, out of which we present here the insight of positional encodings for GNNs using Laplacian eigenvectors, for an example, while the remainder is included in the appendix.\\
\vspace{-10pt}

\noindent\textbf{Graph Positional Encoding.} 
Nodes in a graph do not have any canonical positional information. In the absence of available features, nodes are anonymous, such as the nodes in CSL, CYCLES or GraphTheoryProp datasets in our benchmark. 
As such, message passing based GCNs perform either poorly or fail completely to detect the class of the graph, such as isomorphic class, or cycles \citep{murphy2019relational, Loukas2020What}. We proposed the use of Laplacian eigenvectors \citep{belkin2003laplacian} as node positional encoding 
\edits{by building on top of corresponding dataset files in the \textbf{data} module 
as shown in the pseudo-code snippet alongside.} 
\finaledits{In other words, the positional encoding $p_i$ for a node $i$ can be added to its features $x_i$ as $x_i = x_i + p_i$.}
\edits{Similarly, other ideas can be explored by leveraging respective modules of the framework (in Fig \ref{fig:framework_diagram}) for which we direct to \texttt{README} of our GitHub repository.}

\begin{wrapfigure}{r}{0.45\textwidth}
    \vspace{-23pt}
    {\color{black}
    \begin{small}
\begin{verbnobox}[{\highlight}\fontsize{8pt}{8pt}\selectfont\mbox{}]
class NameOfDataset(torch.utils.data.Dataset):
    def __init__(self, name=`name_of_dataset'):
        # existing code to load dataset  
    
    def _add_positional_encodings(self, args):
        # new code that precomputes and adds
        # positional encoding using eigenvectors
\end{verbnobox}
\end{small}}
\vspace{-17pt}
\caption{Primary code block in \textbf{data} module to implement Graph PE.}
\vspace{-10pt}
\end{wrapfigure}

We used the benchmark to validate and also quantified the improvement provided by this idea.
The Laplacian PE effectively improved the MP-GCNs (message-passing based Graph Convolutional Networks)
on the the 3 synthetics datasets mentioned previously and other real-world datasets, including the newly added AQSOL dataset.
A detailed presentation of the PE with experiments are in Appendix \ref{sec:lap_pe}.
After the introduction of Laplacian PE through this benchmark, new ideas followed up in the literature for improving PE \citep{beaini2021directional, wang2022equivariant, lim2022sign,  kreuzer2021rethinking, ying2021transformers, mialon2021graphit}, thus demonstrating how the identification of first principles using the proposed benchmark can steer GNN research.

\vspace{-4pt}
\section{Conclusion}
This paper introduces an open-source benchmarking framework for Graph Neural Networks that is modular, easy-to-use, and can be leveraged to quickly yet robustly test new GNN ideas and explore insights that direct further research.
The benchmark led us to propose graph PE 
that has remained an interesting avenue of exploration since the first release of our benchmark.
We also perform additional studies on investigation of different GNN categories, and edge representations for link prediction, the details of which are included in the appendix for interested readers.

\section*{Acknowledgments}
XB is supported by NRF Fellowship NRFF2017-10, NUS-R-252-000-B97-133 and A*STAR Grant ID A20H4g2141. This research is supported by Nanyang Technological University, under SUG Grant (020724-00001). The authors thank the reviewers and the editor for their comments and suggestions, which greatly improved the manuscript.

\newpage
\appendix

\section{Related Work}
In the last few years, graph neural networks (GNNs) have seen a great surge of interest with promising methods being developed for myriad of domains including chemistry~\citep{duvenaud2015convolutional,gilmer2017neural}, physics~\citep{cranmer2019learning,sanchez2020learning}, social sciences~\citep{kipf2017semi,monti2019fake}, knowledge graphs~\citep{schlichtkrull2018modeling,chami2020low}, recommendation~\citep{monti2017geometric,ying2018graph}, and neuroscience~\citep{griffa2017transient}. 
Historically, three classes of GNNs have been developed. The first models \citep{art:ScarselliGoriTsoiHagenbuchnerMonfardini09,bruna2013spectral,NIPS2016_6081,NIPS2016_6398,kipf2017semi,hamilton2017inductive} aimed at extending the original convolutional neural networks~\citep{lecun1995convolutional,lecun1998gradient} to graphs. 
The second class enhanced the original models with {anisotropic} operations on graphs~\citep{perona1990scale}, such as attention and gating mechanisms~\citep{battaglia2016interaction,marcheggiani2017encoding,Monti_2017,velickovic2018graph,bresson2017residual}.
The recent third class has introduced GNNs that improve upon theoretical limitations of previous models \citep{xu2018how,morris2019weisfeiler,maron2019provably,chen2019equivalence,murphy2019relational,srinivasan2019equivalence}. 
Specifically, the first two classes can only differentiate
simple non-isomorphic graphs and cannot separate automorphic nodes.
Developing powerful and theoretically expressive GNN architectures is a key concern towards practical applications and real-world adoption of graph machine learning.
However, tracking recent progress has been challenging as most models are evaluated on small  
datasets such as Cora, Citeseer and TU, which are inappropriate to differentiate complex, simple and graph-agnostic architectures \citep{Hoang2019RevisitingGN,chen2019powerful}, and do not have consensus on a unifying experimental setting~\citep{errica2019fair,hu2020ogb}.

Consequently, our motivation is to benchmark GNNs to identify and quantify what types of architectures, first principles or mechanisms are universal, generalizable, and scalable when we move to larger and more challenging datasets. Benchmarking provides a strong paradigm to answer these fundamental questions. It has proved to be beneficial for driving progress, identifying essential ideas, and solving domain-specific problems in several areas of science~\citep{weber2019essential}. 
Recently, the famous 2012 ImageNet challenge~\citep{imagenet_cvpr09} has provided a benchmark dataset that has triggered the deep learning revolution \citep{DBLP:conf/nips/KrizhevskySH12,10.1145/3065384}. Nevertheless, designing successful benchmarks is highly challenging as it requires both a coding framework with a rigorous experimental setting for fair comparisons, all while being reproducible, as well as using appropriate datasets that can statistically separate model performance. 
The lack of benchmarks has been a major issue in GNN literature as the aforementioned requirements have not been rigorously enforced.

\section{Graph Neural Network Pipeline}
\label{sec:gnns}

\begin{figure}[!h]
    \centering
    \includegraphics[width=0.99\textwidth]{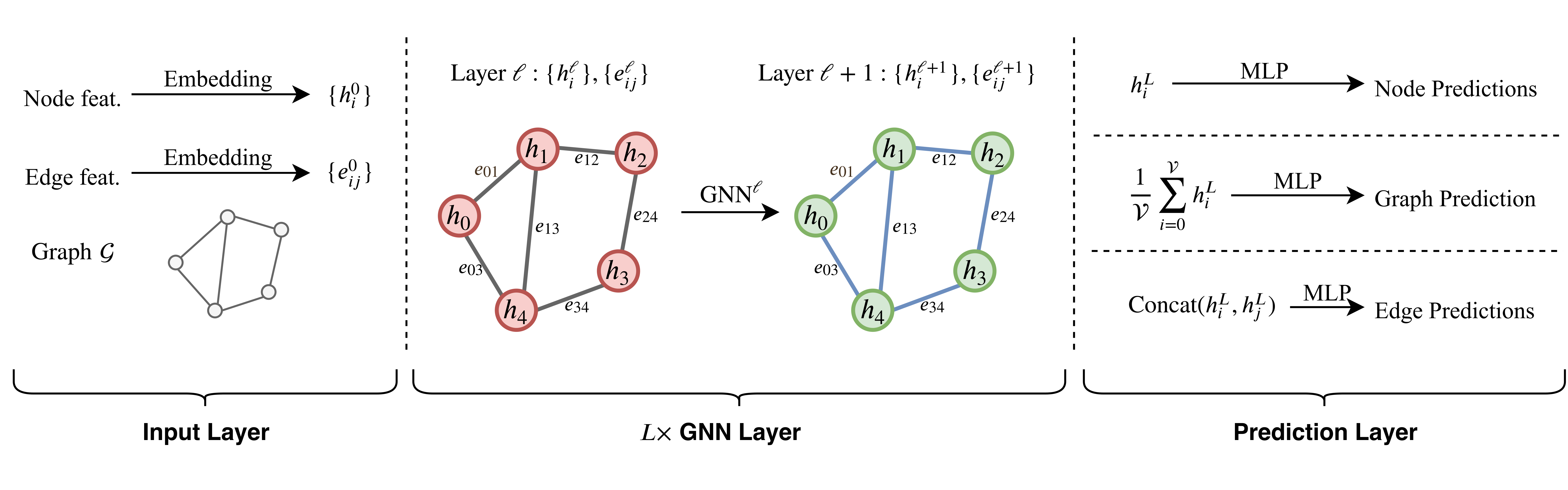}
    \caption{A standard experimental pipeline for GCNs, which embeds the graph node and edge features,
    performs several GNN layers to compute convolutional features,
    and finally makes a prediction through a task-specific MLP layer.} 
    \label{fig:pipeline_mpgcns}
\end{figure}

\begin{figure}[!t]
    \centering
    \includegraphics[width=1.02\textwidth]{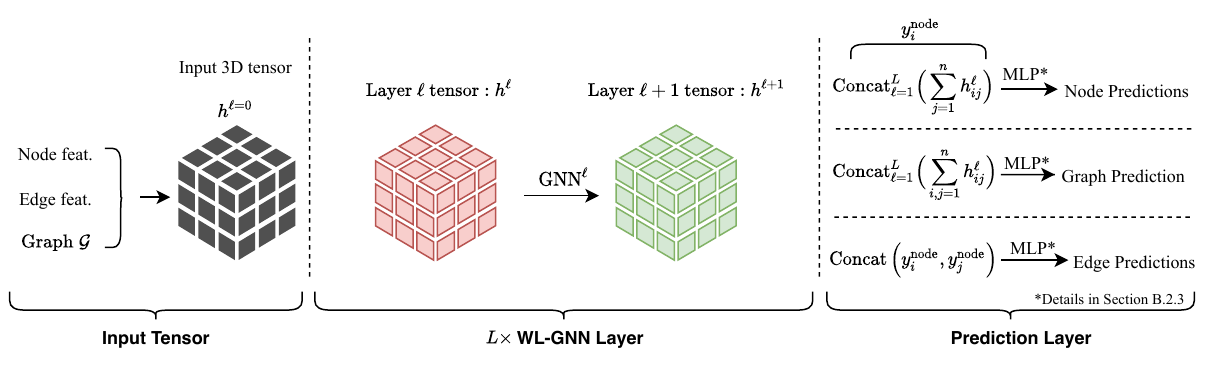}
    \caption{A standard experimental pipeline for 
    WL-GNNs, which inputs to a GNN a graph with all node and edge information (if available) represented by a dense tensor, performs several GNN layer computations over the dense tensor,
    and finally makes a prediction through a task-specific MLP layer.} 
    \label{fig:pipeline_wlgnns}
\end{figure}

In this section, we describe the experimental pipeline for the two broad classes of GNN architectures that are benchmarked in this framework as representative GNN classes -- Message Passing Graph Convolutional Networks (MP-GCNs), which are based on the message passing framework formalized in \cite{gilmer2017neural}, and Weisfeiler Lehman GNNs (WL-GNNs), which improves the theoretical limitations of MP-GCNs and align expressivity power to the WL-tests to distinguish non-isomorphic graphs. The two pipelines are illustrated in Figure \ref{fig:pipeline_mpgcns} for \textbf{GCNs} and Figure \ref{fig:pipeline_wlgnns} for \textbf{WL-GNNs}.

In Section \ref{sec:mpgcns}, we describe the components of the setup of the GCN class with \textit{vanilla} GCN \citep{kipf2017semi}, GraphSage \citep{hamilton2017inductive}, MoNet \citep{Monti_2017}, GAT \citep{velickovic2018graph}, and GatedGCN \citep{bresson2017residual}, including the input layers, the GNN layers and the task based MLP classifier layers. We also include the description of GIN \citep{xu2018how} in this section as this model can be interpreted as a GCN, although it was designed to differentiate non-isomorphic graphs.
In Section \ref{sec:wlgnns}, we present the  GNN layers and the task based MLP classifier layers 
for the class of WL-GNN models with 
Ring-GNNs \citep{chen2019equivalence} and 3WL-GNNs \citep{maron2019provably}.

\subsection{Message-Passing GCNs}
\label{sec:mpgcns}

For this class, we consider the widely used message passing-based graph convolutional networks (MP-GCNs), which update node representations from one layer to the other according to the formula: $h^{\ell+1}_{i} =f(h^{\ell}_{i},\{ h^{\ell}_{j} \}_{j\in\mathcal{N}_i})$.
Note that the update equation is {\it local}, only depending on the neighborhood $\mathcal{N}_i$ of node $i$, and {\it independent of graph size}, making the space/time complexity $O(E)$ reducing to $O(n)$ for sparse graphs.
Thus, MP-GCNs are highly parallelizable on GPUs and are implemented via sparse matrix multiplications in modern graph machine learning frameworks~\citep{wang2019dgl,fey2019fast}.
MP-GCNs draw parallels to ConvNets for computer vision \citep{lecun1998gradient} by considering a convolution operation with shared weights across the graph domain.

\subsubsection{Input Layer}

Given a graph, we are given node features $\alpha_i \in \mathbb{R}^{a \times 1}$ for each node $i$ and (optionally) edge features $\beta_{ij} \in \mathbb{R}^{b \times 1}$ for each edge connecting node $i$ and node $j$.
The input features $\alpha_i$ and $\beta_{ij}$ are embedded to $d$-dimensional hidden features $h_i^{\ell=0}$ and $e_{ij}^{\ell=0}$ via a simple linear projection before passing them to a graph neural network:
\begin{equation}
    \label{eqn:input}
    h_i^{0} = U^{0} \alpha_i + u^{0} \;\ ; \;\ e_{ij}^{0} = V^{0} \beta_{ij} + v^{0} ,
\end{equation}
where $U^{0} \in \mathbb{R}^{d \times a}$, $V^{0} \in \mathbb{R}^{d \times b}$ and $u^{0},v^{0}\in \mathbb{R}^{d}$. If the input node/edge features are one-hot vectors of discrete variables, then  biases $u^{0},v^{0}$ are not used.

\subsubsection{GCN layers}

Each GCN layer computes $d$-dimensional representations for the nodes/edges of the graph through recursive neighborhood diffusion (or message passing), where each graph node gathers features from its neighbors to represent local graph structure.
Stacking $L$ GCN layers allows the network to build node representations from the $L$-hop neighborhood of each node.

\begin{figure}[h]
    \centering
    \includegraphics[width=0.50\textwidth]{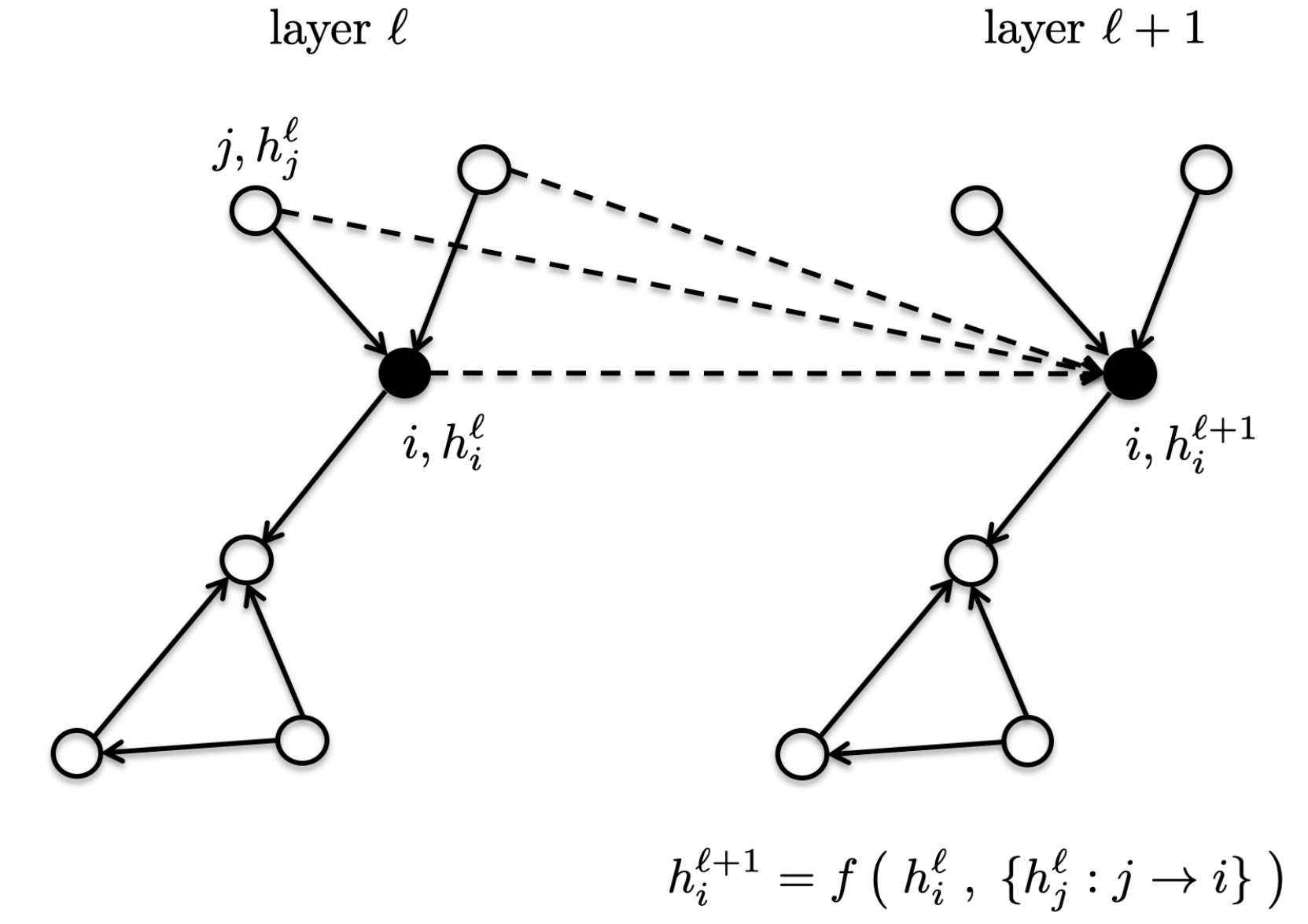}
    \caption{A generic graph neural network layer. Figure adapted from \cite{bresson2017residual}.}
    \label{fig:generic-gnn}
\end{figure}

Let $h_i^{\ell}$ denote the feature vector at layer $\ell$ associated with node $i$.
The updated features $h_i^{\ell+1}$ at the next layer $\ell+1$ are obtained by applying non-linear transformations to the central feature vector $h_i^{\ell}$ and the feature vectors $h_{j}^{\ell}$ for all nodes $j$ in the neighborhood of node $i$ (defined by the graph structure). 
This guarantees the transformation to build local reception fields, such as in standard ConvNets for computer vision, and be invariant to both graph size and vertex re-indexing.

Thus, the most generic version of a feature vector $h_i^{\ell+1}$ at vertex $i$ at the next layer in the GNN is:
\begin{equation} 
    \label{eqn:generic-gnn}
    h_{i}^{\ell+1} =  f \left( \ h_i^{\ell} \  , \ \{ h_{j}^{\ell}: j \rightarrow i \}  \ \right) ,
\end{equation}
where $\{ j \rightarrow i \}$ denotes the set of neighboring nodes $j$ pointed to node $i$, which can be replaced by $\{ j \in \mathcal{N}_i \}$, the set of neighbors of node $i$, if the graph is undirected. In other words, a GNN is defined by a mapping $f$ taking as input a vector $h_i^{\ell}$ (the feature vector of the center vertex) as well as an un-ordered set of vectors $\{ h_{j}^{\ell}\}$ (the feature vectors of all neighboring vertices), see Figure \ref{fig:generic-gnn}. 
The arbitrary choice of the mapping $f$ defines an instantiation of a class of GNNs.

\begin{figure}[t!]
\centering
\begin{minipage}{.5\textwidth}
\centering
  \vspace{2.4mm}
  \includegraphics[width=0.5\columnwidth]{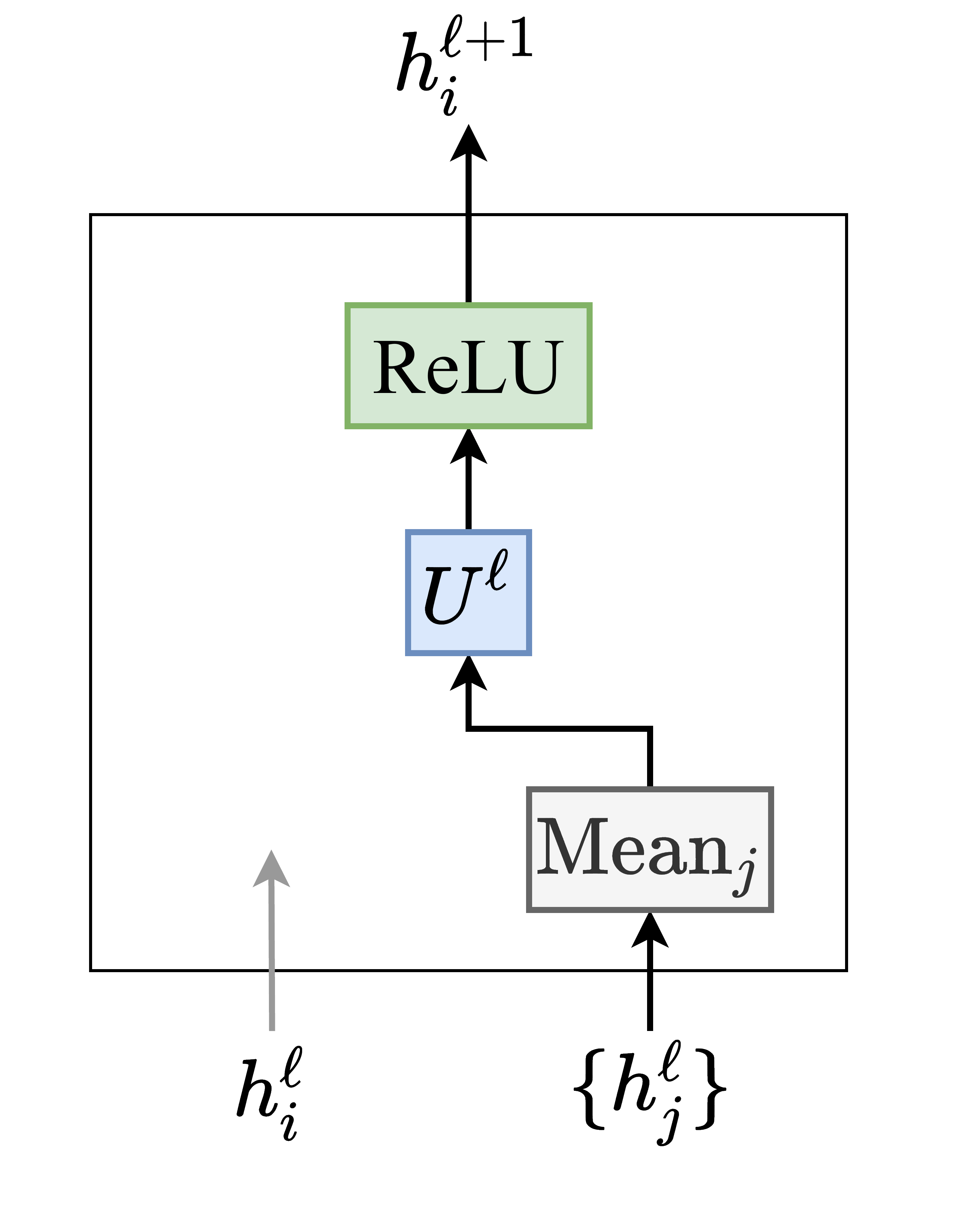}
  \vspace{2.4mm}
  \caption{GCN Layer}
  \label{fig:gcn}
\end{minipage}%
\begin{minipage}{.5\textwidth}
\centering
  \includegraphics[width=0.55\columnwidth]{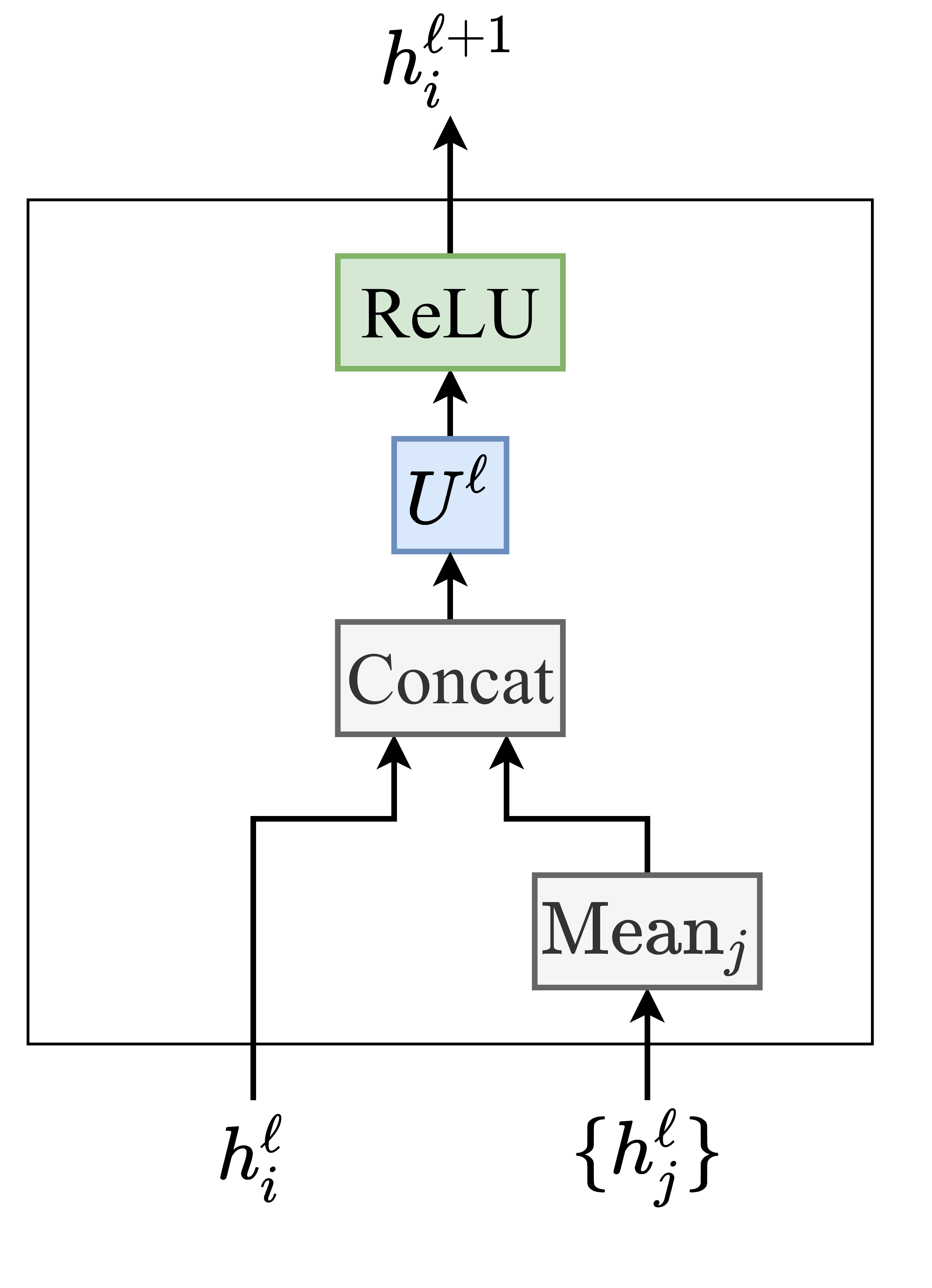}
  \caption{GraphSage Layer}
  \label{fig:graphsage}
\end{minipage}%
\end{figure}

\paragraph{Graph ConvNets (GCN) \citep{kipf2017semi}}
In the simplest formulation of GNNs, \textit{vanilla} Graph ConvNets iteratively update node features via an isotropic averaging operation over the neighborhood node features, \textit{i.e.},
\begin{eqnarray}
    h_{i}^{\ell+1} &=& \text{ReLU} \Big( U^{\ell} \   \text{Mean}_{j \in \mathcal{N}_i} \ h_j^{\ell}  \Big), \label{eqn:gcn} \\
    &=&  \text{ReLU} \Big( U^{\ell} \ \frac{1}{\text{deg}_i} \sum_{j \in \mathcal{N}_i} \  h_j^{\ell}  \Big), \label{eqn:gcn-app}
\end{eqnarray}
where $U^{\ell} \in \mathbb{R}^{d \times d}$ (a bias is also used, but omitted for clarity purpose), $\text{deg}_i$ is the in-degree of node $i$,
see Figure \ref{fig:gcn}.
Eq. \eqref{eqn:gcn} is called a \textit{convolution} as it is a linear approximation of a localized spectral convolution. Note that it is possible to add the central node features $h_{i}^{\ell}$ in the update \eqref{eqn:gcn} by using self-loops or residual connections.

The GCN model in \cite{kipf2017semi} use symmetric normalization instead of the isotropic averaging, to result in the following node update equation:

\begin{eqnarray}
    h_{i}^{\ell+1} &=&  \text{ReLU} \Big( U^{\ell} \ \frac{1}{\sqrt{\text{deg}_i}\sqrt{\text{deg}_j}} \sum_{j \in \mathcal{N}_i} \  h_j^{\ell}  \Big), \label{eqn:gcn-aniso}
\end{eqnarray}


\paragraph{GraphSage \citep{hamilton2017inductive}}

GraphSage improves upon the simple GCN model by explicitly incorporating each node's own features from the previous layer in its update equation:
\begin{equation}
    \label{eqn:graphsage-mean}
    \hat h_{i}^{\ell+1} = \text{ReLU} \Big( U^{\ell} \ \text{Concat} \left( h_{i}^{\ell} , \ \text{Mean}_{j \in \mathcal{N}_i} \ h_j^{\ell} \right) \Big),
\end{equation}
where $U^{\ell} \in \mathbb{R}^{d \times 2d}$,
see Figure \ref{fig:graphsage}. Observe that the transformation applied to the central node features $ h_{i}^{\ell}$ is different to the transformation carried out to the neighborhood features $ h_j^{\ell}$. The node features are then projected onto the $\ell_2$-unit ball before being passed to the next layer:
\begin{equation}
    \label{eqn:graphsage-l2}
    h_{i}^{\ell+1} = \frac{\hat h_{i}^{\ell+1}}{\|\hat h_{i}^{\ell+1}\|_2}.
\end{equation}
The authors also define more sophisticated neighborhood aggregation functions, such as Max-pooling or LSTM aggregators:
\begin{eqnarray}
    \label{eqn:graphsage-maxpool}
    \hat h_{i}^{\ell+1} &=& \text{ReLU} \Big( U^{\ell} \ \text{Concat} \left( h_{i}^{\ell} , \ \text{Max}_{j \in \mathcal{N}_i} \text{ReLU}\left( V^{\ell} h_j^{\ell}\right) \right) \Big), \\
    \label{eqn:graphsage-lstm}
    \hat h_{i}^{\ell+1} &=& \text{ReLU} \Big( U^{\ell} \ \text{Concat} \left( h_{i}^{\ell} , \ \text{LSTM}^\ell_{j \in \mathcal{N}_i} \left( h_j^{\ell} \right) \right) \Big),
\end{eqnarray}
where $V^{\ell} \in \mathbb{R}^{d \times d}$ and the $\text{LSTM}^\ell$ cell also uses learnable weights. In our experiments, we use the Max-pooling version of GraphSage, Eq.\eqref{eqn:graphsage-maxpool}.

\begin{figure}[t!]
\centering
\begin{minipage}{.5\textwidth}
\centering
  \includegraphics[width=0.8\columnwidth]{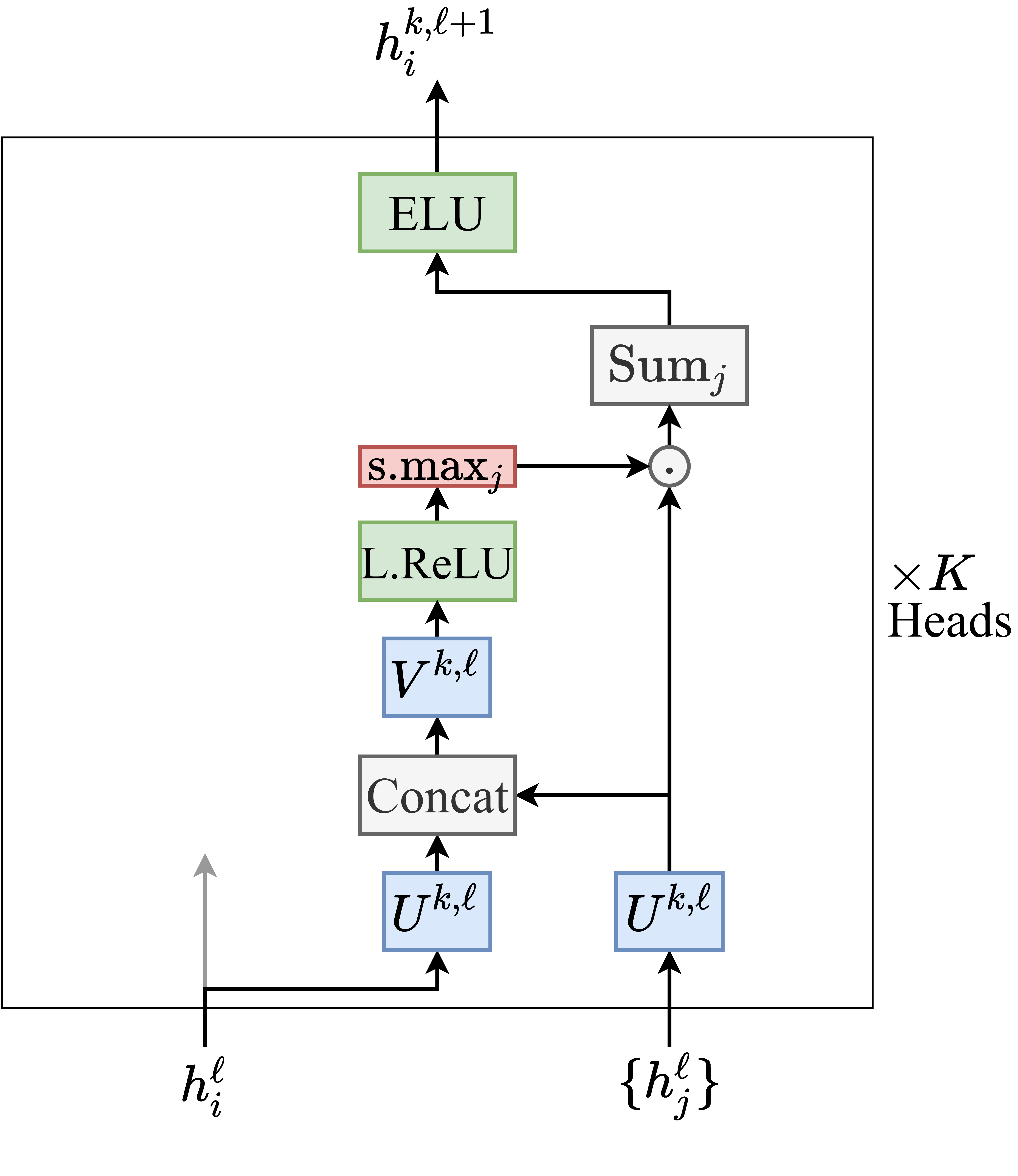}
  \caption{GAT Layer}
  \label{fig:gat}
\end{minipage}%
\begin{minipage}{.5\textwidth}
\centering
  \includegraphics[width=0.8\columnwidth]{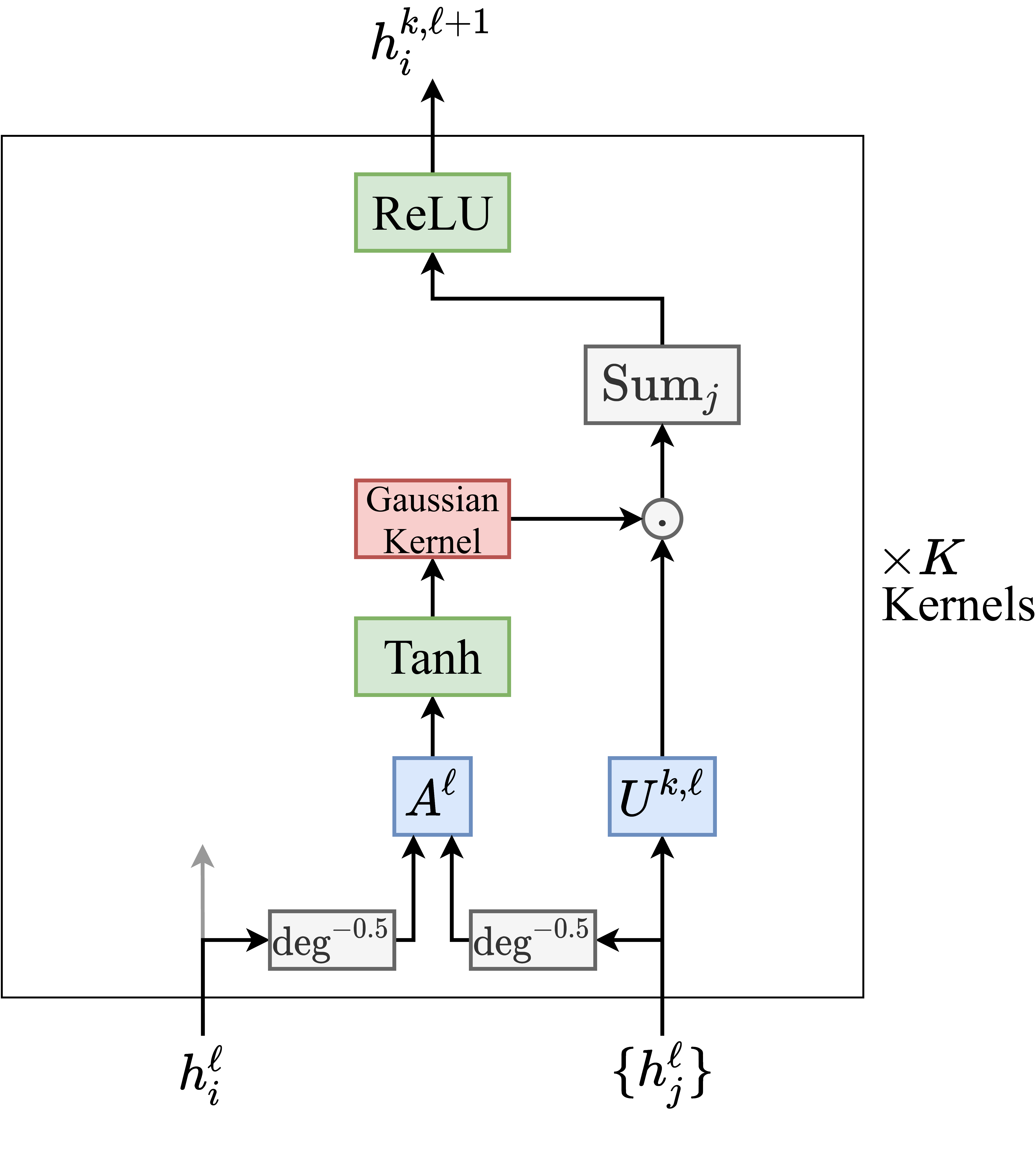}
  \caption{MoNet Layer}
  \label{fig:monet}
\end{minipage}%
\end{figure}

\paragraph{Graph Attention Network (GAT) \citep{velickovic2018graph}}
GAT uses an attention mechanism \citep{bahdanau2014neural} to introduce anisotropy in the neighborhood aggregation function. The network employs a multi-headed architecture to increase the learning capacity, similar to the Transformer \citep{vaswani2017attention, joshi2020transformers}. The node update equation is given by:
\begin{equation}
    \label{eqn:gat}
    h_{i}^{\ell+1} =  \text{Concat}_{k=1}^{K} \Big( \text{ELU} \Big( \sum_{j \in \mathcal{N}_i} e_{ij}^{k,\ell} \ U^{k,\ell} \ h_{j}^{\ell} \Big) \Big),
\end{equation}
where $U^{k,\ell} \in \mathbb{R}^{\frac{d}{K} \times d}$ are the $K$ linear projection heads, and $e_{ij}^{k,\ell}$ are the attention coefficients for each head defined as:
\begin{eqnarray}
    e_{ij}^{k,\ell} &=& \frac{\exp(\hat e_{ij}^{k,\ell})}{\sum_{j' \in \mathcal{N}_i} \exp(\hat e_{ij'}^{k,\ell}) }, \label{gat-attn} \\
     \hat e_{ij}^{k,\ell} &=& \text{LeakyReLU} \Big( V^{k,\ell} \ \text{Concat} \big( U^{k,\ell} h_{i}^{\ell} , \ U^{k,\ell} h_{j}^{\ell} \big) \Big), \label{gat-attn2}   
\end{eqnarray}

where $V^{k,\ell} \in \mathbb{R}^{\frac{2d}{K}}$, see Figure \ref{fig:gat}. GAT learns a mean over each node's neighborhood features sparsely weighted by the importance of each neighbor.

\paragraph{MoNet \citep{Monti_2017}} The MoNet model introduces a general architecture to learn on graphs and manifolds using the Bayesian Gaussian Mixture Model (GMM) \citep{dempster1977maximum}. In the case of graphs, the node update equation is defined as:
\begin{eqnarray}
    h_i^{\ell+1} &=& \text{ReLU} \Big( \sum_{k=1}^K \sum_{j \in \mathcal{N}_i}  e_{ij}^{k,\ell}\  U^{k,\ell} \ h_j^\ell  \Big), \label{eqn:monet1}\\
    e_{ij}^{k,\ell}  &=& \!\!\!\! \exp\!\Big(\!\!-\!\frac{1}{2} \! (u_{ij}^\ell -\mu_k^\ell)^T\! (\Sigma_k^\ell)^{-1} (u_{ij}^\ell -\mu_k^\ell) \! \Big), \label{eqn:monet2}\\
    u_{ij}^\ell &=& \text{Tanh}\Big(A^\ell (\text{deg}_i^{-1/2},\text{deg}_j^{-1/2})^T + a^\ell\Big), \label{eqn:monet3}
\end{eqnarray}
where $U^{k,\ell} \in \mathbb{R}^{d \times d}$, $\mu_k^\ell,{(\Sigma_k^\ell)}^{-1},a^\ell \in \mathbb{R}^{2} \text{ and } A^\ell\in \mathbb{R}^{2\times 2} $ are the (learnable) parameters of the GMM, see Figure \ref{fig:monet}.

\paragraph{Gated Graph ConvNet (GatedGCN) \citep{bresson2017residual}}

\begin{figure}[t!]
\centering
\begin{minipage}{.5\textwidth}
\centering
  \includegraphics[width=0.85\columnwidth]{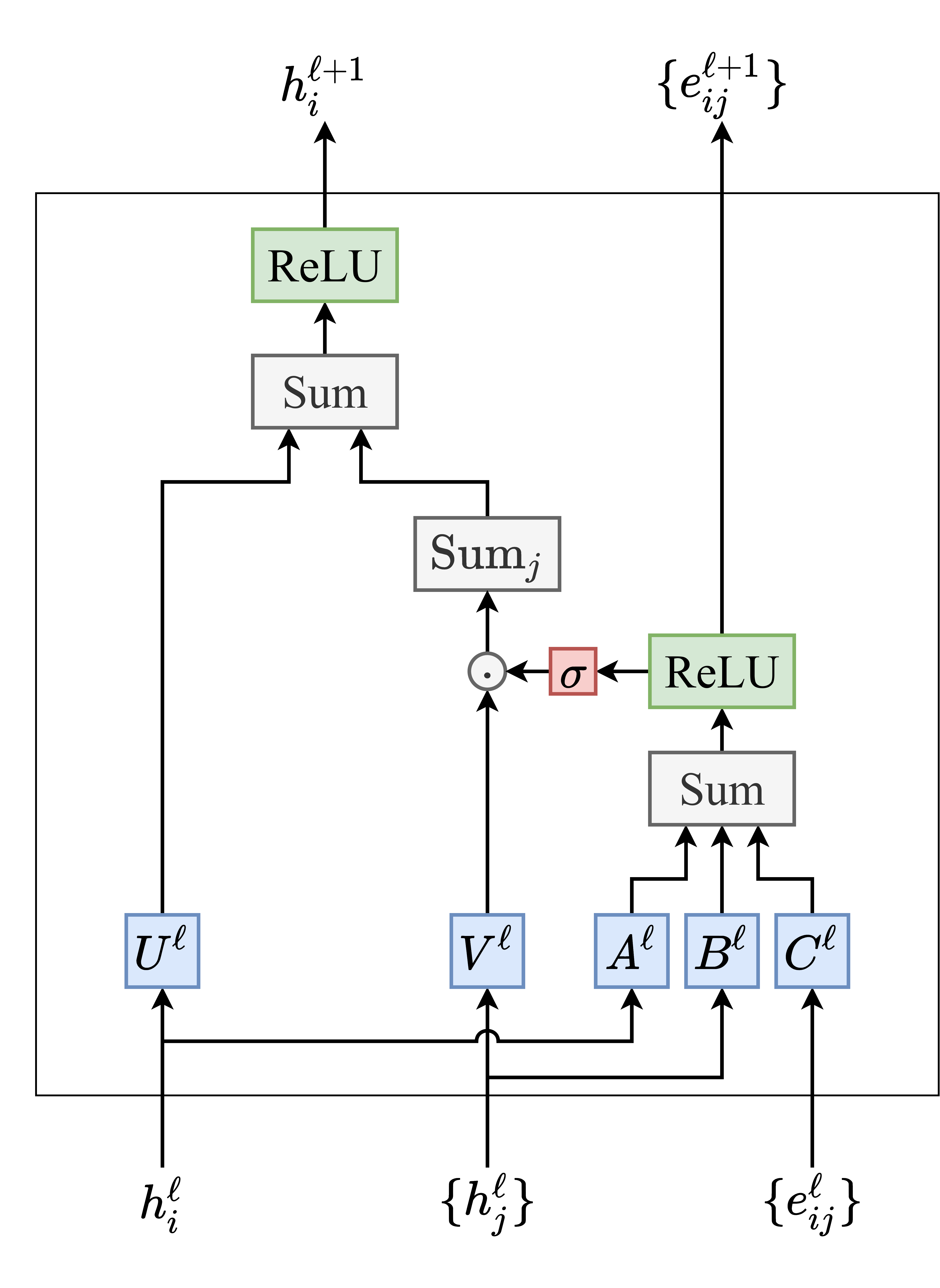}
  \caption{GatedGCN Layer}
  \label{fig:gatedgcn}
\end{minipage}%
\begin{minipage}{.5\textwidth}
\centering
  \includegraphics[width=0.55\columnwidth]{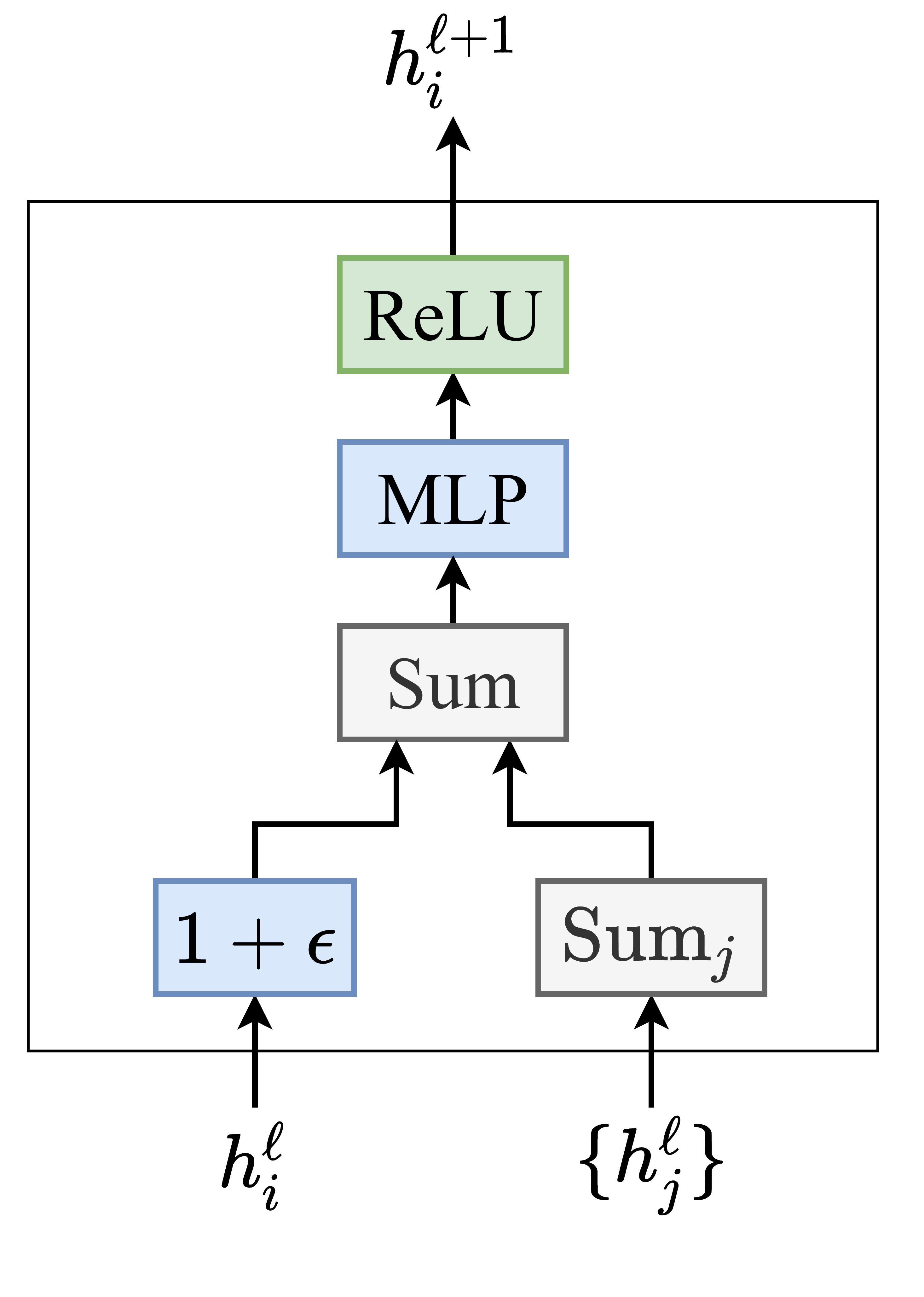}
  \caption{GIN Layer}
  \label{fig:gin}
\end{minipage}%
\end{figure}

GatedGCN considers residual connections, batch normalization and edge gates \citep{marcheggiani2017encoding} to design another anisotropic variant of GCN. The authors propose to explicitly update edge features along with node features:  
\begin{equation}
    \label{eqn:gated-gcn-h}
    h_{i}^{\ell+1} = h_{i}^{\ell} + \text{ReLU} \Big( \text{BN} \Big( U^{\ell} h_{i}^{\ell} + \sum_{j \in \mathcal{N}_i} e_{ij}^{\ell} \odot V^{\ell} h_{j}^{\ell} \Big)\Big),
\end{equation}
where $U^{\ell}, V^{\ell} \in \mathbb{R}^{d \times d}$, $\odot$ is the Hadamard product, and the edge gates $e_{ij}^{\ell}$ are defined as:
\begin{eqnarray}
    e_{ij}^{\ell} &=& \frac{\sigma(\hat e_{ij}^{\ell})}{\sum_{j' \in \mathcal{N}_i} \sigma(\hat e_{ij'}^{\ell}) + \varepsilon }, \label{eqn:gated-gcn-eta} \\
     \hat e_{ij}^{\ell} &=& \hat e_{ij}^{\ell-1} + \text{ReLU} \Big( \text{BN} \big( A^{\ell} h_{i}^{\ell-1} + B^{\ell} h_{j}^{\ell-1} + C^{\ell} \hat e_{ij}^{\ell-1} \big) \Big),  \label{eqn:gated-gcn-e}   
\end{eqnarray}
where $\sigma$ is the sigmoid function, $\varepsilon$ is a small fixed constant for numerical stability, $A^\ell, B^\ell, C^\ell \in \mathbb{R}^{d \times d}$, see Figure \ref{fig:gatedgcn}. 
Note that the edge gates \eqref{eqn:gated-gcn-eta} can be regarded as a soft attention process, related to the standard sparse attention mechanism \citep{bahdanau2014neural}.
Different from other anisotropic GNNs, the GatedGCN architecture explicitly maintains edge features $\hat e_{ij}$ at each layer, following \cite{bresson2019two,joshi2019efficient}.

\paragraph{Graph Isomorphism Networks (GIN) \citep{xu2018how}}
The GIN architecture is based the Weisfeiler-Lehman Isomorphism Test \citep{weisfeiler1968reduction} to study the expressive power of GNNs. 
The node update equation is defined as:
\begin{eqnarray}
    h_i^{\ell+1} &=& \!\!\! \text{ReLU}\big( \ U^{\ell} \big( \text{ReLU}\big( \ \text{BN}\big( \ V^{\ell} \hat{h}_i^{\ell+1} \big)\ \big)\ \big)\ \big),\label{eqn:gin2}\\
    \hat{h}_{i}^{\ell+1} &=& \left( 1 + \epsilon \right) h_{i}^{\ell} + \sum_{j \in \mathcal{N}_i} h_{j}^{\ell}, \label{eqn:gin1}
\end{eqnarray}
where $\epsilon$ is a learnable constant, $U^{\ell}, V^{\ell} \in \mathbb{R}^{d \times d}$, BN denotes Batch Normalization. See Figure \ref{fig:gin} for illustration of the update equation.

\paragraph{Normalization and Residual Connections}
As a final note, we augment each message-passing GCN layer with batch normalization (BN) \citep{ioffe2015batch} and residual connections \citep{He_2016_CVPR}. 
As such, we consider a more specific class of GCNs than \eqref{eqn:generic-gnn}:
\begin{eqnarray}
    \label{eqn:generic-gnn-with-tricks-1}
    h_{i}^{\ell+1} &=& h_{i}^{\ell} \ + \ \sigma \Big( \text{BN} \left(  {\hat{h}}_{i}^{\ell+1}  \right) \Big), \\
    \label{eqn:generic-gnn-with-tricks-2}
    {\hat{h}}_{i}^{\ell+1} &=& g_{\text{GCN}} \left( \ h_i^{\ell} \  , \ \{ h_{j}^{\ell}: j \rightarrow i \} \ \right),
\end{eqnarray}
where $\sigma$ is a non-linear activation function and $g_{\text{GCN}}$ is a specific message-passing GCN layer.

\subsubsection{Task-based Layer}
\label{sec:mpgcns_task_network}
The final component of each network is a prediction layer to compute task-dependent outputs, which are given to a loss function to train the network parameters in an end-to-end manner. The input of the prediction layer is the result of the final message-passing GCN layer for each node of the graph (except GIN, which uses features from all intermediate layers).

\paragraph{Graph classifier layer}
To perform graph classification, we first build a $d$-dimensional graph-level vector representation $y_\mathcal{G}$ by averaging over all node features in the final GCN layer:
\begin{equation}
    \label{eqn:graph-repres}
    y_\mathcal{G} =  \frac{1}{\mathcal{V}} \sum_{i=0}^{\mathcal{V}}{h_{i}^{L}},
\end{equation}
The graph features are then passed to a MLP, which outputs un-normalized logits/scores $y_\text{pred}\in \mathbb{R}^C$ for each class:
\begin{equation}
    \label{eqn:graph-clf}
    y_\text{pred} = P \ \text{ReLU} \left( Q \ y_\mathcal{G} \right),
\end{equation}
where $P \in \mathbb{R}^{d \times C}, Q \in \mathbb{R}^{d \times d}, C$ is the number of classes. 
Finally, we minimize the cross-entropy loss between the logits and groundtruth labels.

\paragraph{Graph regression layer}
For graph regression, we compute $y_\mathcal{G}$ using Eq.\eqref{eqn:graph-repres} and pass it to a MLP which gives the prediction score $y_\text{pred} \in \mathbb{R}$:
\begin{equation}
    \label{eqn:graph-regress}
    y_\text{pred} = P \ \text{ReLU} \left( Q \ y_\mathcal{G} \right),
\end{equation}
where $P \in \mathbb{R}^{d \times 1}, Q \in \mathbb{R}^{d \times d}$.
The L1-loss between the predicted score and the groundtruth score is minimized during the training.

\paragraph{Node classifier layer}
For node classification, we independently pass each node's feature vector to a MLP for computing the un-normalized logits $y_{i,\text{pred}}\in \mathbb{R}^C$ for each class:
\begin{equation}
    \label{eqn:node-clf}
    y_{i,\text{pred}} = P \ \text{ReLU} \left( Q \ h_{i}^{L} \right),
\end{equation}
where $P \in \mathbb{R}^{d \times C}, Q \in \mathbb{R}^{d \times d}$. The cross-entropy loss weighted inversely by the class size is used during training.

\paragraph{Edge classifier layer}
To make a prediction for each graph edge $e_{ij}$, we first concatenate node features $h_i$ and $h_j$ from the final GNN layer. 
The concatenated edge features are then passed to a MLP for computing the un-normalized logits $y_{ij,\text{pred}}\in \mathbb{R}^C$ for each class:
\begin{equation}
    \label{eqn:edge-clf}
    y_{ij,\text{pred}} = P \ \text{ReLU} \left( Q \ \text{Concat} \left( h_{i}^{L}  , \ h_{j}^{L} \right) \right),
\end{equation}
where $P \in \mathbb{R}^{d \times C}, Q \in \mathbb{R}^{d \times 2d}$. The standard cross-entropy loss between the logits and groundtruth labels is used.

\subsection{Weisfeiler-Lehman GNNs}
\label{sec:wlgnns}

Weisfeiler-Lehman GNNs are the second GNN class we include in our benchmarking framework which are based on the WL test \citep{weisfeiler1968reduction}. \cite{xu2018how} introduced GIN--Graph Isomorphism Network, a provable $1$-WL GNN, which can distinguish two non-isomorphic graphs w.r.t. the $1$-WL test.
Higher $k$-WL isomorphic tests lead to more discriminative $k$-WL GNNs in \citep{morris2019weisfeiler,maron2019provably}. However, $k$-WL GNNs require the use of tensors of rank $k$, which is intractable in practice for $k>2$. As a result, \cite{maron2019provably} proposed a model, namely $3$-WL GNNs, that uses rank-2 tensors while being $3$-WL provable. This $3$-WL model improves the space/time complexities of \cite{morris2019weisfeiler} from $O(n^3)/O(n^4)$ to $O(n^2)/O(n^3)$ respectively. We use 3WLGNNs \citep{maron2019provably} and RingGNNs \citep{chen2019equivalence} as the GNN instances in this class, the experimental pipeline of which are described as follows.

\subsubsection{Input Tensor}

For a given graph with adjacency matrix $A \in \mathbb{R}^{n \times n}$, node features $h^\textrm{node} \in \mathbb{R}^{n\times d}$ and edge features $h^\textrm{edge} \in \mathbb{R}^{n\times n\times d_e}$, the input tensor to the RingGNN and 3WL-GNN networks is defined as 
\begin{eqnarray}
    &&h^{\ell=0} \in \mathbb{R}^{n \times n \times (1 + d + d_e)},  
\end{eqnarray}
where
\begin{eqnarray}
    &&h^{\ell=0}_{i,j,1} = A_{ij} \in \mathbb{R}, \quad \forall i,j \\
    &&h^{\ell=0}_{i,j,2:d+1} = 
    \left\{
\begin{array}{llll}
&h^\textrm{node}_i \in \mathbb{R}^{d}&\!\!\!\!\!\!, &\quad \forall i=j\\
&0&\!\!\!\!\!\!,  &\quad \forall i\not =j\\
\end{array}
\right. \\
    &&h^{\ell=0}_{i,j,d+2:d+d_e+1} = h_{ij}^\textrm{edge} \in \mathbb{R}^{d_e}
\end{eqnarray}

\subsubsection{WL-GNN layers}

\begin{figure}[t!]
\centering
\begin{minipage}{.5\textwidth}
\centering
  \includegraphics[width=0.65\columnwidth]{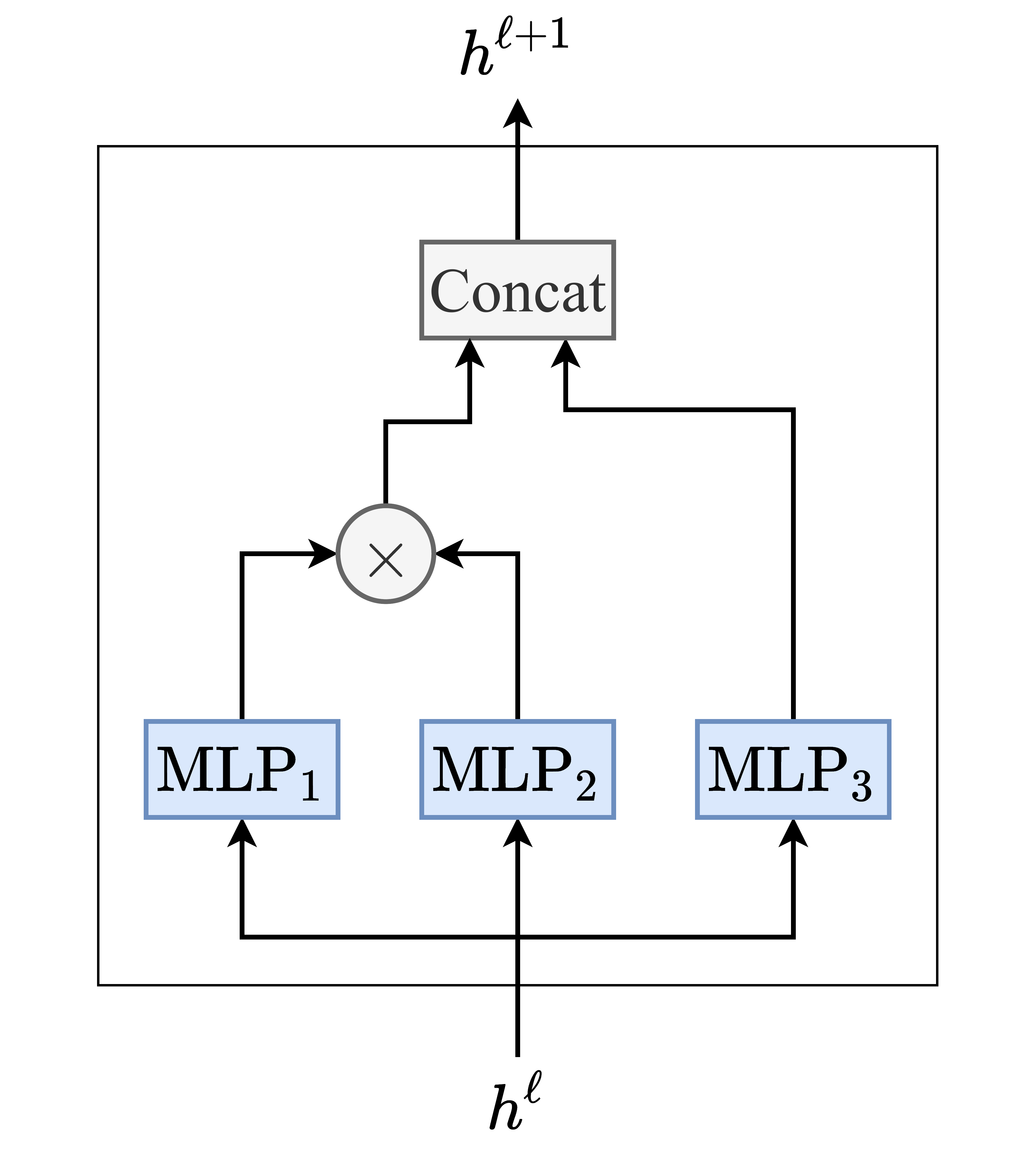}
  \caption{3WL-GNN Layer}
  \label{fig:3wlgnn}
\end{minipage}%
\begin{minipage}{.5\textwidth}
\centering
  \includegraphics[width=0.6\columnwidth]{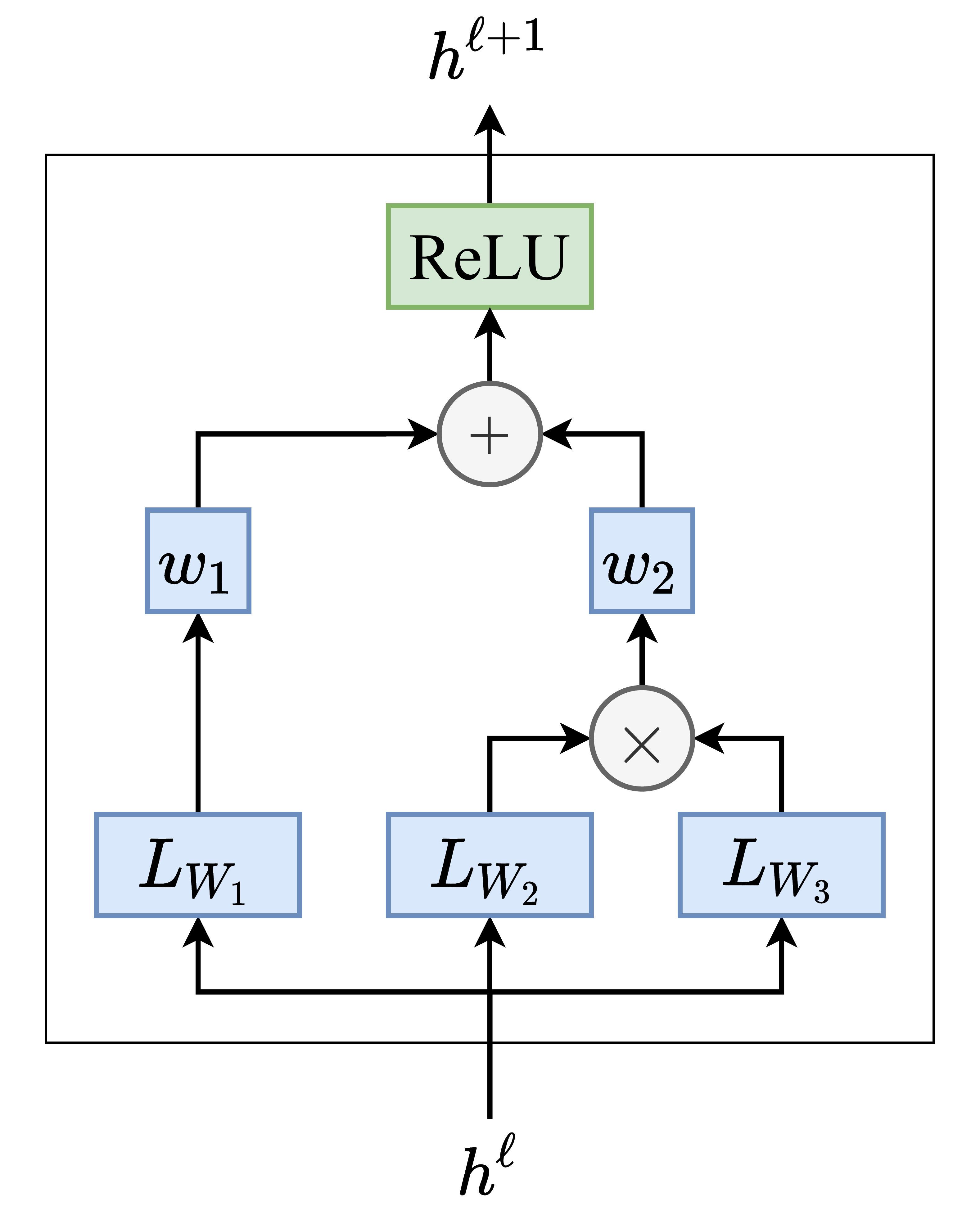}
  \caption{RingGNN Layer}
  \label{fig:ringgnn}
\end{minipage}%
\end{figure}

\paragraph{3WL-GNNs \citep{maron2019provably}}
These networks introduced an architecture that can distinguish two non-isomorphic graphs with the 3-WL test. The layer update equation of 3WL-GNNs is defined as: 
\begin{equation}
    \label{eqn:3wlgnn}
    h^{\ell+1}=\textrm{Concat}\Big( M_{W_1^\ell} (h^{\ell}) \ . \ M_{W_2^\ell} (h^{\ell}), \ M_{W_3^\ell} (h^{\ell}) \Big),
\end{equation}
where $h^{\ell}, h^{\ell+1}  \in \mathbb{R}^{n\times n\times d}$,  and $M_{W}$ are 2-layer MLPs applied along the feature dimension:
\begin{equation}
    \label{eqn:mw}
    M_{W=\{W_a,W_b\}}(h)=\sigma\Big( h \ W_a  \Big) W_b,
\end{equation}
where $W_a,W_b\in\mathbb{R}^{d\times d}$. As $h\in\mathbb{R}^{n\times n\times d}$, the MLP \eqref{eqn:mw} is implemented with a standard 2D-convolutional layer with $1\times 1$ kernel size. Eventually, the matrix multiplication in \eqref{eqn:3wlgnn} is carried out along the first and second dimensions such that:
\begin{equation}
    \label{eqn:matmul}
 \Big( M_{W_1} (h) \ . \ M_{W_2} (h) \Big)_{i,j,k} = \sum_{p=1}^n \Big( M_{W_1} (h) \Big)_{i,p,k} \ . \ \Big( M_{W_2} (h) \Big)_{p,j,k},
\end{equation}
with complexity $O(n^3)$.

\paragraph{Ring-GNNs \citep{chen2019equivalence}}
These models proposed to improve the order-2 equivariant GNNs of \cite{maron2018invariant} with the multiplication of two equivariant linear layers. The layer update equation of Ring-GNNs is designed as: 
\begin{equation}
    \label{eqn:ring-gnn}
    h^{\ell+1}=\sigma\Big(w_1^\ell \ L_{W_1^\ell} (h^\ell) + w_2^\ell \ L_{W_2^\ell} (h^\ell) . L_{W_3^\ell} (h^\ell) \Big),
\end{equation}
where $h^{\ell}, h^{\ell+1}  \in \mathbb{R}^{n\times n\times d}$, $w_{1,2}^\ell \in \mathbb{R}$, and $L_{W}$ are the equivariant linear layers defined as 
\begin{equation}
    \label{eqn:equilayer}
   \Big(  L_W(h) \Big)_{i,j,k}=\sum_{p=1}^{17}\sum_{q=1}^d W_{p,q,k} \Big( L_i(h) \Big)_{i,j,q},
\end{equation}
where $W \in \mathbb{R}^{d\times d\times 17}$ and $\{L_i\}_{i=1}^{15}$ is the set of all basis functions for all linear equivariant functions from $\mathbb{R}^{n\times n}\rightarrow\mathbb{R}^{n\times n}$ (see Appendix A in \cite{maron2018invariant} for the complete list of these 15 operations) and $\{L_i\}_{i=16}^{17}$ are the basis for the bias terms. Matrix multiplication in \eqref{eqn:ring-gnn} also implies a time complexity $O(n^3)$.

\subsubsection{Task-based network layers}
\label{sec:wlgnns_task_network}

We describe the final network layers depending on the task at hand. The loss functions corresponding to the task are the same as the GCNs, and presented in Section \ref{sec:mpgcns_task_network}.

\paragraph{Graph classifier layer} 
We have followed the original author implementations in \cite{maron2019provably,maron2018invariant,chen2019equivalence} to design the classifier layer for 3WL-GNNs and Ring-GNNs. Similar to \cite{xu2018representation,xu2018how}, the classifier layer for Ring-GNNs uses features from all intermediate layers and then passes the features to a MLP:
\begin{eqnarray}
    \label{eqn:wlgnn_classif}
    y_\mathcal{G} &=& \textrm{Concat}_{\ell=1}^L \Big( \sum_{i,j=1}^{n} h^\ell_{ij} \Big) \ \in \mathbb{R}^{Ld}, \\
    y_\text{pred} &=& P \ \text{ReLU} \left( Q \ y_\mathcal{G} \right) \ \in \mathbb{R}^{C}, 
\end{eqnarray}
where $P \in \mathbb{R}^{d \times C}, Q \in \mathbb{R}^{Ld \times d}, C$ is the number of classes.

For 3WL-GNNs, Eqn. \eqref{eqn:wlgnn_classif} is replaced by a diagonal and off-diagonal max pooling readout \cite{maron2019provably, maron2018invariant} at every layer:
\begin{eqnarray}
    \label{eqn:diag_offdiag_maxpool}
    y_\mathcal{G}^{\ell} &=& \textrm{Concat} \Big( \max_{i}\ h^{\ell}_{ii} \ , \ \max_{i\not=j}\ h^{\ell}_{ij} \Big) \ \in \mathbb{R}^{2d}, 
\end{eqnarray}
and the final prediction score is defined as:
\begin{eqnarray}
    \label{eqn:3wlgnn_out}
    y_{\text{pred}} &=& \sum_{\ell=1}^{L} P^{\ell}y_{\mathcal{G}}^{\ell} \ \in \mathbb{R}^{C},
\end{eqnarray}
where $P^{\ell} \in \mathbb{R}^{2d \times C}$, $C$ is the number of classes. 

\paragraph{Graph regression layer} 
Similar to the graph classifier layer with $P \in \mathbb{R}^{d \times 1}$ for Ring-GNNs, and $P^{\ell} \ \in \mathbb{R}^{2d \times 1}$ for 3WL-GNNs.

\paragraph{Node classifier layer}
For node classification, the prediction in Ring-GNNs is done as follows:
\begin{eqnarray}
    \label{eqn:wlgnn_node}
    y_i^\textrm{node} &=& \textrm{Concat}_{\ell=1}^L \Big( \sum_{j=1}^{n} h^\ell_{ij} \Big) \ \in \mathbb{R}^{Ld}, \\
    y_{i,\text{pred}} &=& P \ \text{ReLU} \left( Q \ y_i^\textrm{node} \right) \ \in \mathbb{R}^{C},
\end{eqnarray}
where $P \in \mathbb{R}^{d \times C}, Q \in \mathbb{R}^{Ld \times d}, C$ is the number of classes.

In 3WL-GNNs, the final prediction score is defined as:
\begin{eqnarray}
    \label{eqn:wlgnn_node_3wlgnn}
    y_i^{\textrm{node}, \ell} &=&  \sum_{j=1}^{n} h^\ell_{ij} \ \in \mathbb{R}^{d}, \\
    y_{i,\text{pred}} &=& \sum_{\ell=1}^{L}  P^{\ell} \ y_i^{\textrm{node}, \ell} \ \in \mathbb{R}^{C},
\end{eqnarray}
where $P^{\ell} \in \mathbb{R}^{d \times C}, C$ is the number of classes. 

\paragraph{Edge classifier layer}
For link prediction, for both Ring-GNNs and 3WL-GNNs, the edge features are obtained by concatenating the node features such as:
\begin{eqnarray}
    \label{eqn:wlgnn_edge}
    y_i^\textrm{node} &=& \textrm{Concat}_{\ell=1}^L \Big( \sum_{j=1}^{n} h^\ell_{ij} \Big) \ \in \mathbb{R}^{Ld}, \\
    y_{ij,\text{pred}} &=& P \ \text{ReLU} \left( Q \ \text{Concat} \left( y_i^\textrm{node},y_j^\textrm{node}  \right) \right) \ \in \mathbb{R}^{C},
\end{eqnarray}
where $P \in \mathbb{R}^{d \times C}, Q \in \mathbb{R}^{2Ld \times d}, C$ is the number of classes.

\section{Datasets and Benchmarking Experiments}
\label{sec:expsetup}

\begin{table}[!t]
  \begin{center}
    \scalebox{0.72}{
    \begin{tabular}{lccrrcc}
    \toprule
    \textbf{Dataset} & \textbf{\#Graphs} & \textbf{\#Classes} & \textbf{Avg. Nodes} & \textbf{Avg. Edges} & \textbf{Node feat. (dim)} & \textbf{Edge feat. (dim)}\\
    \midrule
    ZINC & 12000 & -- & 23.16 & 49.83 & Atom Type (28) & Bond Type (4)\\
    AQSOL & 9823 & -- & 17.57 & 35.76 & Atom Type (65) & Bond Type (5)\\
    OGBL-COLLAB & 1 & -- & 235868.00 & 2358104.00 & Word Embs (128) & Year \& Weight (2) \\
    WikiCS & 1 & 10 & 11701.00 & 216123.00 & Word Embs (300) & N.A.\\
    MNIST & 70000 & 10 & 70.57 & 564.53 & Pixel+Coord (3) & Node Dist (1)\\
    CIFAR10 & 60000 & 10 & 117.63 & 941.07 & Pixel[RGB]+Coord (5) & Node Dist (1)\\
    \midrule
    PATTERN & 14000 & 2 & 117.47 & 4749.15 & Node Attr (3) & N.A.\\
    CLUSTER & 12000 & 6 & 117.20 & 4301.72 & Node Attr (7) & N.A.\\
    TSP & 12000 & 2 & 275.76 & 6894.04 & Coord (2) & Node Dist (1) \\
    CSL & 150 & 10 & 41.00 & 164.00 & N.A. & N.A.\\
    CYCLES & 20000 & 2 & 48.96 & 87.84 & N.A. & N.A.\\
    GraphTheoryProp & 7040 & -- & 18.82 & 95.17 & Source S.P.+Random(2)  & N.A.\\
    \bottomrule
    \end{tabular}
    }
    \caption{Summary statistics of all datasets. 
    Numbers in parentheses of Node features and Edge features are the dimensions.
    S.P. denotes shortest path.
    }
    \label{tab:data_stats}
  \end{center}
\end{table}

In this section, we provide details on the datasets included in the benchmarking framework (Table \ref{tab:data_summary}) and the numerical results of the experiments using the GNN described in Section \ref{sec:gnns}, which also consists experiments from a simple graph-agnostic baseline for every dataset that parallelly applies an MLP on each node's feature vector, independent of other nodes. For complete statistics of the data, see Table \ref{tab:data_stats}. The experimental overview in terms of the training strategy, reporting of results and the parameter budget used for fair comparison are described first, as follows.

{\bf Training.} We use the Adam optimizer \citep{kingma2014adam} with the same learning rate decay strategy for all models. An initial learning rate is selected in 
$\{10^{-2},10^{-3},10^{-4}\}$ 
which is reduced by half if the validation loss does not improve after a fixed number of epochs, in the range 5-25. We do not set a maximum number of epochs -- the training is stopped either when the learning rate has reached the small value of $10^{-6}$,
or the computational time reaches 12 hours. We run each experiment with 4 different seeds and report the statistics of the 4 results. More details are provided in each experimental sub-sections.

{\bf Task-based network layer.} The node representations generated by the final layer of GCNs, or the dense tensor obtained at the final layer of the higher order WL-GNNs, are passed to a network suffix which is usually a downstream MLP
of 3 layers. 
For GIN, RingGNN, and 3WL-GNN, we follow the original instructions of network suffixes to consider feature outputs from each layer of the network, similar to that of Jumping Knowledge Networks \citep{xu2018representation}. 
Refer to the equations in the Sections \ref{sec:mpgcns_task_network} and \ref{sec:wlgnns_task_network} for more details.

{\bf Parameter budgets.} Our goal is not to find the optimal set of hyperparameters for a specific GNN model (which is computationally expensive), but to compare and benchmark the model and/or their building blocks within a budget of parameters.
Therefore, we decide on using two parameter budgets: (1)~100k parameters for each GNNs for all the tasks, and (2)~500k parameters for GNNs for which we investigate scaling a model to larger parameters and deeper layers. 
The number of hidden layers and hidden dimensions are selected accordingly to match these budgets. The configuration details of each single experiment can be found in our modular coding infrastructure on GitHub. 

\subsection{Graph Regression with ZINC dataset}
\label{sec:zinc_data_exp}
For the ZINC dataset in our benchmark, we use a subset (12K) of ZINC molecular graphs (250K) dataset \citep{irwin2012zinc} to regress a molecular property known as the constrained solubility which is the term $\text{log}P-\text{SA}-\text{cycle}$ (octanol-water partition coefficients, $\text{log}P$, penalized by the synthetic accessibility score, $\text{SA}$, and number of long cycles, $\text{cycle}$). 
For each molecular graph, the node features are the types of heavy atoms and the edge features are the types of bonds between them. ZINC has been used popularly for research related to molecular graph generation \citep{jin2018junction, bresson2019two}.

\noindent {\bf Splitting.} ZINC has $10,000$ train, $1,000$ validation and $1,000$ test graphs.\\
{\bf Training.} For the learning rate strategy across all GNNs, an initial learning rate is set to $1\times10^{-3}$, the reduce factor is $0.5$, and the stopping learning rate is $1\times10^{-5}$. The patience value is $5$ for 3WLGNN and RingGNN, and $10$ for all other GNNs.\\
{\bf Performance Measure.} The performance measure is the mean absolute error (MAE) between the predicted and the groundtruth constrained solubility for each molecular graph.
{\bf Results.} The numerical results are presented in Table \ref{tab:results_zinc} and analysed in Section \ref{sec:eval} collectively with other benchmarking results.

\begin{table}[t!]
    \centering
    \scalebox{0.78}{
    \begin{tabular}{rr|ccccc}
        \toprule
        & & \multicolumn{5}{c}{\textbf{ZINC}} \\
        \textbf{Model} & \textbf{$L$} & \textbf{\#Param} & \textbf{Test MAE$\pm$s.d.} & \textbf{Train MAE$\pm$s.d.} & \textbf{\#Epoch} & \textbf{Epoch/Total} \\
        \cline{1-7}
        \cline{1-7}
        MLP & 4 & 108975 & 0.706$\pm$0.006 & 0.644$\pm$0.005 & 116.75 & 1.01s/0.03hr \\
        \cline{1-7}
        \textit{vanilla} GCN & 4 & 103077 & 0.459$\pm$0.006 & 0.343$\pm$0.011 & 196.25 & 2.89s/0.16hr \\
         & 16 & 505079 & 0.367$\pm$0.011 & 0.128$\pm$0.019 & 197.00 & 12.78s/0.71hr\\
        GraphSage & 4 & 94977 & 0.468$\pm$0.003 & 0.251$\pm$0.004 & 147.25 & 3.74s/0.15hr \\
         & 16 & 505341 & 0.398$\pm$0.002 & 0.081$\pm$0.009 & 145.50 & 16.61s/0.68hr \\
        \cline{1-7}
        GCN & 4 & 103077 & 0.416$\pm$0.006 & 0.313$\pm$0.011 & 159.50 & 1.53s/0.07hr\\
        & 16 & 505079 & \good{0.278$\pm$0.003} & 0.101$\pm$0.011 & 159.25 & 3.66s/0.16hr\\
        MoNet & 4 & 106002 & 0.397$\pm$0.010 & 0.318$\pm$0.016 & 188.25 & 1.97s/0.10hr \\
        & 16 & 504013 & 0.292$\pm$0.006 & 0.093$\pm$0.014 & 171.75 & 10.82s/0.52hr\\
        GAT & 4 & 102385 & 0.475$\pm$0.007 & 0.317$\pm$0.006 & 137.50 & 2.93s/0.11hr \\
         & 16 & 531345 & 0.384$\pm$0.007 & 0.067$\pm$0.004 & 144.00 & 12.98s/0.53hr \\
        GatedGCN & 4 & 105735 & 0.435$\pm$0.011 & 0.287$\pm$0.014 & 173.50 & 5.76s/0.28hr \\
        GatedGCN-E & 4 & 105875 & 0.375$\pm$0.003 & 0.236$\pm$0.007 & 194.75 & 5.37s/0.29hr \\
         & 16 & 504309 & \good{0.282$\pm$0.015} & 0.074$\pm$0.016 & 166.75 & 20.50s/0.96hr \\
        GatedGCN-E-PE & 16 & 505011 & \best{0.214$\pm$0.013} & 0.067$\pm$0.019 & 185.00 & 10.70s/0.56hr \\
        \cline{1-7}
        GIN & 4 & 103079 & 0.387$\pm$0.015 & 0.319$\pm$0.015 & 153.25 & 2.29s/0.10hr \\
         & 16 & 509549 & 0.526$\pm$0.051 & 0.444$\pm$0.039 & 147.00 & 10.22s/0.42hr \\
        RingGNN & 2 & 97978 & 0.512$\pm$0.023 & 0.383$\pm$0.020 & 90.25 & 327.65s/8.32hr  \\
        RingGNN-E & 2 & 104403 & 0.363$\pm$0.026 & 0.243$\pm$0.025 & 95.00 & 366.29s/9.76hr \\
         & 2 & 527283 & 0.353$\pm$0.019 & 0.236$\pm$0.019 & 79.75 & 293.94s/6.63hr \\
          & 8 & 510305 & Diverged & Diverged & Diverged & Diverged \\
        3WLGNN & 3 & 102150 & 0.407$\pm$0.028 & 0.272$\pm$0.037 & 111.25 & 286.23s/8.88hr  \\
        3WLGNN-E & 3 & 103098 & \good{0.256$\pm$0.054} & 0.140$\pm$0.044 & 117.25 & 334.69s/10.90hr \\
         & 3 & 507603 & 0.303$\pm$0.068 & 0.173$\pm$0.041 & 120.25 & 329.49s/11.08hr \\
         & 8 & 582824 & 0.303$\pm$0.057 & 0.246$\pm$0.043 & 52.50 & 811.27s/12.15hr \\
        \bottomrule

\end{tabular}
    }
    \vspace{3pt}
    \caption{
    Benchmarking results for ZINC for graph regression. Results (lower is better) are averaged over 4 runs with 4 different seeds. \best{Red}: the best model, \good{Violet}: good models. The suffix -E denotes the use of available edge features, and the suffix -PE denote the use of Laplacian Eigenvectors as node positional encodings with dimension 8.}
    \label{tab:results_zinc}
\end{table}

\subsection{Graph Regression with AQSOL dataset}
AQSOL dataset is based on AqSolDB \citep{sorkun2019aqsoldb} which is a standardized database of $9,982$ molecular graphs with their aqueous solubility values, collected from $9$ different data sources. The aqueous solubility targets are collected from experimental measurements and standardized to $\text{Log}S$ units in AqSolDB. We use these final values as the property to regress in the AQSOL dataset which is the resultant collection after we filter out few graphs with no edges (bonds) and a small number of graphs with missing node feature values. Thus, the total molecular graphs are $9,823$. For each molecular graph, the node features are the types of heavy atoms and the edge features are the types of bonds between them.

\noindent {\bf Splitting.}
We provide a scaffold splitting \citep{hu2020ogb} of the dataset in the ratio $8:1:1$ to have $7,831$ train, $996$ validation and $996$ test graphs.\\
{\bf Training.}
For the learning rate strategy across all GNNs, an initial learning rate is set to $1\times10^{-3}$, the reduce factor is $0.5$, and the stopping learning rate is $1\times10^{-5}$. The patience value is $5$ for 3WLGNN and RingGNN, and $10$ for all other GNNs. \\
{\bf Performance Measure.} 
Similar to ZINC, the performance measure is the mean absolute error (MAE) between the predicted and the actual aqueous solubility values. \\
{\bf Results.} The numerical results are presented in Table \ref{tab:results_aqsol} and analysed in Section \ref{sec:eval}.

\begin{table}[!htb]
    \centering
    \scalebox{0.78}{
    \begin{tabular}{rr|ccccc}
        \toprule
        & & \multicolumn{5}{c}{\textbf{AQSOL}} \\
        \textbf{Model} & \textbf{$L$} & \textbf{\#Param} & \textbf{TestMAE$\pm$s.d.} & \textbf{TrainMAE$\pm$s.d.} & \textbf{Epochs} & \textbf{Epoch/Total} \\
        \midrule
        MLP & 4 & 114525 & 1.744$\pm$0.016 & 1.413$\pm$0.042 & 85.75 & 0.61s/0.02hr\\
        \cmidrule{1-7}
        \textit{vanilla} GCN & 4 & 108442 & 1.483$\pm$0.014 & 0.791$\pm$0.034 & 110.25 & 1.14s/0.04hr\\
        & 16 & 511443 & 1.458$\pm$0.011 & 0.567$\pm$0.027 & 121.50 & 2.83s/0.10hr\\
        GraphSage & 4 & 109620 & 1.431$\pm$0.010 & 0.666$\pm$0.027 & 106.00 & 1.51s/0.05hr\\
        & 16 & 509078 & 1.402$\pm$0.013 & 0.402$\pm$0.013 & 110.50 & 3.20s/0.10hr\\
        \cmidrule{1-7}
        GCN & 4 & 108442 & 1.372$\pm$0.020 & 0.593$\pm$0.030 & 135.00 & 1.28s/0.05hr\\
        & 16 & 511443 & 1.333$\pm$0.013 & 0.382$\pm$0.018 & 137.25 & 3.31s/0.13hr\\
        MoNet & 4 & 109332 & 1.395$\pm$0.027 & 0.557$\pm$0.022 & 125.50 & 1.68s/0.06hr\\
        & 16 & 507750 & 1.501$\pm$0.056 & 0.444$\pm$0.024 & 110.00 & 3.62s/0.11hr\\
        GAT & 4 & 108289 & 1.441$\pm$0.023 & 0.678$\pm$0.021 & 104.50 & 1.92s/0.06hr\\
        & 16 & 540673 & 1.403$\pm$0.008 & 0.386$\pm$0.014 & 111.75 & 4.44s/0.14hr\\
        GatedGCN & 4 & 108325 & 1.352$\pm$0.034 & 0.576$\pm$0.056 & 142.75 & 2.28s/0.09hr\\
        & 16 & 507039 & 1.355$\pm$0.016 & 0.465$\pm$0.038 & 99.25 & 5.52s/0.16hr\\
        GatedGCN-E & 4 & 108535 & 1.295$\pm$0.016 & 0.544$\pm$0.033 & 116.25 & 2.29s/0.08hr\\
        & 16 & 507273 & 1.308$\pm$0.013 & 0.367$\pm$0.012 & 110.25 & 5.61s/0.18hr\\
        GatedGCN-E-PE & 16 & 507663 & \textbf{\best{0.996$\pm$0.008}} & 0.372$\pm$0.016 & 105.25 & 5.70s/0.30hr\\
        \cmidrule{1-7}
        GIN & 4 & 107149 & 1.894$\pm$0.024 & 0.660$\pm$0.027 & 115.75 & 1.55s/0.05hr\\
        & 16 & 514137 & 1.962$\pm$0.058 & 0.850$\pm$0.054 & 128.50 & 3.97s/0.14hr\\
        RingGNN & 2 & 116643 & 20.264$\pm$7.549 & 0.625$\pm$0.018 & 54.25 & 113.99s/1.76hr\\
        RingGNN-E & 2 & 123157 & 3.769$\pm$1.012 & 0.470$\pm$0.022 & 63.75 & 125.17s/2.26hr\\
        & 2 & 523935 & Diverged & Diverged & Diverged & Diverged\\
        & 8 & - & Diverged & Diverged & Diverged & Diverged\\
        3WLGNN & 3 & 110919 & 1.154$\pm$0.050 & 0.434$\pm$0.026 & 66.75 & 130.92s/2.48hr\\
        & 3 & 525423 & 1.108$\pm$0.036 & 0.405$\pm$0.031 & 70.75 & 131.12s/2.62hr\\
        3WLGNN-E & 3 & 112104 & \good{1.042$\pm$0.064} & 0.307$\pm$0.024 & 68.50 & 139.04s/2.70hr\\
        & 3 & 528123 & \good{1.052$\pm$0.034} & 0.287$\pm$0.023 & 67.00 & 140.43s/2.67hr\\
        & 8 & - & Diverged & Diverged & Diverged & Diverged\\
        \bottomrule
        
    \end{tabular}
    }
    \caption{Benchmarking results for AQSOL for graph regression. Results (lower is better) are averaged over 4 runs with 4 different seeds. \textbf{\best{Red}}: the best model, \good{Violet}: good models. The suffix -E denotes the use of available edge features, and the suffix -PE denote the use of Laplacian Eigenvectors as node positional encodings with dimension 4.}
    \label{tab:results_aqsol}
\end{table}

\subsection{Link Prediction with OGBL-COLLAB dataset}

\begin{table}[t!]
    \centering
    \scalebox{0.78}{
    \begin{tabular}{rr|ccccc}
        \toprule
        & & \multicolumn{5}{c}{\textbf{OGBL-COLLAB}}\\
        \multirow{1}{*}{\textbf{Model}} & \textbf{$L$} & \textbf{\#Param} & \multirow{1}{*}{\textbf{Test Hits$\pm$s.d.}} & \multirow{1}{*}{\textbf{Train Hits$\pm$s.d.}} & \multirow{1}{*}{\textbf{\#Epoch}} & \multirow{1}{*}{\textbf{Epoch/Total}} \\
        \midrule
        MLP & 3 & 39441 & 20.350$\pm$2.168 & 29.807$\pm$3.360 & 147.50 & 2.09s/0.09hr \\
        \midrule
        \textit{vanilla} GCN & 3 & 40479 & 50.422$\pm$1.131 & 92.112$\pm$0.991 & 122.50 & 351.05s/12.04hr \\
        GraphSage & 3 & 39856 & \good{51.618$\pm$0.690} & 99.949$\pm$0.052 & 152.75 & 277.93s/11.87hr \\
        \midrule
        GCN & 3 & 40479 & 48.956$\pm$1.143 & 87.385$\pm$2.056 & 142.25 & 7.66s/0.31hr\\
        MoNet & 3 & 39751 & 36.144$\pm$2.191 & 61.156$\pm$3.973 & 167.50 & 26.69s/1.26hr \\
        GAT & 3 & 42637 & \good{51.501$\pm$0.962} & 97.851$\pm$1.114 & 157.00 & 18.12s/0.80hr \\
        GatedGCN & 3 & 40965 & \good{52.635$\pm$1.168} & 96.103$\pm$1.876 & 95.00 & 453.47s/12.09hr \\
        GatedGCN-PE & 3 & 41889 & \best{52.849$\pm$1.345} & 96.165$\pm$0.453 & 94.75 & 452.75s/12.08hr \\
        GatedGCN-E & 3 & 40965 & 49.212$\pm$1.560 & 88.747$\pm$1.058 & 95.00 & 451.21s/12.03hr \\
        \midrule
        GIN & 3 & 39544 & 41.730$\pm$2.284 & 70.555$\pm$4.444 & 140.25 & 8.66s/0.34hr \\
        \multirow{3}{*}{RingGNN} & \multirow{3}{*}{-} & \multirow{3}{*}{-} & \multirow{3}{*}{OOM} & \multicolumn{3}{c}{ 
        \multirow{6}{2in}{\textit{RingGNN and 3WLGNN rely on dense tensors which leads to \textbf{OOM} on both GPU and CPU memory.} 
        }} \\
        & \\
         & \\
        \multirow{3}{*}{3WLGNN} & \multirow{3}{*}{-} & \multirow{3}{*}{-} & \multirow{3}{*}{OOM} & \\
        &\\
        & \\
        \midrule
        Matrix Fact. & - & 60546561 & 44.206$\pm$0.452 & 100.000$\pm$0.000 & 254.33 & 2.66s/0.21hr \\
        \bottomrule

\end{tabular}
    }
    \vspace{3pt}
    \caption{
    Benchmarking results for OGBL-COLLAB for link prediction. Results (higher is better) are averaged over 4 runs with 4 different seeds. \best{Red}: the best model, \good{Violet}: good models. The suffix -E denotes the use of available edge features, and the suffix -PE denote the use of Laplacian Eigenvectors as node positional encodings with dimension 20.
    }
    \label{tab:results_collab}
\end{table}

OGBL-COLLAB is a link prediction dataset proposed by OGB \citep{hu2020ogb} corresponding to a collaboration network between approximately 235K scientists, indexed by Microsoft Academic Graph~\citep{wang2020microsoft}. 
Nodes represent scientists and edges denote collaborations between them.
For node features, OGB provides 128-dimensional vectors, obtained by averaging the word embeddings of a scientist's papers. 
The year and number of co-authored papers in a given year are concatenated to form edge features.
The graph can also be viewed as a dynamic multi-graph, since two nodes may have multiple temporal edges between if they collaborate over multiple years.

\noindent {\bf Splitting.} 
We use the realistic training, validation and test edge splits provided by OGB. 
Specifically, they use collaborations until 2017 as training edges, those in 2018 as validation edges, and those in 2019 as test edges. \\
{\bf Training.} All GNNs use a consistent learning rate strategy: an initial learning rate is set to $1\times10^{-3}$, the reduce factor is $0.5$, the patience value is $10$, and the stopping learning rate is $1\times10^{-5}$. \\
{\bf Performance Measure.} 
We use the evaluator provided by OGB, which aims to measure a model's ability to predict future collaboration relationships given past collaborations. 
Specifically, they rank each true collaboration among a set of 100,000 randomly-sampled negative collaborations, and count the ratio of positive edges that are ranked at K-place or above (Hits@K, with K = 50). \\
{\bf Matrix Factorization Baseline.}
In addition to GNNs, we report performance for a simple matrix factorization baseline \citep{hu2020ogb}, which trains 256-dimensional embeddings for each of the 235K nodes.
Comparing GNNs to matrix factorization tells us whether models leverage node features in addition to graph structure, as matrix factorization can be thought of as feature-agnostic.\\
{\bf Results.} The numerical results are presented in Table \ref{tab:results_collab} and discussed in Section \ref{sec:eval}.

\subsection{Node Classification with WikiCS dataset}

\begin{table}[t!]
    \centering
    \scalebox{0.78}{
    \begin{tabular}{rr|ccccc}
        \toprule
        & & \multicolumn{5}{c}{\textbf{WikiCS}}\\
        \multirow{1}{*}{\textbf{Model}} & \textbf{$L$} & \textbf{\#Param} & \multirow{1}{*}{\textbf{Test Acc.$\pm$s.d.}} & \multirow{1}{*}{\textbf{Train Acc.$\pm$s.d.}} & \multirow{1}{*}{\textbf{\#Epoch}} & \multirow{1}{*}{\textbf{Epoch/Total}} \\
        \midrule
        MLP & 4 & 110710 & 59.452$\pm$2.327 & 85.347$\pm$5.440 & 322.46 & 0.01s/0.03hr \\
        \midrule
        \textit{vanilla} GCN & 4 & 104560 & 77.103$\pm$0.830 & 98.918$\pm$0.619 & 293.84 & 0.05s/0.10hr \\
        GraphSage & 4 & 101775 & 74.767$\pm$0.950 & 99.976$\pm$0.095 & 303.68 & 0.06s/0.12hr\\
        \midrule
        GCN & 4 & 104560 & \good{77.469$\pm$0.854} & 98.925$\pm$0.590 & 299.85 & 0.06s/0.11hr\\
        MoNet &  4 & 106182 & \good{77.431$\pm$0.669} & 98.737$\pm$0.710 & 355.81 & 0.17s/0.36hr\\
        MoNet-PE & 4 & 107862 & \textbf{\best{77.481$\pm$0.712}} & 98.767$\pm$0.726 & 357.74 & 0.19s/0.81hr\\
        GAT & 4 & 105520 & 76.908$\pm$0.821 & 99.914$\pm$0.262 & 275.48 & 1.12s/2.22hr \\
        GatedGCN & 4 & 109280 & \multicolumn{2}{c}{OOM}\\
        \midrule
        GIN & 4 & 109782 & 75.857$\pm$0.577 & 99.575$\pm$0.388 & 321.25 & 0.06s/0.13hr\\
        \bottomrule
\end{tabular}
    }
    \vspace{3pt}
    \caption{
    Benchmarking results for WikiCS for node classification. Results (higher is better) are averaged over 4 runs with 4 different seeds. \best{Red}: the best model, \good{Violet}: good models. The suffix -PE denote the use of Laplacian Eigenvectors as node positional encodings with dimension 20.
    }
    \label{tab:results_wikics}
\end{table}

WikiCS is a node classification dataset based on an extracted subset of Wikipedia's Computer Science articles \citep{mernyei2020wiki}. It is a single graph dataset with $11,701$ nodes and $216,123$ edges where each node corresponds to an article, and each edge corresponds to a hyperlink. Each node belongs to a label out of total 10 classes representing the article's category. The average of the article text's pre-trained GloVe word embeddings \citep{pennington2014glove} is assigned as 300-dimensional node features. Compared to previous single-graph node classification benchmarks such as Cora and Citeseer, WikiCS dataset has denser node neighborhoods and each node's connectivity is spread across nodes from varying class labels. Additionally, as shown in \cite{mernyei2020wiki}, the average shortest path length in WikiCS is smaller compared to Cora and Citeseer. Thus, on average, a larger node neighborhood and smaller shortest path length makes WikiCS an appropriate benchmark to test out neighborhood computation functions in GNNs' design.

\noindent {\bf Splitting.} 
We follow the splitting defined in \cite{mernyei2020wiki} that has \edits{20 different training, validation and early stopping splits consisting of 5\% nodes, 22.5\% nodes and 22.5\% nodes of each class respectively. 50\% nodes from each class, which are not in the training or validation split, are assigned as test splits. We combine the two original validation (22.5\% nodes) and early stopping (22.5\% nodes) splits to make the new validation (45\% nodes) splits since we do not use separate early stopping splits in our benchmark.}\\
{\bf Training.} As consistent learning rate strategy across GNNs, an initial learning rate is set to $1\times10^{-2}$, the reduce factor is $0.5$, the patience value is $25$, and the stopping learning rate is $1\times10^{-5}$. Since there are 20 different training and validation splits, the training is done 20 times using these splits, and evaluated on the single test split. This is done for 4 times with 4 different seeds. Finally, the average of the 20 $\times$ 4 = 80 runs is reported.\\
{\bf Performance Measure.} 
The performance measure is the classification accuracy between the predicted and groundtruth label for each node. \\
{\bf Results.} \edits{The numerical results are presented in Table \ref{tab:results_wikics} and discussed in Section \ref{sec:eval}.}

\subsection{Graph Classification with Super-pixel (MNIST/CIFAR10) datasets}
The super-pixels datasets test graph classification using the popular MNIST and CIFAR10 image classification datasets.
Our main motivation to use these datasets is as sanity-checks: we expect most GNNs to perform close to 100\% for MNIST and well enough for CIFAR10. Besides, the use of super-pixel image datasets is suggestive of the way image datasets can be used for graph learning research.

\begin{figure}[t]
\begin{tabular}{c c}
\subfloat[MNIST]{\includegraphics[width = 2.8in]{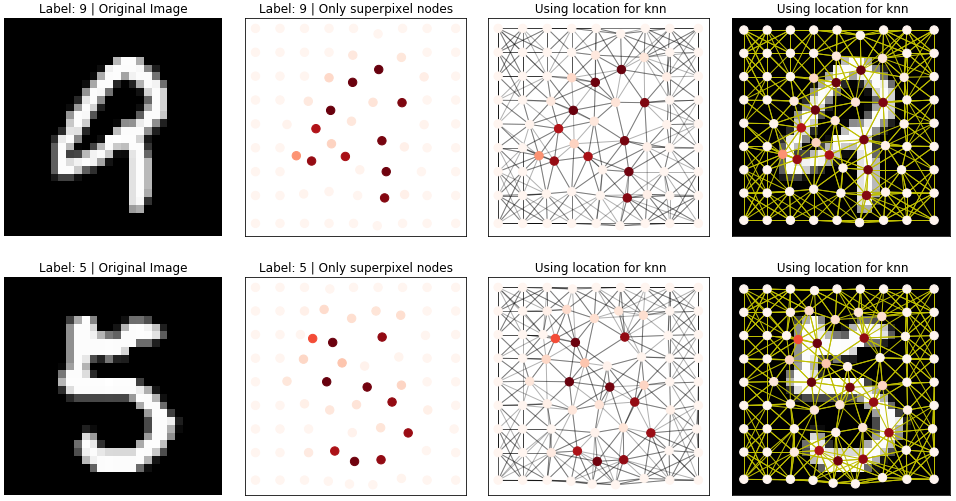}} &
\subfloat[CIFAR10]{\includegraphics[width = 2.8in]{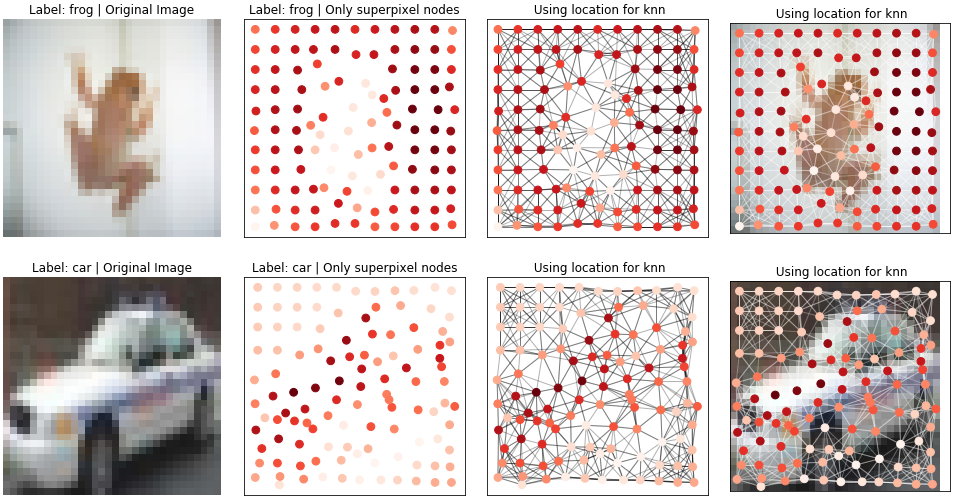}} \\
\end{tabular}
\caption{Sample images and their superpixel graphs. The graphs of SLIC superpixels (at most 75 nodes for MNIST and 150 nodes for CIFAR10) are  8-nearest neighbor graphs in the Euclidean space and node colors denote the mean pixel intensities.  }
\label{fig:superpixel_graph}
\end{figure}

The original MNIST and CIFAR10 images are converted to graphs using super-pixels. Super-pixels represent small regions of homogeneous intensity in images, and can be extracted with the SLIC technique \citep{10.1109/TPAMI.2012.120}. 
We use SLIC super-pixels from \citep{knyazev2019understanding}\footnote{\small{\url{https://github.com/bknyaz/graph_attention_pool}}}. 
For each sample, we build a $k$-nearest neighbor adjacency matrix with
 \begin{equation}
W_{ij}^{k-\tiny{\textrm{NN}}} = \exp \left( -\frac{\|x_i - x_j\|^2}{\sigma^2_x} \right),
\label{knn_weights}
\end{equation}
where $x_i, x_j$ are the 2-D coordinates of super-pixels $i,j$, and $\sigma_x$ is the scale parameter defined as the averaged distance $x_k$ of the $k$ nearest neighbors for each node. We use $k=8$ for both MNIST and CIFAR10, whereas the maximum number of super-pixels (nodes) are 75 and 150 for MNIST and CIFAR10, respectively. The resultant graphs are of sizes $40$-$75$ nodes for MNIST and $85$-$150$ nodes for CIFAR10. Figure \ref{fig:superpixel_graph} presents visualizations of the super-pixel graphs. 



\noindent {\bf Splitting.} We use the standard splits of MNIST and CIFAR10. MNIST has $55,000$ train, $5,000$ validation, $10,000$ test graphs and CIFAR10 has $45,000$ train, $5,000$ validation, $10,000$ test graphs. The $5,000$ graphs for validation set are randomly sampled from the training set and the same splits are used for every GNN.\\
{\bf Training.} The learning decay rate strategy is adopted with an initial learning rate of $1\times10^{-3}$, reduce factor $0.5$, patience value $10$, and the stopping learning rate $1\times10^{-5}$ for all GNNs, except for 3WLGNN and RingGNN where we experienced a difficulty in training, leading us to slightly adjust their learning rate schedule hyperparameters. For both 3WLGNN and RingGNN, the patience value is changed to $5$. For RingGNN, the initial learning rate is changed to $1\times10^{-4}$ and the stopping learning rate is changed to $1\times10^{-6}$.\\
{\bf Performance Measure.} The classification accuracy between the predicted and groundtruth label for each graph is the performance measure.\\
{\bf Results.} The numerical results are presented in Table \ref{tab:results_superpixels} and discussed in Section \ref{sec:eval}.

\begin{table}[t!]
    \centering
    \scalebox{0.5}{
    \begin{tabular}{rr|ccccc|ccccc}
        \toprule
        & & \multicolumn{5}{c}{\textbf{MNIST}} & \multicolumn{5}{c}{\textbf{CIFAR10}}\\
        \textbf{Model} & \textbf{$L$} & \textbf{\#Param} & \textbf{Test Acc.$\pm$s.d.} & \textbf{Train Acc.$\pm$s.d.} & \textbf{\#Epoch} & \textbf{Epoch/Total} & \textbf{\#Param} & \textbf{Test Acc.$\pm$s.d.} & \textbf{Train Acc.$\pm$s.d.} & \textbf{\#Epoch} & \textbf{Epoch/Total} \\
        \midrule
        \midrule
        MLP & 4 & 104044 & 95.340$\pm$0.138 & 97.432$\pm$0.470 & 232.25 & 22.74s/1.48hr & 104380 & 56.340$\pm$0.181 & 65.113$\pm$1.685 & 185.25 & 29.48s/1.53hr\\
        \midrule
        \textit{vanilla} GCN & 4 & 101365 & 90.705$\pm$0.218 & 97.196$\pm$0.223 & 127.50 & 83.41s/2.99hr & 101657 & 55.710$\pm$0.381 & 69.523$\pm$1.948 & 142.50 & 109.70s/4.39hr\\
        GraphSage & 4 & 104337 & \best{97.312$\pm$0.097} & 100.000$\pm$0.000 & 98.25 & 113.12s/3.13hr & 104517 & \good{65.767$\pm$0.308} & 99.719$\pm$0.062 & 93.50 & 124.61s/3.29hr\\
        \midrule
        GCN & 4 & 101365 & 90.120$\pm$0.145 & 96.459$\pm$1.020 & 116.75 & 37.06s/1.22hr & 101657 & 54.142$\pm$0.394 & 70.163$\pm$3.429 & 140.50 & 47.16s/1.86hr\\
        MoNet & 4 & 104049 & 90.805$\pm$0.032 & 96.609$\pm$0.440 & 146.25 & 93.19s/3.82hr & 104229 & 54.655$\pm$0.518 & 65.911$\pm$2.515 & 141.50 & 97.13s/3.85hr\\
        GAT & 4 & 110400 & 95.535$\pm$0.205 & 99.994$\pm$0.008 & 104.75 & 42.26s/1.25hr & 110704 & 64.223$\pm$0.455 & 89.114$\pm$0.499 & 103.75 & 55.27s/1.62hr\\
        GatedGCN & 4 & 104217 & \best{97.340$\pm$0.143} & 100.000$\pm$0.000 & 96.25 & 128.79s/3.50hr & 104357 & \best{67.312$\pm$0.311} & 94.553$\pm$1.018 & 97.00 & 154.15s/4.22hr\\
        \midrule
        GIN & 4 & 105434 & \good{96.485$\pm$0.252} & 100.000$\pm$0.000 & 128.00 & 39.22s/1.41hr & 105654 & 55.255$\pm$1.527 & 79.412$\pm$9.700 & 141.50 & 52.12s/2.07hr\\
        RingGNN & 2 & 105398 & 11.350$\pm$0.000 & 11.235$\pm$0.000 & 14.00 & 2945.69s/12.77hr & 105165 & 19.300$\pm$16.108 & 19.556$\pm$16.397 & 13.50 & 3112.96s/13.00hr\\
        & 2 & 505182 & 91.860$\pm$0.449 & 92.169$\pm$0.505 & 16.25 & 2575.99s/12.63hr & 504949 & 39.165$\pm$17.114 & 40.209$\pm$17.790 & 13.75 & 2998.24s/12.60hr \\
        & 8 & 506357 & Diverged & Diverged & Diverged & Diverged & 510439 & Diverged & Diverged & Diverged & Diverged\\
        3WLGNN & 3 & 108024 & 95.075$\pm$0.961 & 95.830$\pm$1.338 & 27.75 & 1523.20s/12.40hr & 108516 & 59.175$\pm$1.593 & 63.751$\pm$2.697 & 28.50 & 1506.29s/12.60hr \\
        & 3 & 501690 & 95.002$\pm$0.419 & 95.692$\pm$0.677 & 26.25 & 1608.73s/12.42hr & 502770 & 58.043$\pm$2.512 & 61.574$\pm$3.575 & 20.00 & 2091.22s/12.55hr\\
        & 8 & 500816 & Diverged & Diverged & Diverged & Diverged & 501584 & Diverged & Diverged & Diverged & Diverged \\
        \midrule
\end{tabular}
    }
    \vspace{3pt}
    \caption{
    Benchmarking results for Super-pixels datasets for graph classification. Results (higher is better) are averaged over 4 runs with 4 different seeds. \best{Red}: the best model, \good{Violet}: good models. 
    }
    \label{tab:results_superpixels}
\end{table}

\subsection{Node Classification with SBM (PATTERN/CLUSTER) datasets}
\label{sec:sbms_data_exp}
The SBM datasets consider node-level tasks of graph pattern recognition \citep{art:ScarselliGoriTsoiHagenbuchnerMonfardini09} -- PATTERN and semi-supervised graph clustering -- CLUSTER. 
The graphs are generated with the Stochastic Block Model (SBM) \citep{abbe2017community}, which is widely used to model communities in social networks by modulating the intra- and extra-communities connections, thereby controlling the difficulty of the task.
A SBM is a random graph which assigns communities to each node as follows: any two vertices are connected with the probability $p$ if they belong to the same community, or they are connected with the probability $q$ if they belong to different communities (the value of $q$ acts as the noise level).

{\bf PATTERN:} The graph pattern recognition task, presented in \cite{art:ScarselliGoriTsoiHagenbuchnerMonfardini09}, aims at finding a fixed graph pattern $P$ embedded in larger graphs $G$ of variable sizes.
For all data, we generate graphs $G$ with 5 communities with sizes randomly selected between $[5,35]$. 
The SBM of each community is $p=0.5, q=0.35$, and the node features on $G$ are generated with a uniform random distribution with a vocabulary of size $3$, \textit{i.e.} $\{0,1,2\}$. 
We randomly generate $100$ patterns $P$ composed of $20$ nodes with intra-probability $p_P=0.5$ and extra-probability $q_P=0.5$ (\textit{i.e.}, 50\% of nodes in $P$ are connected to $G$). 
The node features for $P$ are also generated as a random signal with values $\{0,1,2\}$. 
The graphs are of sizes $44$-$188$ nodes. 
The output node labels have value 1 if the node belongs to $P$ and value 0 if it is in $G$. 

{\bf CLUSTER:} For the semi-supervised clustering task, we generate 6 SBM clusters with sizes randomly selected between $[5,35]$ and probabilities $p=0.55, q=0.25$. 
The graphs are of sizes $40$-$190$ nodes. 
Each node can take an input feature value in $\{0,1,2,..,6\}$. If the value is 1, the node belongs to class 0, value 2 corresponds to class 1, \dots, value 6 corresponds to class 5. 
Otherwise, if the value is 0, the class of the node is unknown and will be inferred by the GNN. 
There is only one labelled node that is randomly assigned to each community and most node features are set to 0.
The output node labels are defined as the community/cluster class labels.

\noindent {\bf Splitting.} The PATTERN dataset has $10,000$ train, $2,000$ validation, $2,000$ test graphs and CLUSTER dataset has $10,000$ train, $1,000$ validation, $1,000$ test graphs. We save the generated splits and use the same sets in all models for fair comparison.\\
{\bf Training.} As presented in the standard experimental protocol in Section \ref{sec:expsetup}, we use Adam optimizer with a learning rate decay strategy. For all GNNs, an initial learning rate is set to $1\times10^{-3}$, the reduce factor is $0.5$, the patience value is $5$, and the stopping learning rate is $1\times10^{-5}$. \\
{\bf Performance Measure.} The performance measure is the average node-level accuracy weighted with respect to the class sizes.\\
{\bf Results.} Our numerical results are presented in Table \ref{tab:results_sbm} and discussed in Section \ref{sec:eval} together with other benchmark results.

\begin{table}[t!]
    \centering
    \scalebox{0.5}{
    \begin{tabular}{rr|ccccc|ccccc}
        \toprule
        & & \multicolumn{5}{c}{\textbf{PATTERN}} & \multicolumn{5}{c}{\textbf{CLUSTER}}\\
        \textbf{Model} & \textbf{$L$} & \textbf{\#Param} & \textbf{Test Acc.$\pm$s.d.} & \textbf{Train Acc.$\pm$s.d.} & \textbf{\#Epoch} & \textbf{Epoch/Total} & \textbf{\#Param} & \textbf{Test Acc.$\pm$s.d.} & \textbf{Train Acc.$\pm$s.d.} & \textbf{\#Epoch} & \textbf{Epoch/Total} \\
        \midrule
        \midrule
        MLP & 4 & 105263 & 50.519$\pm$0.000 & 50.487$\pm$0.014 & 42.25 & 8.95s/0.11hr & 106015 & 20.973$\pm$0.004 & 20.938$\pm$0.002 & 42.25 & 5.83s/0.07hr\\
        \midrule
        \textit{vanilla} GCN & 4 & 100923 & 63.880$\pm$0.074 & 65.126$\pm$0.135 & 105.00 & 118.85s/3.51hr & 101655 & 53.445$\pm$2.029 & 54.041$\pm$2.197 & 70.00 & 65.72s/1.30hr\\
        & 16 & 500823 & 71.892$\pm$0.334 & 78.409$\pm$1.592 & 81.50 & 492.19s/11.31hr & 501687 & 68.498$\pm$0.976 & 71.729$\pm$2.212 & 79.75 & 270.28s/6.08hr\\
        GraphSage & 4 & 101739 & 50.516$\pm$0.001 & 50.473$\pm$0.014 & 43.75 & 93.41s/1.17hr & 102187 & 50.454$\pm$0.145 & 54.374$\pm$0.203 & 64.00 & 53.56s/0.97hr\\
        & 16 & 502842 & 50.492$\pm$0.001 & 50.487$\pm$0.005 & 46.50 & 391.19s/5.19hr & 503350 & 63.844$\pm$0.110 & 86.710$\pm$0.167 & 57.75 & 225.61s/3.70hr\\
        \midrule
        GCN & 4 & 100923 & \good{85.498$\pm$0.045} & 85.598$\pm$0.043 & 65.00 & 19.21s/0.36hr & 101655 & 47.828$\pm$1.510 & 48.258$\pm$1.607 & 63.50 & 12.84s/0.23hr\\
        & 16 & 500823 & \good{85.614$\pm$0.032} & 86.034$\pm$0.087 & 66.00 & 37.08s/0.70hr & 501687 & 69.026$\pm$1.372 & 73.749$\pm$2.570 & 77.75 & 30.20s/0.66hr\\
        MoNet & 4 & 103775 & \good{85.482$\pm$0.037} & 85.569$\pm$0.044 & 89.75 & 35.71s/0.90hr & 104227 & 58.064$\pm$0.131 & 58.454$\pm$0.183 & 76.25 & 24.29s/0.52hr\\
        & 16 & 511487 & \good{85.582$\pm$0.038} & 85.720$\pm$0.068 & 81.75 & 68.49s/1.58hr & 511999 & 66.407$\pm$0.540 & 67.727$\pm$0.649 & 77.75 & 47.82s/1.05hr\\
        GAT & 4 & 109936 & 75.824$\pm$1.823 & 77.883$\pm$1.632 & 96.00 & 20.92s/0.57hr & 110700 & 57.732$\pm$0.323 & 58.331$\pm$0.342 & 67.25 & 14.17s/0.27hr\\
        & 16 & 526990 & 78.271$\pm$0.186 & 90.212$\pm$0.476 & 53.50 & 50.33s/0.77hr & 527874 & \good{70.587$\pm$0.447} & 76.074$\pm$1.362 & 73.50 & 35.94s/0.75hr\\
        GatedGCN & 4 & 104003 & 84.480$\pm$0.122 & 84.474$\pm$0.155 & 78.75 & 139.01s/3.09hr & 104355 & 60.404$\pm$0.419 & 61.618$\pm$0.536 & 94.50 & 79.97s/2.13hr\\
        & 16 & 502223 & \good{85.568$\pm$0.088} & 86.007$\pm$0.123 & 65.25 & 644.71s/11.91hr & 502615 & \good{73.840$\pm$0.326} & 87.880$\pm$0.908 & 60.00 & 400.07s/6.81hr\\
        GatedGCN-PE & 16 & 502457 & \best{86.508$\pm$0.085} & 86.801$\pm$0.133 & 65.75 & 647.94s/12.08hr & 504253 & \best{76.082$\pm$0.196} & 88.919$\pm$0.720 & 57.75 & 399.66s/6.58hr \\
        \midrule
        GIN & 4 & 100884 & \good{85.590$\pm$0.011} & 85.852$\pm$0.030 & 93.00 & 15.24s/0.40hr & 103544 & 58.384$\pm$0.236 & 59.480$\pm$0.337 & 74.75 & 10.71s/0.23hr\\
        & 16 & 508574 & 85.387$\pm$0.136 & 85.664$\pm$0.116 & 86.75 & 25.14s/0.62hr & 517570 & 64.716$\pm$1.553 & 65.973$\pm$1.816 & 80.75 & 20.67s/0.47hr\\
        RingGNN & 2 & 105206 & \good{86.245$\pm$0.013} & 86.118$\pm$0.034 & 75.00 & 573.37s/12.17hr & 104746 & 42.418$\pm$20.063 & 42.520$\pm$20.212 & 74.50 & 428.24s/8.79hr\\
        & 2 & 504766 & \good{86.244$\pm$0.025} & 86.105$\pm$0.021 & 72.00 & 595.97s/12.15hr & 524202 & 22.340$\pm$0.000 & 22.304$\pm$0.000 & 43.25 & 501.84s/6.22hr\\
        & 8 & 505749 & \textrm{Diverged} & \textrm{Diverged} & \textrm{Diverged} & \textrm{Diverged} & 514380 & \textrm{Diverged} & \textrm{Diverged} & \textrm{Diverged} & \textrm{Diverged}\\
        3WLGNN & 3 & 103572 & \good{85.661$\pm$0.353} & 85.608$\pm$0.337 & 95.00 & 304.79s/7.88hr & 105552 & 57.130$\pm$6.539 & 57.404$\pm$6.597 & 116.00 & 219.51s/6.52hr\\
        & 3 & 502872 & \good{85.341$\pm$0.207} & 85.270$\pm$0.198 & 81.75 & 424.23s/9.56hr & 507252 & 55.489$\pm$7.863 & 55.736$\pm$8.024 & 66.00 & 319.98s/5.79hr \\
        & 8 & 581716 & \textrm{Diverged} & \textrm{Diverged} & \textrm{Diverged} & \textrm{Diverged} & 586788 & \textrm{Diverged} & \textrm{Diverged} & \textrm{Diverged} & \textrm{Diverged}\\
        \midrule
        \end{tabular}
    }
    \vspace{3pt}
    \caption{
    Benchmarking results for SBMs datasets for node classification. Results (higher is better) are averaged over 4 runs with 4 different seeds. \best{Red}: the best model, \good{Violet}: good models. The suffix -PE denote the use of Laplacian Eigenvectors as node positional encodings with dimension 2 for PATTERN and 20 for CLUSTER. 
    }
    \label{tab:results_sbm}
\end{table}

\subsection{Edge Classification/Link Prediction with TSP dataset}
\label{sec:tsp_data_exp}

\begin{figure}[!t]
	\centering
	\subfloat[TSP50]{
      \includegraphics[width=0.27\textwidth]{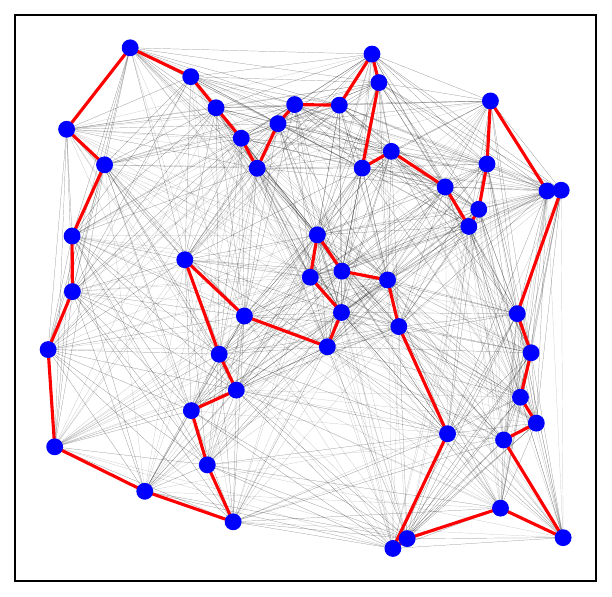}}
    \subfloat[TSP200]{
      \includegraphics[width=0.27\textwidth]{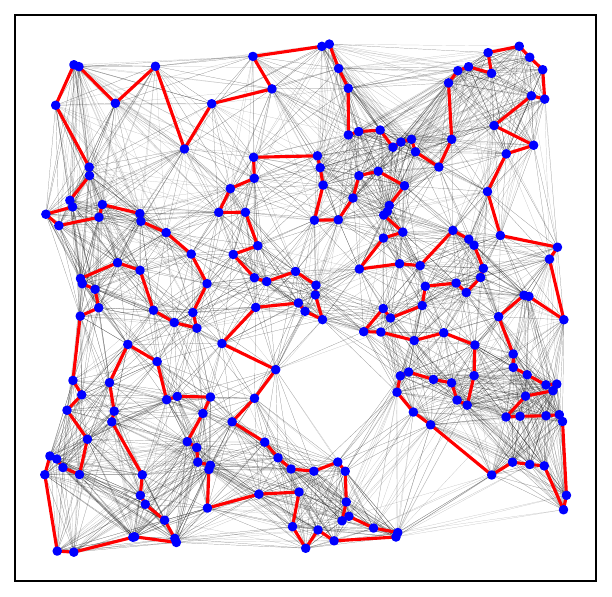}}
    \subfloat[TSP500]{
      \includegraphics[width=0.27\textwidth]{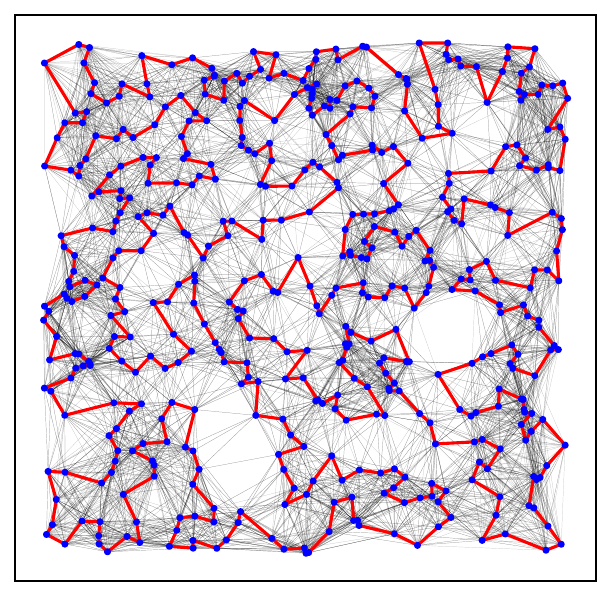}}
	\caption{Sample graphs from the TSP dataset. Nodes are colored blue and edges on the groundtruth TSP tours are colored red.
	}
	\label{fig:tsp}
\end{figure}

Leveraging machine learning for solving NP-hard combinatorial optimization problems (COPs) has been the focus of intense research in recent years \citep{vinyals2015pointer,bengio2018machine}.
Recently proposed learning-driven solvers for COPs \citep{khalil2017learning,kool2018attention,joshi2019efficient} combine GNNs with classical search to predict approximate solutions directly from problem instances (represented as graphs).
Consider the intensively studied Travelling Salesman Problem (TSP), which asks the following question:
\textit{“Given a list of cities and the distances between each pair of cities, what is the shortest possible route that visits each city and returns to the origin city?" }
Formally, given a 2D Euclidean graph, one needs to find an optimal sequence of
nodes, called a tour, with minimal total edge weights (tour length).
TSP's \textit{multi-scale} nature makes it a challenging graph task which requires reasoning about both local node neighborhoods as well as global graph structure.

For our experiments with TSP, we follow the learning-based approach to COPs described in \cite{joshi2022learning}, where a GNN is the backbone architecture for assigning probabilities to each edge as belonging/not belonging to the predicted solution set.
The probabilities are then converted into discrete decisions through graph search techniques.
Each instance is a graph of $n$ node locations sampled uniformly in the unit square $S = \{x_i\}_{i=1}^n$ and $x_i \in {[0,1]}^2$.
We generate problems of varying size and complexity by uniformly sampling the number of nodes $n \in [50, 500]$ for each instance.

In order to isolate the impact of the backbone GNN architectures from the search component, we pose TSP as a binary edge classification task,
with the groundtruth value for each edge belonging to the TSP tour given by Concorde~\citep{applegate2006concorde}.
For scaling to large instances, we use sparse $k=25$ nearest neighbor graphs instead of full graphs, following \citep{khalil2017learning}.
See Figure \ref{fig:tsp} for sample TSP instances of various sizes.

\noindent {\bf Splitting.} TSP has $10,000$ train, $1,000$ validation and $1,000$ test graphs.\\
{\bf Training.} All GNNs use a consistent learning rate strategy: an initial learning rate is set to $1\times10^{-3}$, the reduce factor is $0.5$, the patience value is $10$, and the stopping learning rate is $1\times10^{-5}$.\\
{\bf Performance Measure.} Given the high class imbalance, \textit{i.e.}, only the edges in the TSP tour have positive label, we use the F1 score for the positive class as our performance measure. \\
{\bf Non-learnt Baseline.} In addition to reporting performance of GNNs, we compare with a simple $k$-nearest neighbor heuristic baseline, defined as follows: Predict true for the edges corresponding to the $k$ nearest neighbors of each node, and false for all other edges. 
We set $k$ = 2 for optimal performance. 
Comparing GNNs to the non-learnt baseline tells us whether models learn something more sophisticated than identifying a node’s nearest neighbors.\\
{\bf Results.} The numerical results are presented in Table \ref{tab:results_tsp} and analysed in Section \ref{sec:eval}.

\begin{table}[t!]
    \centering
    \scalebox{0.78}{
    \begin{tabular}{rr|ccccc}
        \toprule
        & & \multicolumn{5}{c}{\textbf{TSP}}\\
        \multirow{1}{*}{\textbf{Model}} & \multirow{1}{*}{\textbf{$L$}} & \multirow{1}{*}{\textbf{\#Param}} & \multirow{1}{*}{\textbf{Test F1$\pm$s.d.}} & \multirow{1}{*}{\textbf{Train F1$\pm$s.d.}} & \multirow{1}{*}{\textbf{\#Epoch}} & \multirow{1}{*}{\textbf{Epoch/Total}} \\
        \midrule
        \midrule
        MLP & 4 & 96956 & 0.544$\pm$0.001 & 0.544$\pm$0.001 & 164.25 & 50.15s/2.31hr \\
        \midrule
        \textit{vanilla} GCN & 4 & 95702 & 0.630$\pm$0.001 & 0.631$\pm$0.001 & 261.00 & 152.89s/11.15hr \\
        GraphSage & 4 & 99263 & 0.665$\pm$0.003 & 0.669$\pm$0.003 & 266.00 & 157.26s/11.68hr\\
        \midrule
        GCN & 4 & 95702 & 0.643$\pm$0.001 & 0.645$\pm$0.002 & 261.67 & 57.84s/4.23hr \\
        MoNet & 4 & 99007 & 0.641$\pm$0.002 & 0.643$\pm$0.002 & 282.00 & 84.46s/6.65hr \\
        GAT & 4 & 96182 & 0.671$\pm$0.002 & 0.673$\pm$0.002 & 328.25 & 68.23s/6.25hr \\
        GatedGCN & 4 & 97858 & \good{0.791$\pm$0.003} & 0.793$\pm$0.003 & 159.00 & 218.20s/9.72hr \\
        GatedGCN-E & 4 & 97858 & \good{0.808$\pm$0.003} & 0.811$\pm$0.003 & 197.00 & 218.51s/12.04hr \\
        GatedGCN-E & 16 & 500770 & \best{0.838$\pm$0.002} & 0.850$\pm$0.001 & 53.00 & 807.23s/12.17hr \\
        \midrule
        GIN & 4 & 99002 & 0.656$\pm$0.003 & 0.660$\pm$0.003 & 273.50 & 72.73s/5.56hr \\
        RingGNN & 2 & 106862 & 0.643$\pm$0.024 & 0.644$\pm$0.024 & 2.00 & 17850.52s/17.19hr \\
        & 2 & 507938 & 0.704$\pm$0.003 & 0.705$\pm$0.003 & 3.00 & 12835.53s/16.08hr \\
        & 8 & 506564 & Diverged & Diverged & Diverged & Diverged\\
        3WLGNN & 3 & 106366 & 0.694$\pm$0.073 & 0.695$\pm$0.073 & 2.00 & 17468.81s/16.59hr \\
        & 3 & 506681 & 0.288$\pm$0.311 & 0.290$\pm$0.312 & 2.00 & 17190.17s/16.51hr \\
        & 8 & 508832 & OOM & OOM & OOM & OOM \\
        \midrule
        $k$-NN Heuristic & & $k=$2 & \multicolumn{4}{l}{Test F1: 0.693} \\
        \midrule

\end{tabular}
    }
    \vspace{3pt}
    \caption{
    Benchmarking results for TSP for edge classification. Results (higher is better) are averaged over 4 runs with 4 different seeds. \best{Red}: the best model, \good{Violet}: good models. The suffix -E denotes the use of available edge features.
    }
    \label{tab:results_tsp}
\end{table}

\subsection{Graph Classification and Isomorphism Testing with CSL dataset}
\label{sec:csl_data_main}
The Circular Skip Link dataset is a symmetric graph dataset introduced in \cite{murphy2019relational} to test the expressivity of GNNs. Each CSL graph is a 4-regular graph with edges connected to form a cycle and containing skip-links between nodes. Formally, it is denoted by $\mathcal{G}_{N, C}$ where $N$ is the number of nodes and $C$ is the isomorphism class which is the skip-link length of the graph. We use the same dataset $\mathcal{G}_{41, C}$ with $C \in \{2, 3, 4, 5, 6, 9, 11, 12, 13, 16\}$. The dataset is class-balanced with $15$ graphs for every $C$ resulting in a total of $150$ graphs.

\noindent {\bf Splitting.} We perform a 5-fold cross validation split, following \cite{murphy2019relational}, which gives 5 sets of train, validation and test data indices in the ratio $3:1:1$. We use stratified sampling to ensure that the class distribution remains the same across splits. The indices are saved and used across all experiments for fair comparisons.\\
{\bf Training.} For the learning rate strategy across all GNNs, an initial learning rate is set to $5\times10^{-4}$, the reduce factor is $0.5$, the patience value is $5$, and the stopping learning rate is $1\times10^{-6}$. We train on the 5-fold cross validation with 20 different seeds of initialization, following \cite{chen2019equivalence}.
\\
{\bf Performance Measure.} We use graph classification accuracy between the predicted labels and groundtruth labels as our performance measure. The model performance is evaluated on the test split of the 5 folds at every run, and following \cite{murphy2019relational, chen2019equivalence}, we report the maximum, minimum, average and the standard deviation of the 100 scores, \textit{i.e.}, 20 runs of 5-folds.\\
{\bf Results.} The numerical results are reported in Table \ref{tab:csl} and analyzed in Section \ref{sec:lap_pe}. In this paper, we use CSL primarily to validate the impact of having Graph Positional Encodings (Section \ref{sec:lap_pe}) that is proposed as a demonstration of our benchmarking framework to steer new GNN research.

\begin{table}[t]
    \centering
    \scalebox{0.65}{
    \begin{tabular}{rcc|ccc|ccc|cc}
        \toprule
        \multirow{2}{*}{\textbf{Model}} & \multirow{2}{*}{\textbf{$L$}} & \multirow{2}{*}{\textbf{\#Param}} & \multicolumn{3}{c|}{\textbf{Test Accuracy}} & \multicolumn{3}{c|}{\textbf{Train Accuracy}} & \multirow{2}{*}{\textbf{\#Epoch}} & \textbf{Epoch/}  \\
        \cline{4-9}
        & & & \textbf{Mean$\pm$s.d.} & \textbf{Max} & \textbf{Min} & \textbf{Mean$\pm$s.d.} & \textbf{Max} & \textbf{Min} & & \textbf{Total} \\ 
        \midrule
        \midrule
        & & \multicolumn{9}{c}{\textbf{Node Positional Encoding with Laplacian Eigenvectors }} \\
        \midrule
        MLP & 4 & 101235 & 22.567$\pm$6.089 & 46.667 & 10.000 & 30.389$\pm$5.712 & 43.333 & 10.000 & 109.39 & 0.16s/0.03hr \\
        \midrule
        GCN & 4 & 103847 & \best{100.000$\pm$0.000} & 100.000 & 100.000 & 100.000$\pm$0.000 & 100.000 & 100.000 & 125.64 & 0.40s/0.07hr \\
        GraphSage & 4 & 105867 & \good{99.933$\pm$0.467} & 100.000 & 96.667 & 100.000$\pm$0.000 & 100.000 & 100.000 & 155.00 & 0.50s/0.11hr \\
        \midrule
        MoNet & 4 & 105579 & \good{99.967$\pm$0.332} & 100.000 & 96.667 & 100.000$\pm$0.000 & 100.000 & 100.000 & 130.39 & 0.49s/0.09hr\\
        GAT & 4 & 101710 & \good{99.933$\pm$0.467} & 100.000 & 96.667 & 100.000$\pm$0.000 & 100.000 & 100.000 & 133.18 & 0.61s/0.12hr  \\
        GatedGCN & 4 & 105407 & \good{99.600$\pm$1.083} & 100.000 & 96.667 & 100.000$\pm$0.000 & 100.000 & 100.000 & 147.06 & 0.66s/0.14hr \\
        \midrule
        GIN & 4 & 107304 & \good{99.333$\pm$1.333} & 100.000 & 96.667 & 100.000$\pm$0.000 & 100.000 & 100.000 & 62.98 & 0.44s/0.04hr  \\
        RingGNN & 2 & 102726 & 17.233$\pm$6.326 & 40.000 & 10.000 & 26.122$\pm$14.382 & 58.889 & 10.000 & 122.75 & 2.93s/0.50hr \\
        & 2 & 505086 & 25.167$\pm$7.399 & 46.667 & 10.000 & 54.533$\pm$18.415 & 82.222 & 10.000 & 120.58 & 3.11s/0.51hr \\
        3WLGNN & 3 & 102054 & 30.533$\pm$9.863 & 56.667 & 10.000 & 99.644$\pm$1.684 & 100.000 & 88.889 & 74.66 & 2.33s/0.25hr   \\
        & 3 & 505347 & 30.500$\pm$8.197 & 56.667 & 13.333 & 100.000$\pm$0.000 & 100.000 & 100.000 & 66.64 & 2.38s/0.23hr \\
        \midrule
        & & \multicolumn{9}{c}{\textbf{No Node Positional Encoding}} \\
        \midrule
        All MP-GCNs & 4 & 100K & 10.000$\pm$0.000 & 10.000 & 10.000 & 10.000$\pm$0.000 & 10.000 & 10.000 & - & - \\
        RingGNN & 2 & 101138 & 10.000$\pm$0.000 & 10.000 & 10.000 & 10.000$\pm$0.000 & 10.000 & 10.000 & 103.23 & 3.09s/0.45hr \\
        & 2 & 505325 & 10.000$\pm$0.000 & 10.000 & 10.000 & 10.000$\pm$0.000 & 10.000 & 10.000 & 90.04 & 3.28s/0.42hr \\
        3WLGNN & 3 & 102510 & 95.700$\pm$14.850 & 100.000 & 30.000 & 95.700$\pm$14.850 & 100.000 & 30.000 & 475.81 & 2.29s/1.51hr  \\
        & 3 & 506106 & 97.800$\pm$10.916 & 100.000 & 30.000 & 97.800$\pm$10.916 & 100.000 & 30.000 & 283.80 & 2.28s/0.90hr \\
        \bottomrule
    \end{tabular}
    }
    \vspace{3pt}
    \caption{Results for the CSL dataset, with and without Laplacian Positional Encodings.
    Results are from 5-fold cross validation, run 20 times with different seeds. 
    \best{Red}: the best model, \good{Violet}: good models. The dimension of node positional encoding with Laplacian eigenvectors is 20.}
    \label{tab:csl}
\end{table}

\subsection{Cycle Detection with CYCLES dataset}
\label{sec:cycles_data_main}
The CYCLES is a dataset synthetically generated by \cite{Loukas2020What} which contains equal number of graphs with and without cycles of fixed lengths. The task is a binary classification task to detect whether a graph has cycle or not. Though there are several forms of the dataset used in \cite{Loukas2020What} in terms of the number of nodes and cycle lengths, we select the dataset variant marked with having node size 56 and cycle length 6, based on the difficulty results shown by the author. The graphs have nodes in the range 37-65.

\noindent {\bf Splitting.} We use the same dataset splits as in \cite{Loukas2020What}. Originally there 10,000 graphs each in the training and test sets. We sample 1,000 class balanced graphs from the training set to be used as validation samples. Therefore, the resulting CYCLES dataset has 9,000 train/ 1,000 validation/10,000 test graphs with all the sets having class-balanced samples. We show results on different sizes of training samples following the original author of CYCLES dataset.\\
{\bf Training.} For the learning rate strategy, an initial learning rate is set to $1\times10^{-4}$, the reduce factor is $0.5$, the patience value is $10$, and the stopping learning rate is $1\times10^{-6}$. Following \cite{Loukas2020What}, we train using a varying sample size from $200$ to $5,000$ out of the training graphs and report the results accordingly. The reported results are based on 4 runs with 4 different seeds.\\
{\bf Performance Measure.} The classification accuracy between the predicted and groundtruth label for whether a graph has cycle or not is the performance measure.\\
{\bf Results.} Similar to the CSL dataset (Section \ref{sec:csl_data_main}), we use the CYCLES dataset mainly for the validation of the Graph Positional Encodings (Section \ref{sec:lap_pe}) proposed as an outcome of this benchmarking framework. As such, we train only a subset of MP-GCNs (GINs and GatedGCNs) and report the respective results. The numerical results are reported in Table \ref{tab:cycles} and analyzed in Section \ref{sec:lap_pe}.

\begin{table}[h]
    \centering
    \scalebox{0.78}{
    \begin{tabular}{r|c|c|c|c|c|c}
        \toprule
        \multicolumn{3}{r|}{\textbf{Train samples $\rightarrow$}} &\textbf{200} & \textbf{500} & \textbf{1000} & \textbf{5000} \\
        \midrule
        \textbf{Model} & \textbf{$L$} & \textbf{\#Param} & \multicolumn{4}{c}{\textbf{Test Acc$\pm$s.d.}}\\
        \midrule
        GIN & 4 & 100774 & 70.585$\pm$0.636 & 74.995$\pm$1.226 & 78.083$\pm$1.083 & 86.130$\pm$1.140\\
        GIN-PE & 4 & 102864 & \textbf{86.720$\pm$3.376} & \textbf{95.960$\pm$0.393} & \textbf{97.998$\pm$0.300} & \textbf{99.570$\pm$0.089}\\
        \midrule
        GatedGCN & 4 & 103933 & 50.000$\pm$0.000 & 50.000$\pm$0.000 & 50.000$\pm$0.000 & 50.000$\pm$0.000\\
        GatedGCN-PE & 4 & 105263 & \textbf{95.082$\pm$0.346} & \textbf{96.700$\pm$0.381} & \textbf{98.230$\pm$0.473} & \textbf{99.725$\pm$0.027}\\
        \bottomrule
        
    \end{tabular}
    }
    \caption{Test accuracy on the CYCLES dataset. Results (higher is better) are averaged over 4 runs with 4 different seeds. The performance on test sets with models trained on varying train data size is show, following \cite{vignac2020building}. \textbf{Bold} shows the best result out of a GNN's two model instances that use and not use PE. The dimension for PE is 20.}
    \label{tab:cycles}
\end{table}

\subsection{Multi-task graph properties with GraphTheoryProp dataset}
\label{sec:graphtheoryprop_data_main}

\cite{corso2020principal} proposed a synthetic dataset of undirected and unweighted graphs of diverse types randomly generated for a multi-task benchmarking of 6 graph-theoretic properties, 3 at the node-level and 3 at the graph-level. We call this dataset as GraphTheoryProp. The node-level tasks are to determine single source shortest paths (Dist.), node eccentricity (Ecc.), and Laplacian features $LX$ given a node feature vector $X$ (Lap.) The graph-level tasks are graph connectivity (Conn.), diameter (Diam.) and spectral radius (Rad.). The dataset has graph sizes in the range of 15-24 nodes which have random identifiers as input features. This dataset is crucial to benchmark the robustness of a GNN to predict specific or overall of all the 6 properties, as these may share subroutines such as graph traversals, despite the tasks being different graph properties \citep{corso2020principal}.

\begin{table}[h]
    \centering
    \scalebox{0.75}{
    \begin{tabular}{r|c|c|cccccc}
\toprule 
\multirow{2}{*}{\textbf{Model}} & \multirow{2}{*}{\textbf{$L$}} & \multicolumn{7}{c}{\textbf{Test}} \\
\cline{3-9}
& & \textbf{Average} & \textbf{Dist.} & \textbf{Ecc.} & \textbf{Lap.} & \textbf{Conn.} & \textbf{Diam.} & \textbf{Rad.} \\
\midrule
GIN & 8 &-3.19$\pm$0.11 & -2.81$\pm$0.11 & -2.42$\pm$0.09 & \textbf{-4.39$\pm$0.18} & \textbf{-2.07$\pm$0.13} & -3.06$\pm$0.11 & \textbf{-4.39$\pm$0.13}\\
GIN-PE & 8 & \textbf{-3.21$\pm$0.13} & \textbf{-2.87$\pm$0.03} & \textbf{-2.83$\pm$0.07} & -3.99$\pm$0.04 & -2.00$\pm$0.15 & \textbf{-3.27$\pm$0.07} & -4.31$\pm$0.15\\
\midrule
GatedGCN & 8 & -3.22$\pm$0.13 & -2.76$\pm$0.17 & -2.36$\pm$0.12 & -3.92$\pm$0.15 & \textbf{-2.65$\pm$0.11} & -3.35$\pm$0.16 & -4.31$\pm$0.08\\
GatedGCN-PE & 8 & \textbf{-3.51$\pm$0.11} & \textbf{-3.23$\pm$0.08} & \textbf{-3.35$\pm$0.08} & \textbf{-4.03$\pm$0.21} & -2.60$\pm$0.12 & \textbf{-3.57$\pm$0.05} & -4.32$\pm$0.13\\

\bottomrule
\end{tabular}
    }
    \caption{Mean ${Log}_{10}$MSE for each task over 4 runs with 4 different seeds. Average denotes the combined average of all the tasks. ${Log}_{10}$MSE is on the test set (lower is better). \textbf{Bold} shows the best result out of a GNN's two model instances that use and not use PE. The dimension for PE is 12.}
    \label{tab:graphtheoryprop}
\end{table}

\noindent {\bf Splitting.} We use the same splitting sets as in \cite{corso2020principal} which has 5,120 train, 640 validation, 1,280 test graphs.\\
{\bf Training.} For the learning rate strategy, an initial learning rate is set to $1\times10^{-3}$, the reduce factor is $0.5$, the patience value is $15$, and the stopping learning rate is $1\times10^{-6}$. The reported results are based on 4 runs with 4 different seeds.\\
{\bf Performance Measure.} For performance measure, ${Log}_{10}$MSE is reported between the predicted and groundtruth values for each single task. Besides, an average performance measure is reported which is the combined average of all the 6 tasks.\\
{\bf Results.} As with the CSL and CYCLES datasets (Sections \ref{sec:csl_data_main}, \ref{sec:cycles_data_main}), we use GraphTheoryProp in this paper for the validation of Graph Positional Encodings, Section \ref{sec:lap_pe}. The numerical results are reported in Table \ref{tab:graphtheoryprop} and analyzed in Section \ref{sec:lap_pe}.


\section{Analysis and Discussion of Benchmarking Results}
\label{sec:eval}

This section highlights the main take-home messages from the experiments in Section \ref{sec:expsetup} on the datasets in the proposed framework, which evaluate the GNNs from Section \ref{sec:gnns} with the experimental setup described in Section \ref{sec:expsetup} and respective sub-sections of each datasets.

{\bf Graph-agnostic NNs perform poorly.} As a sanity check, we compare all GNNs to a simple graph-agnostic MLP baseline which updates each node independent of one-other, $h^{\ell+1}_{i} =  \sigma \left( W^{\ell} \ h^{\ell}_{i} \right)$, and passes 
these features to the task-based layer. MLP presents consistently low scores across all datasets (Tables \ref{tab:results_zinc}-\ref{tab:csl}), 
which shows the necessity to use graph structure for these tasks. All proposed datasets used in our study are appropriate to statistically separate GNN performance, which has remained an issue with the widely used but small graph datasets \citep{errica2019fair,luzhnica2019graph}.

{\bf GCNs outperform WL-GNNs on the proposed datasets.} 
Although provably powerful in terms of graph isomorphism tests and invariant function approximation \citep{maron2019universality,chen2019equivalence,morris2019weisfeiler}, the recent 3WLGNNs and RingGNNs were not able to outperform GCNs for our medium-scale datasets, as shown in Tables \ref{tab:results_zinc}-\ref{tab:results_collab} and \ref{tab:results_superpixels}-\ref{tab:results_tsp}. These new models are limited in terms of space/time complexities, with $O(n^2)/O(n^3)$ respectively, not allowing them to scale to larger datasets. On the contrary, GCNs with linear complexity \textit{w.r.t.} the number of nodes for sparse graphs, can scale conveniently to 16 layers and show the best performance on all datasets. 
3WL-GNNs and RingGNNs face loss divergence and/or out-of-memory errors when trying to build deeper networks.

{\bf Anisotropic mechanisms improve GCNs.} Among the models in the GCN class, the best results point towards the anisotropic models, particularly GAT and GatedGCN, which are based on sparse and dense attention mechanisms, respectively. 
For instance, results for ZINC, AQSOL, WikiCS, MNIST, CIFAR10, PATTERN and CLUSTER in respective Tables \ref{tab:results_zinc}, \ref{tab:results_aqsol}, \ref{tab:results_wikics}, \ref{tab:results_superpixels}, \ref{tab:results_sbm}
show that the performance of the 100K-parameter anisotropic GNNs (GCN with symmetric normalization, GAT, MoNet, GatedGCN) are consistently better than the isotropic models (\textit{vanilla} GCN, GraphSage), except for \textit{vanilla} GCN-WikiCS, GraphSage-MNIST and MoNet-CIFAR10.
Table \ref{tab:edge-analysis}, discussed later, dissects and demonstrates the importance of anisotropy for the link prediction tasks, TSP and COLLAB. 
Overall, our results suggest that understanding the expressive power of attention-based neighborhood aggregation functions is a meaningful avenue of research.

{\bf Underlying challenges for training WL-GNNs.} 
We consistently observe a relatively high standard deviation in the performance of WL-GNNs (recall that we average across 4 runs using 4 different seeds). 
We attribute this fluctuation to the absence of universal training procedures like batching and batch normalization, as these GNNs operate on {\it dense} rank-2 tensors of variable sizes. 
On the other hand, GCNs running on {\it sparse} tensors better leverage batched training and normalization for stable and fast training.
Leading graph machine learning libraries represent batches of graphs as sparse block diagonal matrices, enabling batched training of GCNs through parallelized computation \citep{jia2019redundancy}.

\vspace{-0.1cm}
Dense tensors are incompatible with the prevalent approach, disabling the use of batch normalization for WL-GNNs.
We experimented with layer normalization \citep{ba2016layer} but without success.
We were also unable to train WL-GNNs on CPU memory for the single COLLAB graph, see Table \ref{tab:results_collab}.
Practical applications of the new WL-GNNs may require redesigning the best practices and common building blocks of deep learning, \textit{i.e.} batching of variable-sized data, normalization schemes, and residual connections.

{\bf 3WL-GNNs perform the best among their class.} Among the models in the WL-GNN class, 3WL-GNN provide better results than its similar counter-part RingGNN and achieves close to the best performance for AQSOL, see Table \ref{tab:results_aqsol}. The GIN model, while being less expressive, is able to scale better and provides overall good performance.

\section{Studies using the Benchmarking Framework}
One of the primary goals of this benchmarking framework is to facilitate researchers to perform new explorations conveniently and develop insights that improve our overall understanding of graph neural networks. This section provides a demonstration of two such studies that we carry out by leveraging the datasets and the coding infrastructure which are part of this framework. First, we explore the absence of positional information in graphs for MP-GCNs which induces their low representation power. As a result, we develop a new insight that Laplacian eigenvectors can very simply be used as graph positional encodings and improve MP-GCNs. This insight has been received keenly in the recent literature and there are a number of works that propose positional encoding schemes with some addressing the challenges of using Laplacian eigenvectors \citep{kreuzer2021rethinking, wang2022equivariant, lim2022sign}. Second, we study and show how the modification of existing MP-GCNs with joint edge representations help the models perform comparatively better than their vanilla counterparts.

\label{sec:studies_detail}

\subsection{Laplacian Positional Encodings}
\label{sec:lap_pe}
As discussed in Section \ref{sec:eval}, MP-GCNs outperforms WL-GNNs on the diverse collection of datasets included in our proposed benchmark despite having theoretical limitations derived from the alignment of MP-GCNs to the WL-tests. Also, WL-GNNs were found to be computationally infeasible on medium and large scale datasets. Motivated by these results, we propose `Graph Positional Encodings' using Laplacian eigenvectors, thus referred as Laplacian Positional Encodings, to improve the theoretical shortcomings of MP-GCNs, which allows us to retain the computationally efficiency offered by the message-passing framework and improve the MP-GCNs performance.

\subsubsection{Related Work}
In \cite{murphy2019relational,srinivasan2019equivalence}, it was pointed out that standard MP-GCNs might perform poorly when dealing with graphs that exhibit some symmetries in their structures, such as node or edge isomorphism. This is related to the limitation of MP-GCNs due to their equivalence to the 1-WL test \citep{xu2018how, morris2019weisfeiler}. The equivalence is based on the condition when MP-GCNs handle anonymous nodes \citep{Loukas2020What}, i.e. nodes do not have unique node features. To address this issue of anonymous MP-GCNs, \cite{murphy2019relational} introduced a framework, called Graph Relational Pooling (GRP), that assigns to each node an identifier that depends on the index ordering. This approach can be computationally expensive as it requires to account for all $n!$ node permutations, thus requiring some sampling in practice. 
\cite{you2019position} proposed learnable position-aware embeddings based on random anchor sets of nodes for pairwise node (or, link) tasks. However, the random selection of anchor sets has limitations and their approach is not applicable on inductive node tasks. Similarly, one could think of using full or partial random node identifiers for breaking node-anonymity. Yet, it suffers from generalization to unseen graphs \citep{you2019position, Loukas2020What}. \cite{li2020distance} proposed the use of distance encoding as node attributes which captures distances between nodes using power(s) of random walk matrix. However, their failure on distance regular graphs \citep{li2020distance} and the cost of computing the power matrices may be limiting to scale to diverse and medium to large-scale graphs. We improve upon these works and propose the use of Laplacian eigenvectors as positional encodings.

\subsubsection{Laplacian eigenvectors as Positional Encodings}
We keep the overall MP-GCN architecture and simply add positional features to each node before processing the graph through the MP-GCN. Intuitively, the positional features should be chosen such that nodes which are far apart in the graph have different positional features whereas nodes which are nearby have similar positional features. 
As node positional features, we propose to use graph Laplacian eigenvectors \citep{belkin2003laplacian}, which have less ambiguities and which better describe the distance between nodes on the graph. Formally, Laplacian eigenvectors are spectral techniques that embed the graphs into the Euclidean space. These vectors form a meaningful local coordinate system, while preserving the global graph structure. Mathematically, they are defined via the factorization of the graph Laplacian matrix;
\vspace{-0.1cm}
\begin{align}
\Delta=\textrm{I}-D^{-1/2}AD^{-1/2}=U^T\Lambda U,
 \label{LapEig}
\end{align}
where $A$ is the $n\times n$  adjacency matrix, $D$ is the degree matrix, and $\Lambda$, $U$ correspond respectively to the eigenvalues and eigenvectors. Laplacian eigenvectors also represent a natural generalization of the Transformer \citep{vaswani2017attention} positional encodings (PE) for graphs as the eigenvectors of a discrete line (NLP graph) are the cosine and sinusoidal functions. The computational complexity $O(E^{3/2})$, with $E$ being the number of edges, can be improved with, \textit{e.g.} the Nystrom method \citep{fowlkes2004spectral}. The eigenvectors are defined up to the factor $\pm 1$ (after being normalized to unit length), so the sign of eigenvectors will be randomly flipped during training. For the experiments, we use the $k$ smallest non-trivial eigenvectors, where the $k$ value is given in the respective experiment tables in Section \ref{sec:expsetup} as the dimensions of the PE. The smallest eigenvectors provide smooth encoding coordinates of neighboring nodes. See Section \ref{sec_PE} for additional discussion about positional encodings and the reasoning behind our decision to use random sign flipping.

\subsubsection{Experiments and Analysis}
\label{sec:lap_pe_experiments}
We first use the mathematical graphs such as CSL, CYCLES and GraphTheoryProp included in our benchmark (Sections \ref{sec:csl_data_main}-\ref{sec:graphtheoryprop_data_main}) to validate the proposed Laplacian PE as simple augmentations in MP-GCNs to improve their performance on the datasets. 
On CSL dataset, Table \ref{tab:csl} compares the MP-GCNs using the Laplacian eigenvectors as PE and the WL-GNNs. The MP-GCN models were the most accurate with 99\% of mean accuracy, while 3WL-GNN obtained 97\% and RingGNN 25\% with our experimental setting. Similarly, in Table \ref{tab:cycles} for CYCLES dataset and Table \ref{tab:graphtheoryprop} for GraphTheoryProp dataset, where we simply select 2 representative MP-GCNs (GINs and GatedGCNs), we observe a consistent improvement in the performance when GINs and GatedGCNs are augmented with Laplacian PE. This demonstrates the importance of positional features to successfully detect cycles in a graph, and also predict critical theoretical and geometric properties in a graph.

Next, we study ZINC, AQSOL, WikiCS, PATTERN, CLUSTER and COLLAB with PE (note that MNIST, CIFAR10 and TSP do not need PE as the nodes in these graphs already have features describing their positions in $\mathbb R^2$). We observe a boost of performance for ZINC, AQSOL and CLUSTER (it was expected as eigenvectors are good indicators of clusters \citep{von2007tutorial}), an improvement for PATTERN, and statistically the same result for COLLAB, see the respective tables in Section \ref{sec:expsetup}. This way, MP-GCNs can be augmented with Laplacian PE to overcome their limitations of not being able to detect simple graph symmetries. 
Additionally, PEs also boost the models' performance on real-world graph learning tasks.

\subsubsection{Challenges with using Laplacian eigenvectors}
\label{sec_PE}

\begin{table}[h!]
    \centering
    \scalebox{0.75}{
    \begin{tabular}{c|rcccccc}
        \toprule
        \multicolumn{1}{c}{} & \textbf{PE type} & \textbf{$L$} & \textbf{\#Param} & \textbf{Test Acc.$\pm$s.d.} & \textbf{Train Acc.$\pm$s.d.} & \textbf{\#Epochs} & \textbf{Epoch/Total} \\
        \midrule
        \parbox[t]{2mm}{\multirow{6}{*}{\rotatebox[origin=c]{90}{CSL}}} & No PE & 4 & 104007 & 10.000$\pm$0.000 & 10.000$\pm$0.000 & 54.00 & 0.58s/0.05hr \\
         & EigVecs-20 & 4 & 105407 & 68.633$\pm$7.143 & 99.811$\pm$0.232 & 107.16 & 0.59s/0.09hr \\
         & Rand sign(EigVecs) & 4 & 105407 & \best{99.767$\pm$0.394} & 99.689$\pm$0.550 & 188.76 & 0.59s/0.16hr \\
         & Abs(EigVecs) & 4 & 105407 & 99.433$\pm$1.133 & 100.000$\pm$0.000 & 143.64 & 0.60s/0.12hr \\
         & Fixed node ordering & 4 & 106807 & 10.533$\pm$4.469 & 76.056$\pm$14.136 & 60.56 & 0.59s/0.05hr \\
         & Rand node ordering & 4 & 106807 & 11.133$\pm$2.571 & 10.944$\pm$2.106 & 91.60 & 0.60s/0.08hr \\
        \midrule
        \parbox[t]{2mm}{\multirow{6}{*}{\rotatebox[origin=c]{90}{PATTERN}}} & No PE & 16 & 502223 & 85.605$\pm$0.105 & 85.999$\pm$0.145 & 62.00 & 646.03s/11.36hr \\
         & EigVecs-2 & 16 & 505421 & 86.029$\pm$0.085 & 86.955$\pm$0.227 & 65.00 & 645.36s/11.94hr \\
         & Rand sign(EigVecs) & 16 & 502457 & \best{86.508$\pm$0.085} & 86.801$\pm$0.133 & 65.75 & 647.94s/12.08hr\\
         & Abs(EigVecs) & 16 & 505421 & 86.393$\pm$0.037 & 87.011$\pm$0.172 & 62.00 & 645.90s/11.41hr \\
         & Fixed node ordering & 16 & 516887 & 80.133$\pm$0.202 & 98.416$\pm$0.141 & 45.00 & 643.23s/8.27hr \\
         & Rand node ordering & 16 & 516887 & 85.767$\pm$0.044 & 85.998$\pm$0.063 & 64.50 & 645.09s/11.79hr \\
        \midrule
        \parbox[t]{2mm}{\multirow{6}{*}{\rotatebox[origin=c]{90}{CLUSTER}}} & No PE & 16 & 502615 & 73.684$\pm$0.348 & 88.356$\pm$1.577 & 61.50 & 399.44s/6.97hr \\
         & EigVecs-20 & 16 & 504253 & 75.520$\pm$0.395 & 89.332$\pm$1.297 & 49.75 & 400.50s/5.70hr \\
         & Rand sign(EigVecs) & 16 & 504253 & \best{76.082$\pm$0.196} & 88.919$\pm$0.720 & 57.75 & 399.66s/6.58hr \\
         & Abs(EigVecs) &  16 & 504253 & 73.796$\pm$0.234 & 91.125$\pm$1.248 & 58.75 & 398.97s/6.68hr \\
         & Fixed node ordering & 16 & 517435 & 69.232$\pm$0.265 & 92.298$\pm$0.712 & 51.00 & 400.40s/5.82hr \\
         & Rand node ordering & 16 & 517435 & 74.656$\pm$0.314 & 82.940$\pm$1.718 & 61.00 & 397.75s/6.88hr \\
         \midrule
        \parbox[t]{2mm}{\multirow{4}{*}{\rotatebox[origin=c]{90}{COLLAB}}} & No PE & 3 & 40965 & 52.635$\pm$1.168 & 96.103$\pm$1.876 & 95.00 & 453.47s/12.09hr \\
         & EigVecs-20 & 3 & 41889 & 52.326$\pm$0.678 & 96.700$\pm$1.296 & 95.00 & 452.40s/12.10hr \\
         & Rand sign(EigVecs) & 3 & 41889 & \best{52.849$\pm$1.345} & 96.165$\pm$0.453 & 94.75 & 452.75s/12.08hr \\
         & Abs(EigVecs) & 3 & 41889 & 51.419$\pm$1.109 & 95.984$\pm$1.157 & 95.00 & 451.36s/12.07hr \\
        \midrule
        \multicolumn{1}{c}{} & \textbf{PE type} & \textbf{$L$} & \textbf{\#Param} & \textbf{Test MAE$\pm$s.d.} & \textbf{Train MAE$\pm$s.d.} & \textbf{\#Epochs} & \textbf{Epoch/Total} \\
        \midrule
        \parbox[t]{2mm}{\multirow{6}{*}{\rotatebox[origin=c]{90}{ZINC}}} & No PE & 16 & 504153 & 0.354$\pm$0.012 & 0.095$\pm$0.012 & 165.25 & 10.52s/0.49hr \\
         & EigVecs-8 & 16 & 505011 & 0.319$\pm$0.010 & 0.038$\pm$0.007 & 143.25 & 10.62s/0.43hr \\
         & Rand sign(EigVecs) & 16 & 505011 & \best{0.214$\pm$0.013} & 0.067$\pm$0.019 & 185.00 & 10.70s/0.56hr \\
         & Abs(EigVecs) & 16 & 505011 & \best{0.214$\pm$0.009} & 0.035$\pm$0.011 & 167.50 & 10.61s/0.50hr \\
         & Fixed node ordering & 16 & 507195 & 0.431$\pm$0.007 & 0.044$\pm$0.009 & 118.25 & 10.62s/0.35hr \\
         & Rand node ordering & 16 & 507195 & 0.321$\pm$0.015 & 0.177$\pm$0.015 & 184.75 & 10.55s/0.55hr \\
        \bottomrule
    \end{tabular}
    }
    \caption{Study of positional encodings (PEs) with the GatedGCN model \citep{bresson2017residual}. Performance reported on the test sets of CSL, ZINC, PATTERN, CLUSTER and COLLAB (higher is better, except for ZINC). \best{Red}: the best model.  
    }
    \label{tab:PE_study}
\end{table}

Ideally, positional encodings (PEs) should be unique for each node, and nodes which are far apart in the graph should have different positional features whereas nodes which are nearby have similar positional features. Note that in a graph that has some symmetries, positional features cannot be assigned in a canonical way. For example, if node $i$ and node $j$ are structurally symmetric, and we have positional features  $p_i = a$,  $p_j = b$ that differentiate them, then it is also possible to arbitrary choose $p_i = b$,  $p_j = a$ since $i$ and $j$ are completely symmetric by definition. In other words, the PE is always arbitrary up to the number of symmetries in the graph. As a consequence, the network will have to learn to deal with these ambiguities during training. The simplest possible positional encodings is to give an (arbitrary) ordering to the nodes, among $n!$ possible orderings. During training, the orderings are uniformly sampled from the $n!$ possible choices in order for the network to learn to be independent to these arbitrary choices \citep{murphy2019relational}.

We propose an alternative to reduce the sampling space, and therefore the amount of ambiguities to be resolved by the network. Laplacian eigenvectors are hybrid positional and structural encodings, as they are invariant by node re-parametrization. However, they are also limited by natural symmetries such as the arbitrary sign of eigenvectors (after being normalized to have unit length). The number of possible sign flips is $2^k$, where $k$ is the number of eigenvectors. In practice we choose $k\ll n$, and therefore $2^k$ is much smaller $n!$ (the number of possible ordering of the nodes). During the training, the eigenvectors will be uniformly sampled at random between the $2^k$ possibilities. If we do not seek to learn the invariance w.r.t. all possible sign flips of eigenvectors, then we can remove the sign ambiguity of eigenvectors by taking the absolute value. This choice seriously degrades the expressivity power of the positional features.

Numerical results for different positional encodings are reported in Table \ref{tab:PE_study}. For all results, we use the GatedGCN model \citep{bresson2017residual}. We study 5 types of positional encodings; {\it EigVecs-$k$} corresponds to the smallest non-trivial $k$ eigenvectors, {\it Rand sign(EigVecs)} randomly flips the sign of the $k$ smallest non-trivial eigenvectors in each batch, {\it Abs(EigVecs)} takes the absolute value of the $k$ eigenvectors, {\it Fixed node ordering} uses the original node ordering of  graphs, and {\it Rand node ordering} randomly permutes ordering of nodes in each batch. We observed that the best results are consistently produced with the Laplacian PEs with random sign flipping at training. For index PEs, randomly permuting the ordering of nodes also improves significantly the performances over keeping fixed the original node ordering. However, Laplacian PEs clearly outperform index PEs.

\subsection{Edge representations for link prediction.}
\label{sec:edge_repr}

\subsubsection{With GatedGCN and GAT}

The TSP and COLLAB edge classification tasks present an interesting empirical result for GCNs: Isotropic models (\textit{vanilla} GCN, GraphSage) are consistently outperformed by their Anisotropic counterparts which use joint representations of adjacent nodes as edge features during aggregation (GAT, GatedGCN). In this section, 
we systematically study the impact of anisotropy by instantiating three variants of GAT and GatedGCN:\\
(1) Isotropic aggregation (such as \textit{vanilla} 
GCNs \citep{kipf2017semi}) with node updates of the form:
\vspace{-0.1cm}
\begin{align}
h^{\ell+1}_i=\sigma\big(\sum_{j\in\mathcal{N}_i} W^\ell h^\ell_j \big), \quad \textrm{ identified by (\textit{E.Feat,E.Repr}=\crossmark,\crossmark) in Table~\ref{tab:edge-analysis};}
 \label{eq:edge1}
\end{align}
(2) Anisotropy using edge features (such as GAT by default \citep{velickovic2018graph}) with node updates as:
\vspace{-0.1cm}
\begin{align}
h^{\ell+1}_i=\sigma\big(\sum_{j\in\mathcal{N}_i} f_{V^\ell}(h_i^\ell,h_j^\ell) \cdot W^\ell h^\ell_j \big), \quad \textrm{with (\textit{E.Feat,E.Repr}=\checkmark,\crossmark);}
 \label{eq:edge2}
\end{align}
and (3) Anisotropy with edge features and explicit edge representations updated at each layer with node/edge updates as (such as in GatedGCN by default \citep{bresson2017residual}):
\vspace{-0.1cm}
\begin{align}
h^{\ell+1}_i=\sigma\big(\sum_{j\in\mathcal{N}_i} e_{ij}^\ell \cdot W^\ell h^\ell_j \big), \ e_{ij}^{\ell+1}=f_{V^\ell}\big(h_i^\ell,  h_j^\ell, e_{ij}^\ell), \quad \textrm{with (\textit{E.Feat,E.Repr}=\checkmark,\checkmark).}
 \label{eq:edge3}
\end{align}

The formal update equations of the three variants of \textbf{GatedGCN} are:

\noindent \textbf{Isotropic}, similar to vanilla GCNs with sum aggregation:
\begin{equation}
    \label{eqn:gated-gcn-iso}
    h_{i}^{\ell+1} = h_{i}^{\ell} + \text{ReLU} \Big( \text{BN} \Big( U^{\ell} h_{i}^{\ell} + \sum_{j \in \mathcal{N}_i} V^{\ell} h_{j}^{\ell} \Big)\Big), \quad \text{where } U^{\ell}, V^{\ell} \in \mathbb{R}^{d \times d}.
\end{equation}

\noindent \textbf{Anisotropic} with intermediate edge features computed as joint representations of adjacent node features at each layer:
\begin{eqnarray}
    h_{i}^{\ell+1} &=& h_{i}^{\ell} + \text{ReLU} \Big( \text{BN} \Big( U^{\ell} h_{i}^{\ell} + \sum_{j \in \mathcal{N}_i} e_{ij}^{\ell} \odot V^{\ell} h_{j}^{\ell} \Big)\Big), \label{eqn:gated-gcn-h-edgefeat} \\
    e_{ij}^{\ell} &=& \frac{\sigma(\hat e_{ij}^{\ell})}{\sum_{j' \in \mathcal{N}_i} \sigma(\hat e_{ij'}^{\ell}) + \varepsilon }, \label{eqn:gated-gcn-eta-edgefeat} \quad \hat e_{ij}^{\ell} = A^{\ell} h_{i}^{\ell-1} + B^{\ell} h_{j}^{\ell-1}, \label{eqn:gated-gcn-e-edgefeat}
\end{eqnarray}
where $U^{\ell}, V^{\ell} \in \mathbb{R}^{d \times d}$, $\odot$ is the Hadamard product, and $e_{ij}^{\ell}$ are the edge gates.
    
\noindent \textbf{Anisotropic} with edge features as well as explicit edge representations updated across layers in addition to node features, as in GatedGCN by default, Eq.\eqref{eqn:gated-gcn-h}:
\begin{eqnarray}
    h_{i}^{\ell+1} &=& h_{i}^{\ell} + \text{ReLU} \Big( \text{BN} \Big( U^{\ell} h_{i}^{\ell} + \sum_{j \in \mathcal{N}_i} e_{ij}^{\ell} \odot V^{\ell} h_{j}^{\ell} \Big)\Big), \label{eqn:gated-gcn-edgereprfeat-h} \\
    e_{ij}^{\ell} &=& \frac{\sigma(\hat e_{ij}^{\ell})}{\sum_{j' \in \mathcal{N}_i} \sigma(\hat e_{ij'}^{\ell}) + \varepsilon }, \label{eqn:gated-gcn-edgereprfeat-eta} \\
    \hat e_{ij}^{\ell} &=& \hat e_{ij}^{\ell-1} + \text{ReLU} \Big( \text{BN} \big( A^{\ell} h_{i}^{\ell-1} + B^{\ell} h_{j}^{\ell-1} + C^{\ell} \hat e_{ij}^{\ell-1} \big) \Big),  \label{eqn:gated-gcn-edgereprfeat-e}   
\end{eqnarray}
where $U^{\ell}, V^{\ell} \in \mathbb{R}^{d \times d}$, $\odot$ is the Hadamard product, and $e_{ij}^{\ell}$ are the edge gates.
The input edge features from the datasets (\textit{e.g.} distances for TSP, collaboration year and frequency for COLLAB) can optionally be used to initialize the edge representations $\hat e_{ij}^{\ell=0}$. Note that there may be a multitude of approaches to instantiating anisotropic GNNs and using edge features \citep{battaglia2016interaction,sanchez2018graph,brockschmidt2019gnn} besides the ones we consider.\\

The formal update equations of the three variants of \textbf{GAT} are:

\noindent \textbf{Isotropic}, similar to multi-headed vanilla GCNs with sum aggregation:
\begin{equation}
    \label{eqn:gat-iso}
    h_{i}^{\ell+1} =  \text{Concat}_{k=1}^{K} \Big( \text{ELU} \Big( \text{BN} \Big( \sum_{j \in \mathcal{N}_i} U^{k,\ell} \ h_{j}^{\ell} \Big) \Big) \Big), \quad \text{where } U^{k,\ell} \in \mathbb{R}^{\frac{d}{K} \times d}.
\end{equation}

\noindent \textbf{Anisotropic} with intermediate edge features computed as joint representations of adjacent node features at each layer, as in GAT by default, Eq.\eqref{eqn:gat}:
\begin{eqnarray}
    h_{i}^{\ell+1} &=& h_{i}^{\ell} + \text{ELU} \Big( \text{BN} \Big( \text{Concat}_{k=1}^{K} \Big( \sum_{j \in \mathcal{N}_i} e_{ij}^{k,\ell} \ U^{k,\ell} \ h_{j}^{\ell} \Big) \Big) \Big), \label{eqn:gat-edgefeat} \\
    e_{ij}^{k,\ell} &=& \frac{\exp(\hat e_{ij}^{k,\ell})}{\sum_{j' \in \mathcal{N}_i} \exp(\hat e_{ij'}^{k,\ell}) }, \quad \hat e_{ij}^{k,\ell} = \text{LeakyReLU} \Big( V^{k,\ell} \ \text{Concat} \big( U^{k,\ell} h_{i}^{\ell} , \ U^{k,\ell} h_{j}^{\ell} \big) \Big), \label{eqn:gat-edgefeat-attn}
\end{eqnarray}
where $U^{k,\ell} \in \mathbb{R}^{\frac{d}{K} \times d}, V^{k,\ell} \in \mathbb{R}^{\frac{2d}{K}}$ are the $K$ linear projection heads and $e_{ij}^{k,\ell}$ are the attention coefficients for each head.

\noindent \textbf{Anisotropic} with edge features as well as explicit edge representations updated across layers in addition to node features:
\begin{eqnarray}
    \label{gat-e}
    h_{i}^{\ell+1} &=& h_{i}^{\ell} + \text{ELU} \Big( \text{BN} \Big( \text{Concat}_{k=1}^{K} \Big( \sum_{j \in \mathcal{N}_i} a_{ij}^{k,\ell} \ U^{k,\ell} \ h_{j}^{\ell} \Big) \Big) \Big), \\
    \label{gat-e-feat}
    e_{ij}^{\ell+1} &=& e_{ij}^{\ell} + \text{ELU} \Big( \text{BN} \Big(  \text{Concat}_{k=1}^{K} \Big( B^{k,\ell} \ \text{Concat} \big( A^{k,\ell} e_{ij}^{\ell}, \  U^{k,\ell} h_{i}^{\ell} , \ U^{k,\ell} h_{j}^{\ell} \big) \Big) \Big) \Big), \\
    \label{gat-e-attn}
    a_{ij}^{k,\ell} &=& \frac{\exp(\hat a_{ij}^{k,\ell})}{\sum_{j' \in \mathcal{N}_i} \exp(\hat a_{ij'}^{k,\ell}) }, \\
    \label{gat-e-attn2}
    \hat a_{ij}^{k,\ell} &=& \text{LeakyReLU} \Big( V^{k,\ell} \ \text{Concat} \big( A^{k,\ell} e_{ij}^{\ell}, \  U^{k,\ell} h_{i}^{\ell} , \ U^{k,\ell} h_{j}^{\ell} \big) \Big),
\end{eqnarray}
where $U^{k,\ell} \in \mathbb{R}^{\frac{d}{K} \times d}, V^{k,\ell} \in \mathbb{R}^{\frac{3d}{K}}, A^{k,\ell} \in \mathbb{R}^{\frac{d}{K} \times d}, B^{k,\ell} \in \mathbb{R}^{\frac{d}{K} \times \frac{3d}{K}}$ are the $K$ linear projection heads and $a_{ij}^{k,\ell}$ are the attention coefficients for each head.
The input edge features from the datasets can optionally be used to initialize the edge representations $e_{ij}^{\ell=0}$.\\

\begin{table}[t]
    \centering
    \scalebox{0.70}{
    \begin{tabular}{c|rcccccccc}
        \toprule
        \multicolumn{1}{c}{} & \textbf{Model} & \textbf{E.Feat.} & \textbf{E.Repr.} & \textbf{$L$} & \textbf{\#Param} & \textbf{Test Acc.$\pm$s.d.} & \textbf{Train Acc.$\pm$s.d.} & \textbf{\#Epochs} & \textbf{Epoch/Total} \\
        \midrule
        \parbox[t]{2mm}{\multirow{8}{*}{\rotatebox[origin=c]{90}{TSP}}} & \multirow{3}{*}{GatedGCN} & \crossmark & \crossmark & 4 & 99026 & 0.646$\pm$0.002 & 0.648$\pm$0.002 & 197.50 & 150.83s/8.34hr \\
         & & \checkmark & \crossmark & 4 & 98174 & 0.757$\pm$0.009 & 0.760$\pm$0.009 & 218.25 & 197.80s/12.06hr \\
         & & \checkmark & \checkmark & 4 & 97858 & \good{0.791$\pm$0.003} & 0.793$\pm$0.003 & 159.00 & 218.20s/9.72hr \\
        \cline{2-10}
         & GatedGCN-E & \checkmark & \checkmark & 4 & 97858 & \best{0.808$\pm$0.003} & 0.811$\pm$0.003 & 197.00 & 218.51s/12.04hr \\
        \cline{2-10}
         & \multirow{3}{*}{GAT} & \crossmark & \crossmark & 4 & 95462 & 0.643$\pm$0.001 & 0.644$\pm$0.001 & 132.75 & 325.22s/12.10hr \\
         & & \checkmark & \crossmark & 4 & 96182 & 0.671$\pm$0.002 & 0.673$\pm$0.002 & 328.25 & 68.23s/6.25hr \\
         & & \checkmark & \checkmark & 4 & 96762 & 0.748$\pm$0.022 & 0.749$\pm$0.022 & 93.00 & 462.22s/12.10hr \\
        \cline{2-10}
         & GAT-E & \checkmark & \checkmark & 4 & 96762 & \good{0.782$\pm$0.006} & 0.783$\pm$0.006 & 98.00 & 438.37s/12.11hr \\
        \midrule
        \parbox[t]{2mm}{\multirow{8}{*}{\rotatebox[origin=c]{90}{COLLAB}}} & \multirow{3}{*}{GatedGCN} & \crossmark & \crossmark & 3 & 26593 & 35.989$\pm$1.549 & 60.586$\pm$4.251 & 148.00 & 263.62s/10.90h \\
         & & \checkmark & \crossmark & 3 & 26715 & \good{50.668$\pm$0.291} & 96.128$\pm$0.576 & 172.00 & 384.39s/18.44hr \\
         & & \checkmark & \checkmark & 3 & 27055 & \best{51.537$\pm$1.038} & 96.524$\pm$1.704 & 188.67 & 376.67s/19.85hr \\
        \cline{2-10}
         & GatedGCN-E & \checkmark & \checkmark & 3 & 27055 & 47.212$\pm$2.016 & 85.801$\pm$0.984 & 156.67 & 377.04s/16.49hr \\
        \cline{2-10}
         & \multirow{3}{*}{GAT} & \crossmark & \crossmark & 3 & 28201 & 41.141$\pm$0.701 & 70.344$\pm$1.837 & 153.50 & 371.50s/15.97hr \\
         & & \checkmark & \crossmark & 3 & 28561 & \good{50.662$\pm$0.687} & 96.085$\pm$0.499 & 174.50 & 403.52s/19.69hr \\
         & & \checkmark & \checkmark & 3 & 26676 & \good{49.674$\pm$0.105} & 92.665$\pm$0.719 & 201.00 & 349.19s/19.59hr \\
        \cline{2-10}
         & GAT-E & \checkmark & \checkmark & 3 & 26676 & 44.989$\pm$1.395 & 82.230$\pm$4.941 & 120.67 & 328.29s/11.10hr \\
        \bottomrule
    \end{tabular}
    }
    \caption{Study of anisotropy and edge representations for link prediction on TSP and COLLAB. \best{Red}: the best model, \good{Violet}: good models.
    }
    \label{tab:edge-analysis}
\end{table}

\noindent {\bf Numerical Experiments and Analysis}

In Table \ref{tab:edge-analysis}, we show the experiments of the three variants of GatedGCN and GAT on TSP and COLLAB. GatedGCN-E and GAT-E in Table are models using input edge features from the datasets to initialize the edge representations $e_{ij}$. As maintaining edge representations comes with a time and memory cost for the large COLLAB graph, all models use a reduced budget of 27K parameters to fit the GPU memory, and are allowed to train for a maximum of 24 hours for convergence.

On both TSP and COLLAB, upgrading isotropic models with edge features significantly boosts performance given the same model parameters (\textit{e.g.} 0.75 vs. 0.64 F1 score on TSP,  50.6\% vs. 35.9\% Hits@50 on COLLAB for GatedGCN with edge features vs. the isotropic variant). 
Maintaining explicit edge representations across layers further improves F1 score for TSP, especially when initializing the edge representations with euclidean distances between nodes (\textit{e.g.} 0.78 vs. 0.67 F1 score for GAT-E vs. standard GAT).
On COLLAB, adding explicit edge representations and inputs degrades performance, suggesting that the features (collaboration frequency and year) are not useful for the link prediction task (\textit{e.g.} 47.2 vs. 51.5 Hits@50 for GatedGCN-E vs. GatedGCN).
As suggested by \cite{hu2020ogb}, it would be interesting to treat COLLAB as a multi-graph with temporal edges, motivating the development of task-specific anisotropic edge representations beyond generic attention and gating mechanisms.

\subsubsection{With GraphSage}

Interestingly, in Table \ref{tab:results_collab} for COLLAB, we found that the isotropic GraphSage with max aggregation performs close to GAT and GatedGCN models, both of which perform anisotropic mean aggregation.
On the other hand, models which use sum aggregation (GIN, MoNet) are unable to beat the simple matrix factorization baseline.
This result indicates that aggregation functions which are invariant to node degree (max and mean) provide a powerful inductive bias for COLLAB.

We instantiate two anisotropic variants of GraphSage, as described in the following paragraphs, and compare them to GAT and GatedGCN on COLLAB in Table~\ref{tab:edge-analysis-2}.
We find that upgrading max aggregators with edge features does not significantly boost performance. 
On the other hand, maintaining explicit edge representations across layers hurts the models, presumably due to using very small hidden dimensions.
(As previously mentioned, maintaining representations for both 235K nodes and 2.3M edges leads to significant GPU memory usage and requires using smaller hidden dimensions.)

\begin{table}[t!]
\centering
    \scalebox{0.68}{
    \begin{tabular}{rccccccccc}
        \toprule
         \multirow{2}{*}{\textbf{Model}} & \textbf{Edge} & \textbf{Edge} & \textbf{Aggregation} & \multirow{2}{*}{\textbf{$L$}} & \multirow{2}{*}{\textbf{\#Param}} & \textbf{Test Acc.} & \textbf{Train Acc.} & \multirow{2}{*}{\textbf{\#Epoch}} & \textbf{Epoch/} \\
         & \textbf{Feat.} & \textbf{Repr.} & \textbf{Function} & & & \textbf{$\pm$s.d}. & \textbf{$\pm$s.d.} & & \textbf{Total} \\
         \midrule
        \multirow{3}{*}{GatedGCN} & \crossmark & \crossmark & Sum & 3 & 26593 & 35.989$\pm$1.549 & 60.586$\pm$4.251 & 148.00 & 263.62s/10.90h \\
         & \checkmark & \crossmark & Weighted Mean & 3 & 26715 & \good{50.668$\pm$0.291} & 96.128$\pm$0.576 & 172.00 & 384.39s/18.44hr \\
         & \checkmark & \checkmark & Weighted Mean & 3 & 27055 & \best{51.537$\pm$1.038} & 96.524$\pm$1.704 & 188.67 & 376.67s/19.85hr \\
        \midrule
         GatedGCN-E & \checkmark & \checkmark & Weighted Mean & 3 & 27055 & 47.212$\pm$2.016 & 85.801$\pm$0.984 & 156.67 & 377.04s/16.49hr \\
        \midrule
         \multirow{3}{*}{GAT} & \crossmark & \crossmark & Sum & 3 & 28201 & 41.141$\pm$0.701 & 70.344$\pm$1.837 & 153.50 & 371.50s/15.97hr \\
         & \checkmark & \crossmark & Weighted Mean & 3 & 28561 & \good{50.662$\pm$0.687} & 96.085$\pm$0.499 & 174.50 & 403.52s/19.69hr \\
         & \checkmark & \checkmark & Weighted Mean & 3 & 26676 & \good{49.674$\pm$0.105} & 92.665$\pm$0.719 & 201.00 & 349.19s/19.59hr \\
        \midrule
         GAT-E & \checkmark & \checkmark & Weighted Mean & 3 & 26676 & 44.989$\pm$1.395 & 82.230$\pm$4.941 & 120.67 & 328.29s/11.10hr \\
        \midrule
         \multirow{3}{*}{GraphSage} & \crossmark & \crossmark & Max & 3 & 26293 & \good{50.908$\pm$1.122} & 98.617$\pm$1.763 & 157.75 & 241.49s/10.62hr \\
         & \checkmark & \crossmark & Weighted Max & 3 & 26487 & \good{50.997$\pm$0.875} & 99.158$\pm$0.694 & 112.00 & 366.24s/11.46hr \\
         & \checkmark & \checkmark & Weighted Max & 3 & 26950 & 48.530$\pm$1.919 & 90.990$\pm$9.273 & 118.25 & 359.18s/11.88hr \\
        \midrule
         GraphSage-E & \checkmark & \checkmark & Weighted Max & 3 & 26950 & 47.315$\pm$1.939 & 93.475$\pm$5.884 & 120.00 & 359.10s/12.07hr \\
        \bottomrule
    \end{tabular}
    }
    \caption{Study of anisotropic edge features and representations for link prediction on COLLAB, including GraphSage models. \best{Red}: the best model, \good{Violet}: good models.
    }
    \label{tab:edge-analysis-2}
\end{table}


\textbf{Isotropic}, as in GraphSage by default, Eq.\eqref{eqn:graphsage-maxpool}:
\begin{equation}
    \label{eqn:graphsage-maxpool-iso}
    h_{i}^{\ell+1} = h_{i}^{\ell} + \text{ReLU} \Big(  \text{BN} \Big( U^{\ell} \ \text{Concat} \left( h_{i}^{\ell} , \ \text{Max}_{j \in \mathcal{N}_i} \ \text{ReLU} \left( V^{\ell} h_j^{\ell} \right) \right) \Big) \Big),
\end{equation}
where $U^{\ell} \in \mathbb{R}^{d \times 2d}, V^{\ell} \in \mathbb{R}^{d \times d}$.

\textbf{Anisotropic} with intermediate edge features computed as joint representations of adjacent node features at each layer:
\begin{eqnarray}
    \label{eqn:graphsage-maxpool-edgefeat} 
    h_{i}^{\ell+1} &=& h_{i}^{\ell} + \text{ReLU} \Big( \text{BN} \Big( U^{\ell} \ \text{Concat} \left( h_{i}^{\ell} , \ \text{Max}_{j \in \mathcal{N}_i} \ \text{ReLU} \left( \sigma \left( e_{ij}^{\ell} \right) \odot V^{\ell} h_j^{\ell} \right) \right) \Big) \Big), \\
    \label{eqn:graphsage-e-edgefeat}
    e_{ij}^{\ell} &=& A^{\ell} \left( h_{i}^{\ell-1} + h_{j}^{\ell-1} \right),
\end{eqnarray}
where $U^{\ell} \in \mathbb{R}^{d \times 2d}, V^{\ell}, A^{\ell} \in \mathbb{R}^{d \times d}$, $\odot$ is the Hadamard product, and $e_{ij}^{\ell}$ are the edge gates.

\textbf{Anisotropic} with edge features as well as explicit edge representations updated across layers in addition to node features:
\begin{eqnarray}
    \label{eqn:graphsage-maxpool-edgereprfeat}
    h_{i}^{\ell+1} &=& h_{i}^{\ell} + \text{ReLU} \Big( \text{BN} \Big( U^{\ell} \ \text{Concat} \left( h_{i}^{\ell} , \ \text{Max}_{j \in \mathcal{N}_i} \ \text{ReLU} \left( \sigma \left( \hat e_{ij}^{\ell} \right) \odot V^{\ell} h_j^{\ell} \right) \right) \Big) \Big), \\
    \label{eqn:graphsage-e-edgereprfeat}
    \hat e_{ij}^{\ell} &=& A^{\ell} \left( h_{i}^{\ell-1} + h_{j}^{\ell-1} \right) + B^{\ell} e_{ij}^{\ell-1}, \quad
    e_{ij}^{\ell+1} = e_{ij}^{\ell} + \text{ReLU} \Big( \text{BN} \Big( \hat e_{ij}^{\ell} \Big) \Big),
\end{eqnarray}
where $U^{\ell} \in \mathbb{R}^{d \times 2d}, V^{\ell}, A^{\ell}, B^{\ell} \in \mathbb{R}^{d \times d}$, $\odot$ is the Hadamard product, and $\hat e_{ij}^{\ell}$ are the edge gates.
The input edge features from the datasets can optionally be used to initialize the edge representations $e_{ij}^{\ell=0}$.


\section{Experiments on TU datasets}
\label{sec:TUs}

\begin{table}[t]
    \centering
    \scalebox{0.53}{
    \begin{tabular}{crcc|cccc|cccc}
        \toprule
        & \multirow{2}{*}{\textbf{Model}} & \multirow{2}{*}{\textbf{$L$}} & \multirow{2}{*}{\textbf{\#Param}} & \multicolumn{4}{c|}{\textbf{seed 1}} & \multicolumn{4}{c}{\textbf{seed 2}} \\
        \cline{5-12}
        & & & & \textbf{Test Acc.$\pm$s.d.} & \textbf{Train Acc.$\pm$s.d.} & \textbf{\#Epoch} & \textbf{Epoch/Total} & \textbf{Test Acc.$\pm$s.d.} & \textbf{Train Acc.$\pm$s.d.} & \textbf{\#Epoch} & \textbf{Epoch/Total} \\ 
        \midrule
        \midrule
        \parbox[t]{2mm}{\multirow{9}{*}{\rotatebox[origin=c]{90}{ENZYMES}}} & MLP & 4 & 101481 & 55.833$\pm$3.516 & 93.062$\pm$7.551 & 332.30 & 0.18s/0.17hr & 53.833$\pm$4.717 & 87.854$\pm$10.765 & 327.80 & 0.19s/0.18hr \\
        \cline{2-12}
        & \textit{vanilla} GCN & 4 & 103407 & \textcolor{blue}{\textbf{65.833$\pm$4.610}} & 97.688$\pm$3.064 & 343.00 & 0.69s/0.67hr & 64.833$\pm$7.089 & 93.042$\pm$4.982 & 334.30 & 0.74s/0.70hr\\
        & GraphSage & 4 & 105595 & 65.000$\pm$4.944 & 100.000$\pm$0.000 & 294.20 & 1.62s/1.34hr & \textbf{68.167$\pm$5.449} & 100.000$\pm$0.000 & 287.30 & 1.76s/1.42hr
 \\
        \cline{2-12}
        & MoNet & 4 & 105307 & 63.000$\pm$8.090 & 95.229$\pm$5.864 & 333.70 & 0.53s/0.49hr & 62.167$\pm$4.833 & 93.562$\pm$5.897 & 324.40 & 0.68s/0.62hr \\
        & GAT & 4 & 101274 & \best{68.500$\pm$5.241} & 100.000$\pm$0.000 & 299.30 & 0.70s/0.59hr & \textcolor{blue}{\textbf{68.500$\pm$4.622}} & 100.000$\pm$0.000 & 309.10 & 0.76s/0.66hr \\
        & GatedGCN & 4 & 103409 & \textbf{65.667$\pm$4.899} & 99.979$\pm$0.062 & 316.80 & 2.31s/2.05hr & \best{70.000$\pm$4.944} & 99.979$\pm$0.062 & 313.20 & 2.63s/2.30hr \\
        \cline{2-12}
        & GIN & 4 & 104864 & 65.333$\pm$6.823 & 100.000$\pm$0.000 & 402.10 & 0.53s/0.61hr & 67.667$\pm$5.831 & 100.000$\pm$0.000 & 404.90 & 0.60s/0.68hr \\
        & RingGNN & 2 & 103538 & 18.667$\pm$1.795 & 20.104$\pm$2.166 & 337.30 & 7.12s/6.71hr & 45.333$\pm$4.522 & 56.792$\pm$6.081 & 497.50 & 8.05s/11.16hr \\
        & 3WLGNN & 3 & 104658 & 61.000$\pm$6.799 & 98.875$\pm$1.571 & 381.80 & 9.22s/9.83hr & 57.667$\pm$9.522 & 96.729$\pm$5.525 & 336.50 & 11.80s/11.09hr\\
        \midrule
        \parbox[t]{2mm}{\multirow{9}{*}{\rotatebox[origin=c]{90}{DD}}} & MLP & 4 & 100447 & 72.239$\pm$3.854 & 73.816$\pm$1.015 & 371.80 & 6.36s/6.61hr & \textbf{72.408$\pm$3.449} & 73.880$\pm$0.623 & 349.60 & 1.13s/1.11hr\\
        \cline{2-12}
        & \textit{vanilla} GCN & 4 & 102293 & 72.758$\pm$4.083 & 100.000$\pm$0.000 & 266.70 & 3.56s/2.66hr & \textcolor{blue}{\textbf{73.168$\pm$5.000}} & 100.000$\pm$0.000 & 270.20 & 3.81s/2.88hr\\
        & GraphSage & 4 & 102577 & \textcolor{blue}{\textbf{73.433$\pm$3.429}} & 100.000$\pm$0.000 & 267.20 & 11.50s/8.59hr & 71.900$\pm$3.647 & 100.000$\pm$0.000 & 265.50 & 6.60s/4.90hr \\
        \cline{2-12}
        & MoNet & 4 & 102305 & 71.736$\pm$3.365 & 81.003$\pm$2.593 & 252.60 & 3.30s/2.34hr & 71.479$\pm$2.167 & 81.268$\pm$2.295 & 253.50 & 2.83s/2.01hr \\
        & GAT & 4 & 100132 & \best{75.900$\pm$3.824} & 95.851$\pm$2.575 & 201.30 & 6.31s/3.56hr & \best{74.198$\pm$3.076} & 96.964$\pm$1.544 & 220.10 & 2.84s/1.75hr\\
        & GatedGCN & 4 & 104165 & \textbf{72.918$\pm$2.090} & 82.796$\pm$2.242 & 300.70 & 12.05s/10.13hr & 71.983$\pm$3.644 & 83.243$\pm$3.716 & 323.60 & 8.78s/7.93hr\\
        \cline{2-12}
        & GIN & 4 & 103046 & 71.910$\pm$3.873 & 99.851$\pm$0.136 & 275.70 & 5.28s/4.08hr & 70.883$\pm$2.702 & 99.883$\pm$0.088 & 276.90 & 2.31s/1.79hr\\
        & RingGNN & 2 & 109857 & OOM & OOM & OOM & OOM & OOM & OOM & OOM & OOM\\
        & 3WLGNN & 3 & 104124 & OOM & OOM & OOM & OOM & OOM & OOM & OOM & OOM\\
        \midrule
        \parbox[t]{2mm}{\multirow{9}{*}{\rotatebox[origin=c]{90}{PROTEINS}}} & MLP & 4 & 100643 & 75.644$\pm$2.681 & 79.847$\pm$1.551 & 244.20 & 0.42s/0.29hr & 75.823$\pm$2.915 & 79.442$\pm$1.443 & 241.20 & 0.35s/0.24hr\\
        \cline{2-12}
        & \textit{vanilla} GCN & 4 & 104865 & 76.098$\pm$2.406 & 81.387$\pm$2.451 & 350.90 & 1.55s/1.53hr & \textbf{75.912$\pm$3.064} & 82.140$\pm$2.706 & 349.60 & 1.46s/1.42hr \\
        & GraphSage & 4 & 101928 & 75.289$\pm$2.419 & 85.827$\pm$0.839 & 245.40 & 3.36s/2.30hr & 75.559$\pm$1.907 & 85.118$\pm$1.171 & 244.40 & 3.44s/2.35hr \\
        \cline{2-12}
        & MoNet & 4 & 103858 & \best{76.452$\pm$2.898} & 78.206$\pm$0.548 & 306.80 & 1.23s/1.06hr & \textcolor{blue}{\textbf{76.453$\pm$2.892}} & 78.273$\pm$0.695 & 289.50 & 1.26s/1.03hr\\
        & GAT & 4 & 102710 & \textbf{76.277$\pm$2.410} & 83.186$\pm$2.000 & 344.60 & 1.47s/1.42hr & 75.557$\pm$3.443 & 84.253$\pm$2.348 & 335.10 & 1.51s/1.41hr\\
        & GatedGCN & 4 & 104855 & \textcolor{blue}{\textbf{76.363$\pm$2.904}} & 79.431$\pm$0.695 & 293.80 & 5.03s/4.13hr & \best{76.721$\pm$3.106} & 78.689$\pm$0.692 & 272.80 & 4.78s/3.64hr \\
        \cline{2-12}
        & GIN & 4 & 103854 & 74.117$\pm$3.357 & 75.351$\pm$1.267 & 420.90 & 1.02s/1.20hr & 71.241$\pm$4.921 & 71.373$\pm$2.835 & 362.00 & 1.04s/1.06hr\\
        & RingGNN & 2 & 109036 & 67.564$\pm$7.551 & 67.607$\pm$4.401 & 150.40 & 28.61s/12.08hr & 56.063$\pm$6.301 & 59.289$\pm$5.560 & 222.70 & 19.08s/11.88hr \\
        & 3WLGNN & 3 & 105366 & 61.712$\pm$4.859 & 62.427$\pm$4.548 & 211.40 & 12.82s/7.58hr & 64.682$\pm$5.877 & 65.034$\pm$5.253 & 200.40 & 13.05s/7.32hr \\
        \bottomrule
    \end{tabular}
    }
    \caption{Performance on the TU datasets with 10-fold cross validation (higher is better). Two runs of all the experiments using the same hyperparameters but different random seeds are shown separately to note the differences in ranking and variation for reproducibility. The top 3 performance scores are highlighted as \best{First}, \textcolor{blue}{\textbf{Second}}, \textbf{Third}.
    }
    \label{tab:tableTU}
\end{table}

Apart from the proposed datasets in our benchmark (Section \ref{sec:expsetup}), we perform experiments on 3 TU datasets for graph classification -- ENZYMES, DD and PROTEINS.  
Our goal is to empirically highlight some of the challenges of using these conventional datasets for benchmarking GNNs. 

{\bf Splitting.} Since the 3 TU datasets that we use do not have standard splits, we perform a 10-fold cross validation split which gives 10 sets of train, validation and test data indices in the ratio $8:1:1$. We use stratified sampling to ensure that the class distribution remains the same across splits. The indices are saved and used across all experiments for fair comparisons. There are $480$ train/$60$ validation/$60$ test graphs for ENZYMES, $941$ train/$118$ validation/$119$ test graphs for DD, and $889$ train/$112$ validation/$112$ test graphs for PROTEINS datasets in each of the folds. \\
{\bf Training.} We use Adam optimizer with a similar learning rate strategy as used in our benchmark's experimental protocol. An initial learning rate is tuned from a range of $1\times10^{-3}$ to $7\times10^{-5}$ using grid search for every GNN models. The learning rate reduce factor is $0.5$, the patience value is $25$ and the stopping learning rate is $1\times10^{-6}$.\\
{\bf Performance Measure.} We use classification accuracy between the predicted labels and groundtruth labels as our performance measure. The model performance is evaluated on the test split of the 10 folds for all TU datasets, and reported as the average and the standard deviation of the 10 scores.

Our numerical results on the TU datasets -- ENZYMES, DD and PROTEINS are presented in Table \ref{tab:tableTU}. We observe all NNs have similar statistical test performance as the standard deviation is quite large. We also report a second run of these experiments with the same experimental protocol, \textit{i.e.} the same 10-fold splitting and hyperparameters but different initialization (seed). 
We observe a change of model ranking, which we attribute to the small size of the datasets and the non-determinism of gradient descent optimizers. We also observe that, for DD and PROTEINS, the graph-agnostic MLP baselines perform as good as GNNs. 
Our observations reiterate how experiments on the small TU datasets are difficult to determine which GNNs are powerful and robust.

\section{A Note on Graph Size Normalization}
Intuitively, batching graphs of variable sizes may lead to node representation at different scales, making it difficult to learn the optimal statistics $\mu$ and $\sigma$ for BatchNorm across irregular batch sizes and variable graphs. 
A preliminary version of this work introduced a graph size normalization technique called GraphNorm, which normalizes the node features $h_{i}^{\ell}$  w.r.t. the graph size, \textit{i.e.}, 
\begin{equation}
\label{eqn:gn}
\bar h^{\ell}_{i} = h^{\ell}_{i} \times \frac{1}{\sqrt{\mathcal{V}}},
\end{equation}
where $\mathcal{V}$ is the number of graph nodes. 
The GraphNorm layer is placed before the BatchNorm layer.

We would like to note that GraphNorm does not have any concrete theoretical basis as of now, and was proposed based on initially promising empirical results on datasets such as ZINC and CLUSTER.
Future work shall investigate more principled approaches towards designing normalization layers for graph structured data.

\section{Elaboration on Benchmarking Design Choices}
\label{sec:elaboration_design_choices}
\edits{In Section \ref{sec:overview_gnn_benchmarking_framework}, we provided a brief overview on the design choices that we had to make to build the proposed benchmarking framework. In particular, the decisions on the selection of the specific graph datasets that we have included in this framework, the necessity to constraint model parameters for comparison of GNNs' performance, and whether a standard codebase with data, training, evaluation pipelines is required can be derived from several reasonings. In this section, we provide an elaborate discussion on these factors and how possible extensions can be developed in future with ease and as per required by a research agenda.

\noindent\textbf{Datasets.} Our collection of datasets is based on medium-scale size and criteria of diversity in terms of the end-application domains, learning tasks at graph-, edge-, or node-levels, and their source of construction being real or mathematical. The medium-scale size of datasets enables quick prototyping of novel ideas and robust analysis could be generated in single experiments in as less as 12 hours of maximum time per experiment. Similarly, the diversity ensures a model can be tested on not just one end-application domain but a number of such domains.
However, despite the best efforts, after any collection of datasets in such a research area where a general GNN architecture is expected to be robust to a variety of tasks and domains, there could always be need of additional datasets. Due to this necessity, the proposed framework can be extended with new datasets conveniently by any researchers adopting it. We have also observed the open-sourced GitHub repo of our framework being used accordingly with an example repo being \url{https://github.com/karl-zhao/benchmarking-gnns-pyg} \citep{zhao2020pipeline} which extends the framework with additional node classification datasets as well as adopts it in Pytorch Geometric \citep{fey2019fast} instead of DGL. Such adoption of our framework demonstrates its flexibility and the supported convenient extensions. We provide detailed instructions on adding new datasets to the framework in our GitHub repository's \texttt{README} .

\noindent\textbf{Parameter Budgets.}
As we have already mentioned, we designed the framework with the objective that it is used to conveniently `identify first principles' in GNNs' research and not drive a model towards achieving SOTA performance. To enforce this, a straightforward and sound choice is to constraint model parameters and fix it to a specific number (as eg. 100k and 500k) when comparing two or more GNNs. With this choice, we can likely rely on the inference that performance gains are coming from architectural designs and not merely large trainable parameters. The parameter budgeting also tells that the proposed framework may not be ideal to optimize a model to achieve SOTA by tuning hyperparameters, increasing model size to as much parameters as a server can fit, etc. However, we believe we condition the framework to be suitable for identifying performance trends and infer which first principles work robustly across different model experiments. Once such principles are identified, models can further be scaled without any constraints to achieve SOTA performance targeted benchmarks, beyond the datasets we included here. 

\noindent\textbf{Codebase.} A major contribution of this work is the release of the open-source coding infrastructure on GitHub. As observed since the first release in March 2020, the framework has been used extensively to develop new ideas in the field. In the existing literature prior to this work \citep{errica2019fair}, it was a major issue that different research papers in this field adopted inconsistent model comparison methods. Our framework addresses this need of having a standard codebase that helps in training and evaluating GNNs on a collection of appropriate datasets with consistent settings. While a limiting perspective to such codebase can be that it restricts on the diverse choices which researchers often adopt in deep learning to fully realise the capabilities of a model, we understand that we have set out specific objectives of the need of the proposed coding infrastructure and any extensions with other training settings to the codebase can be done by augmenting methods or modules that applies to each model in a fair and consistent way.
}

\section{Hardware}
\label{sec:hardware}
Timing research code can be tricky due to differences of implementations and hardware acceleration.
Nonetheless, we take a practical view and report the average wall clock time per epoch and the total training time for each model.
All experiments were implemented in DGL/PyTorch. 
We run experiments for MNIST, CIFAR10, ZINC, AQSOL, TSP, COLLAB, WikiCS, CSL, CYCLES, GraphTheoryProp and TUs on an Intel Xeon CPU E5-2690 v4 server with 4 Nvidia 1080Ti GPUs (11 GB), and for PATTERN and CLUSTER on an Intel Xeon Gold 6132 CPU with 4 Nvidia 2080Ti (11 GB) GPUs.
Each experiment was run on a single GPU and 4 experiments were run on the server at any given time (on different GPUs).
We run each experiment for a maximum of 12 hours.

\section{Memory Usage}
\begin{figure}[t]
\begin{tabular}{cc}
\subfloat{\includegraphics[width = 2.9in]{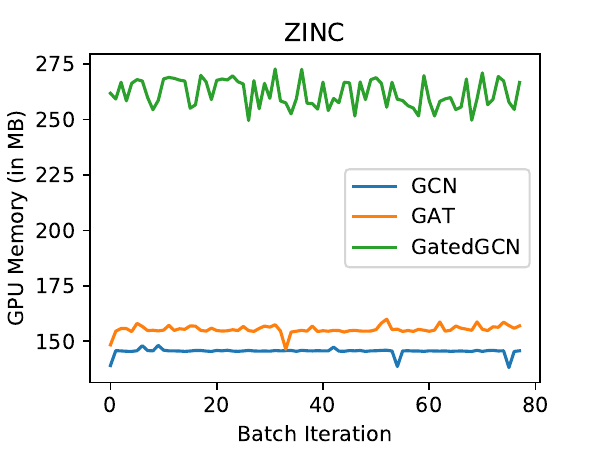}} &
\subfloat{\includegraphics[width = 2.9in]{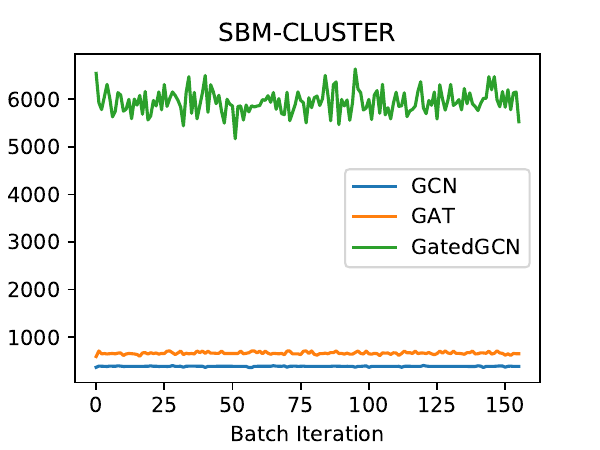}} \\
\end{tabular}
\caption{Memory consumed by the GPU device in a forward pass of a batch during training. All GNNs shown here have 500k learnable parameters. Batch size is 128 for ZINC and 64 for SBM-CLUSTER. }
\label{fig:gpu_memory}
\end{figure}

For datasets that contain graphs with variable sizes, the memory consumed during training by the GPU device changes at each batch of graphs. We report in Figure \ref{fig:gpu_memory} the GPU memory consumption during the training of GCN, GAT and GatedGCN on two datasets--ZINC and SBM-CLUSTER. The plots show the memory allocated during the model's forward pass using a batch of graphs (128 graphs for ZINC and 64 for SBM-CLUSTER) and is computed by using PyTorch's \texttt{torch.cuda.memory\_allocated(device)} functionality. Overall, it can be observed that GatedGCN is a relatively higher memory-intensive model as compared with GCN and GAT, see Section \ref{sec:mpgcns} for the respective models' equations.

\bibliography{neurips_2020}

\end{document}